\documentclass[11pt,a4paper,twoside]{article}

\usepackage{amsmath}
\usepackage{amssymb}
\usepackage{graphicx}
\usepackage{hyperref}
\usepackage[dvipsnames]{xcolor}
\usepackage{subcaption}
\usepackage[authoryear]{natbib}
\usepackage{float}
\usepackage{geometry}

\advance\textheight by \topskip
\usepackage{fancyhdr}
\newlength{\preXLskip}
\newlength{\preLskip}
\newlength{\preMskip}
\newlength{\preSskip}
\newlength{\postMskip}
\newlength{\postSskip}
\usepackage{hyperref}
\definecolor{Red}{rgb}{0.5,0,0}
\definecolor{Blue}{rgb}{0,0,0.5}
\hypersetup{%
hyperindex = {true},
colorlinks = {true},
linktocpage = {true},
plainpages = {false},
linkcolor = {Blue},
citecolor = {Blue},
urlcolor = {Black},
pdfstartview = {Fit},
pdfpagemode = {None},
pdfview = {XYZ null null null}
}

\RequirePackage{amsmath,amsthm,amsfonts,amssymb,mathtools}
\RequirePackage{mathrsfs}
\RequirePackage{caption}
\RequirePackage{subcaption}
\usepackage{float}

\RequirePackage{cleveref} 
\usepackage{algpseudocode} 
\usepackage{algorithm}     
\usepackage{fontawesome}

\numberwithin{equation}{section}

\theoremstyle{plain}

\newtheorem{theorem}{Theorem}[section]

\theoremstyle{definition}

\newtheorem{remark}[theorem]{Remark}

\makeatother

\theoremstyle{remark}

\newcounter{casecount}
\AtBeginEnvironment{proof}{\setcounter{casecount}{0}}

\newcounter{stepcount}
\AtBeginEnvironment{proof}{\setcounter{stepcount}{0}}

\newcommand{\samplesize}{n}
\newcommand{\dimension}{p}
\newcommand{\iteration}{m}
\newcommand{\designmatrix}{A}
\newcommand{\response}{Y}
\newcommand{\truesignal}{f^{*}}
\newcommand{\noiselevel}{\delta}
\newcommand{\noise}{\varepsilon}

\newcommand{\DP}{\tau^{\text{DP}}}

\newcommand{\estimator}{\widehat{f}}
\newcommand{\threshold}{\kappa}

\newcommand{\package}{\texttt{EarlyStopping}-package}
\newcommand{\learningrate}{\omega}

\newcommand{\strongrisk}[2]{\mathcal{R}( {#1}, {#2} )}
\newcommand{\weakrisk}[2]{\mathcal{R}_{\designmatrix}( {#1}, {#2})}

\DeclareMathOperator*{\argmin}{arg\!\min}
\DeclareMathOperator*{\argmax}{arg\!\max}

\DeclareMathOperator*{\tr}{tr}

\providecommand{\norm}[1]{\lVert #1 \rVert}
\providecommand{\abs}[1]{\lvert #1 \rvert}

\usepackage{color}
\definecolor{deepblue}{rgb}{0,0,0.5}
\definecolor{deepred}{rgb}{0.6,0,0}
\definecolor{deepgreen}{rgb}{0,0.5,0}
\definecolor{shorts}{RGB}{52, 125, 235}
\definecolor{basecommands}{RGB}{24, 0, 255}
\definecolor{ourpackage}{RGB}{0, 47, 255}
\definecolor{literate}{RGB}{109, 115, 122}
\definecolor{numbers}{RGB}{12, 77, 117}

\DeclareFixedFont{\ttb}{T1}{txtt}{bx}{n}{9} 
\DeclareFixedFont{\ttm}{T1}{txtt}{m}{n}{9}  

\usepackage{listings}

\newcommand\pythonstyle{\lstset{
language=Python,
xleftmargin=2em, 
xrightmargin=2em, 
basicstyle=\small\ttm,
alsoletter={1234567890},
showstringspaces=false,
morekeywords={=, *, +, -},
otherkeywords={=, *, +, -},
keywordstyle=\ttb\color{literate},
emph={MyClass, TruncatedSVD, SimulationData, SimulationWrapper, SimulationParameters, L2_Boost, Landweber, ConjugateGradients, Algorithm, Dataframe, RegressionTree},          
emphstyle=\ttb\color{deepred},    
emph={[2]np, plt, es, pd},
emphstyle=[2]\color{shorts},
emph={[3]import, as, from, class, def, self, None, if, is, not, and, for, in, else, return, while, alg},
emphstyle=[3]\color{basecommands},
emph={[4], },
emphstyle=[4]\color{ourpackage},
emph={[5]+},
emphstyle=[5]\color{literate},
emph={[6]sin, arrange, exp, diag, abs, normal, iterate, get_quantity, iterate, figure, show, plot, get_weak_balanced_oracle, get_strong_balanced_oracle, get_discrepancy_stop, get_estimate, ylim, sqrt, dia_matrix, __init__, shape, array, sum, get_stopping_index, get_oracle_quantity, __algorithm_one_iteration, range, __update_strong_empirical_risk, __update_weak_empirical_risk, argmax, argmin, run_simulation_conjugate_gradients,run_simulation_truncated_svd,  run_simulation_landweber, get_residual_ratio_stop, get_aic_iteration, get_noise_estimate, get_balanced_oracle, multivariate_normal, get_weak_empirical_oracle, get_strong_empirical_oracle, get_strong_empirical_risk, get_weak_empirical_risk, arange, identity, zeros, __update_oracle_quantity},
emphstyle=[6]\color{deepblue},
emph={[7]0,1,2,3,4,5,6,7,8,9,01, 002, 5000, 250, 10000, 3000, 14, 2000, 500, 10, 1000, 200, True, False, 100},
emphstyle=[7]\color{numbers},
stringstyle=\color{deepgreen},
frame=single,
showstringspaces=false,
escapechar=\%,
literate=%
}}

\lstnewenvironment{python}[1][]
{
\pythonstyle
\lstset{#1, captionpos=b}
}
{}

\crefname{lstlisting}{listing}{listings}
\Crefname{lstlisting}{Codeblock}{Listing}

\newcommand\pythoninline[1]{{\pythonstyle\lstinline!#1!}}

\title{EarlyStopping: Implicit Regularization for Iterative Learning Procedures in Python}
\author{Eric Ziebell\textsuperscript{1}, Ratmir Miftachov\textsuperscript{1,2}, 
Bernhard Stankewitz\textsuperscript{3}, Laura Hucker\textsuperscript{1}\\[2ex]
\normalsize\textsuperscript{1}Institute of Mathematics, Humboldt-Universität zu Berlin\\
\normalsize\textsuperscript{2}School of Business and Economics, Humboldt-Universität zu Berlin\\
\normalsize\textsuperscript{3}Institute of Mathematics, Universität Potsdam
}

\defcitealias{irtools}{\textsc{IRtools}}
\defcitealias{regtools}{\textsc{regtools}}
\defcitealias{xgboost}{{\textsc{XGBoost}}}
\defcitealias{keras}{\textsc{Keras}}
\defcitealias{scikit}{\textsc{scikit-learn}}
\defcitealias{lightgbm}{\textsc{lightgbm}}
\defcitealias{pytorch}{\textsc{Pytorch}}
\defcitealias{tripspy}{\textsc{trips-py}}
\defcitealias{numpy}{\textsc{Numpy}}
\defcitealias{scipy}{\textsc{SciPy}}
\defcitealias{pandas}{\textsc{Pandas}}

\begin{document}
\maketitle

\begin{abstract}
\noindent Iterative learning procedures are ubiquitous in machine learning and modern statistics.
  Regularision is typically required to prevent inflating the expected loss of a procedure in
  later iterations via the propagation of noise inherent in the data.
  Significant emphasis has been placed on achieving this regularisation implicitly by stopping
  procedures early.
  The EarlyStopping-package provides a toolbox of (in-sample) sequential early stopping rules for
  several well-known iterative estimation procedures, such as truncated SVD, Landweber (gradient
  descent), conjugate gradient descent, L2-boosting and regression trees.
  One of the central features of the package is that the algorithms allow the specification of the
  true data-generating process and keep track of relevant theoretical quantities.  
  In this paper, we detail the principles governing the implementation of the \package{} and provide
  a survey of recent foundational advances in the theoretical literature.
  We demonstrate how to use the \package{} to explore core features of implicit regularisation
  and replicate results from the literature.
\end{abstract}

\section{Introduction}
\label{sec_Introduction}

Iterative learning procedures are ubiquitous in machine learning and modern statistics.
They naturally arise from the fact that estimators can often be characterised as solutions to
optimisation problems which have to be approximated iteratively.
Additionally, in the context of high-dimensional data sets, they are instrumental in making large
scale problems computationally tractable.
Any individual step of an iterative procedure is typically cheap to compute, and features of the
data such as smoothness or sparsity often guarantee good estimation results after only a few
iterations.
Consequently, such methods can still be successful in settings where closed-form estimators become
prohibitively expensive or infeasible.
In statistical settings, iterative procedures typically have to be regularised to prevent inflating
the expected loss (risk) of the procedure in later iterations via the propagation of noise inherent
in the data. 
Regularisation can be \emph{explicit} via optimising a penalised loss as in the case of Tikhonov
regularization \citep{EnglEtal1996InverseProblems} or the Lasso \citep{tibshirani1996regression},
implemented for instance in \citetalias{scikit}, \citetalias{regtools}, \citetalias{irtools} and
\citetalias{tripspy}.
Alternatively, the same effect can be achieved \emph{implicitly} by stopping procedures early before
convergence.
Based on recent advances in the statistical literature, we introduce the \package{} implementing
data-driven early stopping rules in the context of inverse problems and regression settings, which
are both statistically and computationally efficient.

Essentially, implicit regularisation translates to the choice of an appropriate iteration number 
\( \widehat{m} \) of the learning procedure.
Classically, this could be regarded as a model selection problem and be addressed via
cross-validation, information criteria such as AIC and BIC, or Lepski's balancing principle.
However, all the \emph{model selection criteria} above require computing the full iteration path of
the procedure to determine $\widehat{m}$, which is fundamentally opposed to saving computational
resources by only computing as many iterations as necessary.
To retain computational efficiency, \emph{early stopping procedures} therefore aim to choose
$\widehat{m}$ \emph{sequentially}, i.e. the moment we have reached $\widehat{m}$, we stop immediately.

One common approach to choosing \( \widehat{m} \) sequentially is based on splitting the observed
sample into a training and a validation set.
The learning procedure is then computed on the training set and stopped when its performance on the
validation set drops below a certain threshold sufficiently often, see \citet{prechelt2002early}. 
While there are not many theoretical guarantees for sequential sample splitting, in practical
application, it is often considered state of the art and is part of most major machine learning
frameworks such as \citetalias{keras}, \citetalias{xgboost}, \citetalias{scikit},
\citetalias{lightgbm} and \citetalias{pytorch}.

From the theoretical side of the literature, more interest has been garnered by sequential
(in-sample) early stopping rules that avoid sample splitting.
On the one hand, only in specific machine learning settings, users can afford to sacrifice a
significant portion of the observations.
On the other hand, sample splitting is unsuitable if the dataset is highly interdependent or
constitutes a single observation of a complex underlying distribution, e.g., in the case of
inverse problems as treated in \citet{BlanchardMathe2012DiscrepancyPrinciple},
\citet{blanchard2018early,blanchard2018optimal}, \citet{Stankewitz2020Smoothed},
\citet{MikaSzkutnik2021DiscrepancyPrinciple}, \citet{Jahn2022Discrepancy},
\citet{hucker2024early} or kernel learning in \citet{CelisseWahl2021Discrepancy}. 
Additionally, optimal early stopping may depend on quantities that cannot be estimated
from a split sample such as an empirical in-sample noise level, see \citet{stankewitz2024early}, 
\citet{kueck2023estimation} and \citet{miftachov2025early}.

These stopping rules have not found their way into mainstream software libraries yet, and
implementations mostly exist isolated in specialised packages, see \citetalias{regtools} or
\citetalias{tripspy}. 
In particular, to our knowledge, there exists no unified library for these closely related
procedures.
Our \package{} fills this gap and provides a toolbox to analyse and experiment with (in-sample)
sequential early stopping rules for several well-known iterative estimation procedures, such as
truncated SVD, Landweber (gradient descent), conjugate Gradient descent, L2-boosting and regression
trees.

The \package{} primarily builds on the theoretical foundations in
\cite{blanchard2018early,blanchard2018optimal}, \cite{hucker2024early}, \cite{stankewitz2024early}, 
\cite{kueck2023estimation} and \cite{miftachov2025early}.
\citet{blanchard2018early} initially analyse the estimation performance of a residual-based
stopping rule (discrepancy stop) for the truncated singular value decomposition (truncated SVD) in a
statistical inverse problem and provide statistical guarantees for the estimation performance. 
They also explore theoretical limitations of stopping rules in terms of minimax lower bounds leading
to the development of an adaptive two-step procedure.
As its default, this procedure uses a sequential stopping rule and guarantees optimal performance
even in pathological cases via an AIC criterion applied to only the iterations up to the stopping
time.
An extension of the adaptation bounds for a residual-based stopping rule is provided by
\citet{blanchard2018optimal} for general spectral regularisation methods such as the Landweber
(gradient descent) iteration. 
One of the computationally most efficient methods for solving systems of linear equations is the
conjugate gradient (CG) algorithm. 
The CG algorithm depends highly non-linearly on the observations, making the theoretical analysis
particularly intricate. 
Adaptation bounds were nevertheless achieved by \citet{hucker2024early} for the discrepancy stop. 

For $L^{2}$-boosting via orthogonal matching pursuit, \citet{stankewitz2024early} showed adaptivity
for the discrepancy stop in high-dimensional linear models given a suitable in-sample noise
estimate, which can be obtained via the scaled Lasso studied in \cite{SunZhang2012ScaledLasso}.
\citet{kueck2023estimation} show that under sparsity assumptions, another stopping rule, the
residual ratio stop, exists that does not rely on an additional noise estimator.
A recent contribution by \citet{miftachov2025early} extends the Early Stopping paradigm to
non-parametric regression using the classification and regression tree (CART) algorithm by
\citet{breiman1984classification}.
The proposed breadth-first search and best-first search early stopping algorithms for constructing a
regression tree are embedded in a much broader framework of generalised projection flows of
iterative regression estimators. 

The \package{} gathers the different early stopping methods from the literature above within one
common framework and provides prototypical implementations of the iterative algorithms in unified
class structures.
One of the central features is that the classes additionally allow the specification of the true
data-generating process and are therefore able to keep track of all relevant theoretical (oracle)
quantities, such as explicit bias-variance decompositions or theoretical risk minimisers.
For both researchers and practitioners, the \package{} can therefore provide a common frame of
reference for experimentation, since in controlled settings, the true performance of the
procedure can immediately be made explicit.
The \package{} also provides a simulation wrapper class, which yields seamless (one-line) executions
of Monte-Carlo simulations.
In the interest of reproducible research, this allows the instant replication of simulation studies
from the literature. 
The \package{} is written in \textsc{Python}, and its code repository and documentation are
available on GitHub under the following URLs:
\begin{itemize}
  \label{itemize:links}
  \item[] \textsc{Repository}: \href{https://github.com/EarlyStop/EarlyStopping}{github.com/EarlyStop/EarlyStopping},
  \item[] \textsc{Documentation}:
  \href{https://earlystop.github.io/EarlyStopping/}{earlystop.github.io/EarlyStopping},
\end{itemize}

The paper consists of two main parts.
\Cref{sec_BackgroundOnEarlyStopping} introduces the necessary theoretical context and provides a survey of the existing early stopping literature.
\Cref{sec:implementation} explains the implementation of the \package{} in detail, provides examples
and replicates major results from the literature.
Both sections feature detailed explanations for each of the iterative estimation procedures.


\newpage
\section{Background on early stopping}
\label{sec_BackgroundOnEarlyStopping}

\begin{figure}[htb]
  \centering
  \begin{subfigure}[b]{0.49\textwidth}
    \centering
    \includegraphics[width=\textwidth]{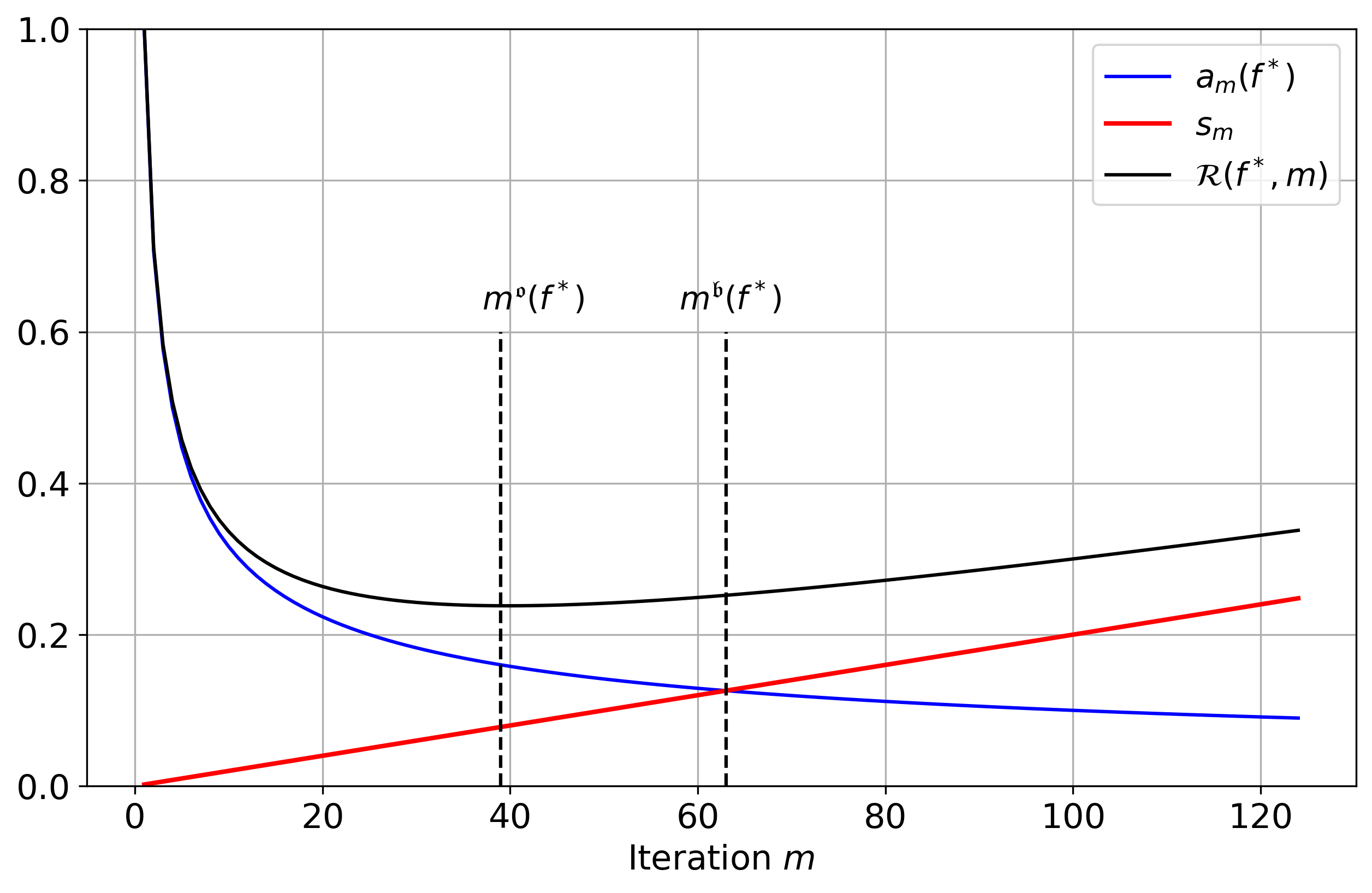}
    \caption{Risk decomposition for a signal $f^{*}$.}
  \end{subfigure}
  \hfill
  \begin{subfigure}[b]{0.49\textwidth}
    \centering
    \includegraphics[width=\textwidth]{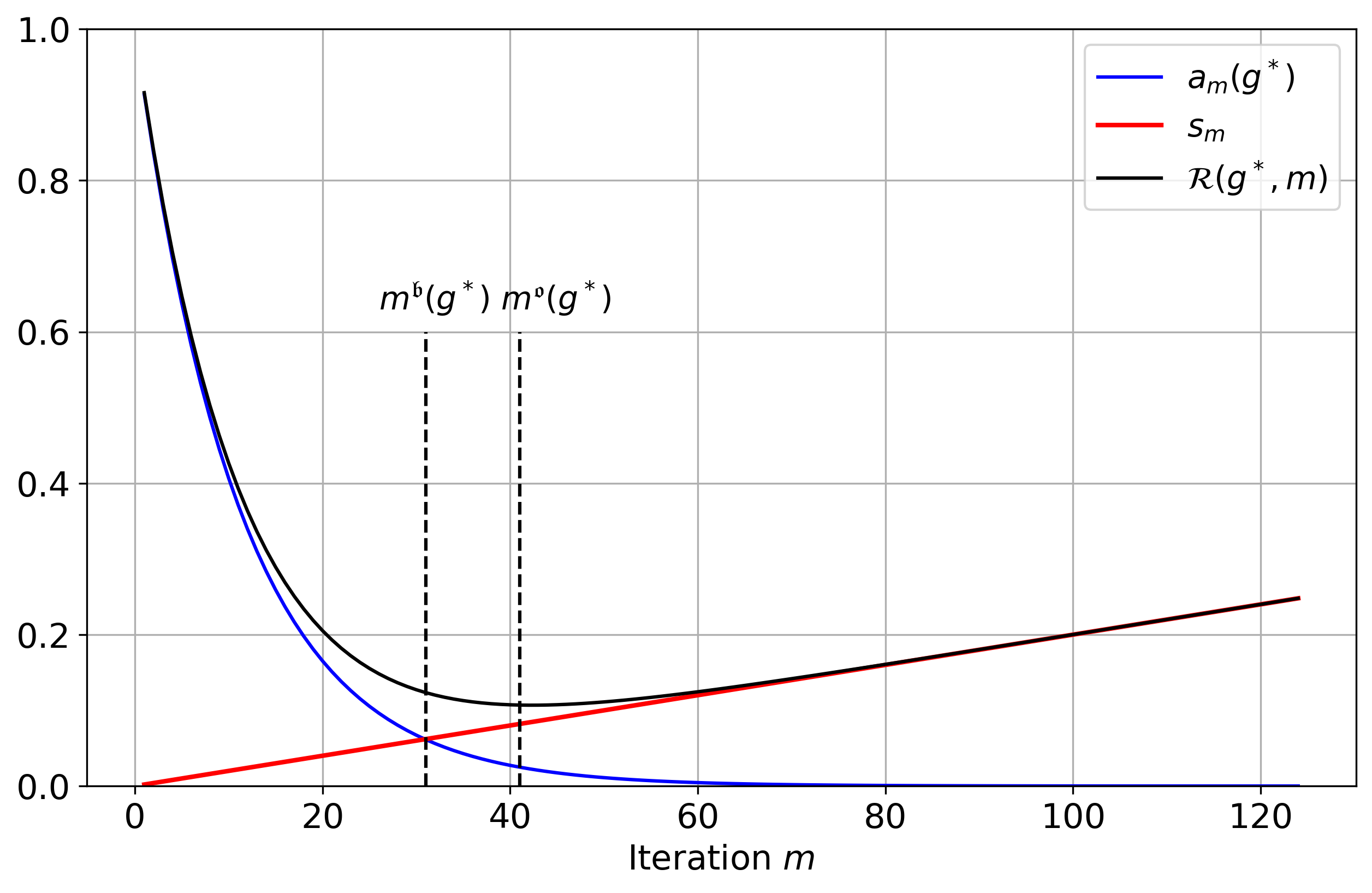}
    \caption{Risk decomposition for a signal $g^{*}$.}
  \end{subfigure}
  \caption{Decomposition of the risk into approximation and stochastic error for two signals 
           \( f^{*} \) and \( g^{*} \) with slowly and fast decaying squared bias respectively.
          }
  \label{fig:GeneralRiskDecomposition}
\end{figure}

Iterative methods have become omnipresent in machine learning and modern statistics.
They naturally arise from the fact that estimators can often be characterised as solutions of
optimisation problems, which have to be solved iteratively.
In statistical settings, when observations are noisy, unregularised algorithms typically cannot be iterated
indefinitely and have to be \emph{stopped early}, i.e. before they converge.
Indeed, the risk given by the expected loss at iteration \( m \) of a sequence of iteratively
computed estimators \( (\estimator^{(m)})_{m \geq 0} \) typically can be decomposed
into
\begin{equation}
\label{eq:general_error_decomposition}
  \mathcal{R}(f^{*}, m) = b_{m}^{2}(f^{*}) + s_{m}, \qquad m \geq 0,
\end{equation}
where \( b_{m}^{2}(f^{*}) \) is a squared bias or approximation error depending on the underlying truth
\( f^{*} \) and \( s_{m} \) is a stochastic error.
The mapping \( m \mapsto b_{m}^{2}(f^{*}) \) is usually decreasing, as later iterations reduce the
approximation error of the iterative procedure, whereas \( m \mapsto s_{m} \) is increasing, since
it accounts for the propagation of the noise in the observations, see
\Cref{fig:GeneralRiskDecomposition}.

Learning \( f^{*} \) amounts to (approximately) minimising the risk \( \mathcal{R}(f^{*}, m) \) with
respect to \( m \geq 0 \).
However, since \( b_{m}^{2}(f^{*}) \) depends on the unknown underlying truth \( f^{*} \), this is
also true for the minimiser
\begin{equation}
  m^{\mathfrak{o}} = m^{\mathfrak{o}}(f^{*}) = \argmin_{m \geq 0} \mathcal{R}(f^{*}, m).
\end{equation}
Consequently, the iteration number \( \hat{m} \) that we ultimately choose for the procedure has to be
adaptive. This means that it is chosen depending on the data in order to distinguish between both different settings in Figure \ref{fig:GeneralRiskDecomposition}.

In case that the full path of estimators \( (\estimator^{(m)})_{m \geq 0} \) is known, criteria from the
model selection literature such as cross-validation \citet{ArlotCelisse2009CrossValidation}, 
information criteria such as the Akaike information criterion (AIC) and the Bayesian information
criterion (BIC) \citet{StoicaSelen2004ModelSelection} or Lepski's balancing principle
\citep{lepski1997optimal} provide adaptive choices of \( \hat{m} \).
However, computing the full path of estimators \( (\hat{f}^{(m)})_{m \geq 0} \) is
fundamentally opposed to the idea of iterative estimation.
In the case of the signal \( g^{*} \) in 
\Cref{fig:GeneralRiskDecomposition}, by computing the full iteration path to then 
select an iteration \( \hat{m} \) close to \( m^{\mathfrak{o}}(g^{*}) \), 
more than \( 75\% \) of computational resources would have been spent on applying a model selection
criterion.
Therefore, we should aim to choose \( \hat{m} \) sequentially to increase the computational efficiency of the method. This means that when we have reached \( \hat{m} \), we stop immediately.

Prima facie, it is not obvious that such rules should exist.
For example, a sequential choice \( \hat{m} \) can never directly mimic the (global) classical
oracle \( m^{ \mathfrak{o} } \), since \( \hat{m} \) does not incorporate information about 
\( b_{m}^{2}(f^{*}) \) for \( m > \hat{m} \). 
However, it can be shown that the (sequential) \emph{balanced oracle}
\begin{align}
  \label{eq_I_BalancedOracle}
      m^{ \mathfrak{b} } = m^{ \mathfrak{b} }( f^{*} ): 
  = 
      \inf \{ m \geq 0: b_{m}^{2}( f^{*} ) \le s_{m} \}
\end{align}
satisfies
\begin{align}
  \label{eq_OptimalityOfBalancedOracle}
        \mathcal{R}( f^{*}, m^{ \mathfrak{b} } ) 
  \le 
        2
        \mathcal{R}( f^{*}, m^{ \mathfrak{o} } ) 
      + 
        \Delta( s_{ m^{ \mathfrak{b} } } )
  = 
        2 
        \min_{ m \geq  0} 
        \mathcal{R}( f^{*}, m ) 
      + 
        \Delta( s_{ m^{ \mathfrak{b} } } ),
\end{align}
where \( \Delta( s_{m} ): = s_{m} - s_{ m - 1 } \) is the discretisation error of the stochastic
error at iteration \( m \), see \citet[Lemma I.1.1]{Stankewitz2023Thesis}.
In many settings, data-driven sequential stopping rules can be constructed that mimic 
\( m^{\mathfrak{b}} \), which is explained in more detail throughout the next section.

\subsection{Early stopping for inverse problems}
\label{sec_EarlyStoppingForInverseProblems} 

We consider a general finite dimensional statistical inverse problem
\begin{equation}
  \label{eq:inverse_problem}
  \response = \designmatrix \truesignal + \noiselevel \noise,
\end{equation}
where $ A \in \mathbb{R}^{n \times p} $ is a large scale matrix, $\truesignal \in \mathbb{R}^{n} $ is an unknown coefficient signal to be recovered and $\noise \in \mathbb{R}^{n} $ is standard Gaussian noise.
Typically, formulations as in \eqref{eq:inverse_problem} stem from discretisations of potentially infinite-dimensional inverse problems, see \citet{EnglEtal1996InverseProblems}.

Assuming that $A$ is injective, $\designmatrix^{\top} \designmatrix$ is invertible and the classical
least squares estimator is given by 
\begin{equation}
  \label{eq_OLSEstimator}
  \estimator : = \argmin_{f \in \mathbb{R}^{p}} \| Y - \designmatrix f \|^{2}
                     = ( \designmatrix^{\top} \designmatrix )^{-1} \designmatrix^{\top} \response
                     = \sum_{j = 1}^{\min(n, p)} \lambda_j^{-1} v_j^{\top} Y u_j.
\end{equation}
The last expression is in terms of the singular value decomposition (SVD)
\begin{equation}
 \label{eq:SVD:designmatrix}
  \designmatrix = V \Lambda U^{\top} = \sum_{j = 1}^{\min(n, p)} \lambda_j v_j u_j^{\top}
\end{equation}
of the design matrix $\designmatrix$.
If the inverse $(\designmatrix^{\top}\designmatrix)^{-1}$ does not exist, e.g. due to ill-posedness,
classical regularisation methods replace $\lambda_j^{-1}$ on the right hand side of
\eqref{eq_OLSEstimator} with $ \lambda_{j} g_{\alpha}(\lambda_j^{2}) $, where
$ (g_{\alpha})_{\alpha>0}$ is an approximating family of piecewise continuous functions satisfying
that for any \( \lambda \in (0, \infty) \),
\begin{equation}
  \label{eq_ApproximationFunctionGAlpha}
  g_{\alpha}(\lambda) \to \lambda^{-1} \qquad \text{ for } \alpha \to 0 .
\end{equation}
For simplicity, we assume in the following that the inverse problem in Equation
\eqref{eq:inverse_problem} is discretised appropriately, i.e., that \( A \) is surjective and 
\( n \le p \).
The regularised estimator is then given by
\begin{equation}
  \label{eq:spectral_regularisation_methods}
  \estimator_{\alpha} 
  \coloneqq
  g_{\alpha}(\designmatrix^{\top}\designmatrix) \designmatrix^{\top}\response
  =
  \sum_{j=1}^n \lambda_{j} g_{\alpha}(\lambda_j^{2}) v_j^{\top}\response u_j.
\end{equation}
Most of the standard regularisation methods can be represented in this way, where \( \alpha \to 0 \)
is interpreted as a continuous idealization of the iterative computation scheme of estimators \(
\widehat{f}^{(m)} \) with iteration number \( \iteration \to \infty \). 

Reparameterised in terms of the iteration $\iteration$, the risk expressed as the mean integrated
squared error (MISE) 
\begin{equation}
  \label{eq:strong_risk}
    \mathcal{R}(\truesignal, \iteration):
  =
    \mathbb{E}\|\estimator^{(\iteration)}-\truesignal\|^2
  = 
    b_\iteration^2(\truesignal) + s_\iteration
\end{equation}
satisfies an error decomposition as in Equation \eqref{eq:general_error_decomposition}.
The approximation error, given by the squared bias 
$ b_\iteration^{2}(\truesignal): = \| \mathbb{E} \estimator^{(m)} - \truesignal \|^{2}$,
is decreasing in $\iteration$, whereas the stochastic error, given by the variance 
$ s_\iteration: = \mathbb{E}\| \widehat{f}^{(m)} - \mathbb{E} \estimator^{(m)} \|^{2} $,
is increasing in $\iteration$. 
Structurally, the same is true for the prediction risk
\begin{align}
  \label{eq:weak_risk}
    \mathcal{R}_{\designmatrix}(\truesignal, \iteration):
  = 
    \mathbb{E} \| \designmatrix (\estimator^{(\iteration)} - \truesignal) \|^{2}
  =
    b_{\iteration, \designmatrix}^{2}(\truesignal) + s_{\iteration, \designmatrix},
\end{align}
which also decomposes into a decreasing approximation error
\( b_{m, A}^{2}(\truesignal): = \| \designmatrix(\mathbb{E} \estimator^{(\iteration)} - \truesignal) \|^{2} \) 
and an increasing noise propagation part
\( s_{m, A}: = \mathbb{E}\|\designmatrix( \widehat{f}^{(\iteration)} - \mathbb{E} \estimator^{(m)} )\|^{2} \). 
Following the terminology from \citet{blanchard2018early}, we call the MISE the strong risk and the
prediction risk the weak risk.
Given the decompositions \eqref{eq:strong_risk} and \eqref{eq:weak_risk}, we can define strong and
weak versions
\begin{align}
  \label{eq:balanced_oracles_weak_strong}
  m^{\mathfrak{b}}_s: = \inf \{ m \geq 0: b_m^2(\truesignal) \leq s_m \},
  \quad
  m^{\mathfrak{b}}_w: = \inf \{ m \geq 0: b_{m, A}^2(\truesignal) \leq s_{m, A} \},
\end{align}
of the balanced oracle in Equation \eqref{eq_I_BalancedOracle}.

In order to stop the learning procedure given by the estimators \( (\widehat{f}^{(m)})_{ m \geq 0} \), we
focus on the \emph{discrepancy principle}, which is likely the most prominent sequential iteration parameter choice rule in the inverse problem literature, see
\citet{EnglEtal1996InverseProblems} and \citet{Werner2018OracleInequalities}.
For a user-specified critical value \( \kappa > 0 \), it prescribes stopping at
\begin{equation}
  \label{eq:discrepancy_stop}
  \DP: = \min \{ m \geq 0: \| Y - A \widehat{f}^{(m)} \|^{2} \le \kappa \}, 
\end{equation}
that is, the first iteration at which the remaining data misfit given by squared residuals 
\( \| Y - A \widehat{f}^{(m)} \|^{2} \) is smaller than \( \kappa \).
In many settings, it can be shown that 
\begin{align}
  \label{eq_ExpectedResidualsForIntuition}
  \mathbb{E} \| Y - A \widehat{f}^{(m)} \|^{2} = b_{m, A}^{2}(\truesignal) - s_{m, A} + \samplesize \delta^{2},
\end{align}
indicating that in expectation, the discrepancy stop \( \DP \) mimics the weak balanced oracle when
\( \kappa \) is approximately \( \samplesize \delta^{2} \).
We consider this in more explicit detail for the learning procedures implemented in the \package.

\subsubsection{Truncated SVD}
\label{sssec_TruncatedSVD}

The spectral cut-off (or truncated SVD) estimator emerges considering the family of functions
$ g_{ \alpha }( \lambda ) \coloneqq \lambda^{-1} \mathbf{1}_{ [ \alpha, \infty ] }( \lambda )$
in Equation \eqref{eq_ApproximationFunctionGAlpha} with 
the discretisation \( \alpha_{m}: = \lambda_{m}^{2} \), \( m \geq 0 \).
As an iterative estimation procedure, the updating rule is given by
\begin{equation}
  \label{eq_TruncatedSVDEstimator}
        \estimator^{(0)}=0, 
  \quad
      \estimator^{(\iteration + 1)} 
  =
      \estimator^{(\iteration)} 
    + 
      \frac{v_{\iteration+1}^{\top} \response }{\lambda_{\iteration+1}} u_{\iteration+1}
  = 
      \sum_{j=1}^{\iteration} \frac{v_{j}^{\top} \response}{\lambda_{j}}  u_{j}. 
\end{equation}
For these estimators, the full SVD is usually unavailable and must be computed.
The calculation of the largest singular value is less costly if deflation or locking methods are
used.
In the \package, the singular values are computed iteratively using the power method and by
sequentially removing the largest singular value from the design through
\begin{equation}
  \designmatrix_0 = \designmatrix, \quad \designmatrix_{i+1} = \designmatrix_{i}-\lambda_i u_i v_i^{\top}. 
\end{equation}
Based on this methodology, the required singular values and vectors are computed with roughly
$O(\iteration n^{2})$ multiplications. 
This results in a computational complexity of \( O(\DP n^{2}) \) for the stopped algorithm compared
to \( O(n^{3}) \) when considering the whole learning trajectory.

The strong risk \eqref{eq:strong_risk} and weak risk \eqref{eq:weak_risk} of the estimators
above are given by
\begin{equation}
  \label{eq:truncatedSVD_weak_strong_tradeoff}
  \begin{aligned}
      \mathcal{R}(\truesignal, m) 
    &=
      b_{\iteration}^{2}(\truesignal) + s_m 
    = 
      \sum_{j=m+1}^{n} \langle \truesignal, u_j\rangle^{2} + \delta^{2}\sum_{j=1}^m \lambda_j^{-2},
    \\
    \mathcal{R}_{\designmatrix}( \truesignal, m) 
    &=
    b_{\iteration, \designmatrix}^{2}(\truesignal) + s_{m,\designmatrix} 
    =
    \sum_{j=m+1}^{p} \lambda_j^{2}\langle \truesignal, u_j\rangle^{2}  + \delta^{2}m.\footnote{
      2025-03-10-TODO-BS: Notation muss vereinheitlicht werden \( \langle \cdot, \cdot \rangle \) oder
      \( v^{\top} Y \)?
    }
  \end{aligned}
\end{equation}
As the number $\iteration$ of iterations increases, the number of modes contributing to the strong
bias decreases, while the number of singular values contributing to the strong variance increases.
The tradeoff is similar in prediction risk, with the weak variance linearly scaling in the number of
iterations $m$ and the singular values contributing to the weak bias.

Here, we explicitly have
\begin{align}
          \mathbb{E} \| Y - A \widehat{f}^{(m)} \|^{2} 
  & = 
          \delta^{2} \mathbb{E} \Big( \sum_{j = m + 1}^{n} \varepsilon_{i}^{2} \Big)
        + 
          \delta \mathbb{E} \langle \varepsilon, A (\truesignal - \widehat{f}^{(m)}) \rangle 
        + 
          b_{m, A}^{2}(\truesignal) 
  \\
  & = 
          b_{m, A}^{2}(\truesignal) - s_{m, A} + n \delta^{2},
  \notag
\end{align}
and the discrepancy stopping \( \DP \) from Equation \eqref{eq:discrepancy_stop} mimics the weak
balanced oracle \( m^{\mathfrak{b}}_{w} \) when \( \kappa \) is approximately equal to 
\( n \delta^{2} \).
In fact, \citet[Theorem 3.3]{blanchard2018early} provides the weakly balanced oracle inequality
\begin{equation}
 \label{eq:bal:oracle:truncated:SVD}
           \weakrisk{\truesignal}{\DP} 
  \lesssim
           \weakrisk{\truesignal}{m^\mathfrak{b}_w}
    +
           \sqrt{\samplesize} \noiselevel^2 
    +   
           |\threshold-\samplesize \noiselevel^2|.
\end{equation}
In light of the result in Equation \eqref{eq_OptimalityOfBalancedOracle}, the stopping therefore
behaves optimally up to a constant as long as $|\threshold - \samplesize \noiselevel^{2}| \lesssim
\sqrt{\samplesize}\noiselevel^2$ and 
\( \mathcal{R}_{A}(\truesignal, m^{\mathfrak{b}_{w}}) \le \sqrt{n} \delta^{2} \).
Given that \( \truesignal \) satisfies a  polynomial decay condition, a similar result \citep[Theorem
2.8]{blanchard2018early} can be obtained in strong norm
\begin{equation}
  \begin{aligned}
  \strongrisk{\truesignal}{\DP} 
  \lesssim \strongrisk{\truesignal}{m^\mathfrak{b}_s \lor \sqrt{\samplesize}}
  \end{aligned}
\end{equation}
showing that the stopping rule is optimally adaptive for signals $\truesignal$ such that
$ m^\mathfrak{b}_s \gtrsim \sqrt{\samplesize}$.

When $m^\mathfrak{b}_s$ is of smaller order than $\sqrt{\samplesize}$, the random variability in the
residuals allows for stopping times \( \DP \ge \sqrt{n} \) with non-vanishing probability.
In fact, no sequential stopping rule can attain the optimal risk in this setting; see
\citet[Section 2]{blanchard2018early}.
This problem can be circumvented by using a two-step procedure.
Initially, we compute the usual discrepancy stop $\DP$ defined in Equation \eqref{eq:discrepancy_stop},
based on the residuals. 
Since this stop might have been too late if $m^\mathfrak{b}_s$ is of order $o(\sqrt{\samplesize})$,
in a second step, the discrepancy stop is combined with the Akaike information criterion (AIC) up to
\( \DP \).
This results in
\begin{equation}
\label{eq:two_step_tSVD}
  \tau^{(\text{2-step})} 
  =
  \argmin_{0 \leq m \leq \DP} \text{AIC}(m)
  =
  \argmin_{0 \leq m \leq \DP} 
  \Big( - \sum_{i=1}^m \lambda_i^{-2}\langle Y,v_i \rangle^{2}
        + 2\delta^{2} \sum_{i=1}^m \lambda_i^{-2} 
  \Big), 
\end{equation}
trading off the first data fit term against the second model complexity term, which provides optimal
results also in the settings in which the application of \( \DP \) alone may fail. 
Since we only consider \( \text{AIC}(m) \) for \( m \le \DP \), which depends on the first 
\( \DP \) quantities in the SVD of \( A \), the two-step procedure has the same computational
complexity of \( O(\DP n^{2}) \) as the sequentially stopped algorithm and provides similar
computational advantages.


\subsubsection{Landweber iteration (gradient-descent)}
\label{sssec_LandweberIteration}

The Landweber iteration for solving inverse problems arises from performing gradient descent
updates with learning rate \( \omega < 1 / \| A \|^{2} \) on the least squares problem  
\( \min_{f \in \mathbb{R}^{p}} \| Y - A f \|^{2} \), i.e., 
\begin{align}
  \widehat{f}^{(0)}: = 0, 
  \qquad 
  \widehat{f}^{(m + 1)}: = \widehat{f}^{(m)} + \omega A^{\top} (Y - A \widehat{f}^{(m)}) 
                         = (I - \omega A^{\top} A) \widehat{f}^{(m)} + \omega A^{\top} Y, 
\end{align}
for \( m \geq 0 \).
In terms of an approximating family as in Equation \eqref{eq_ApproximationFunctionGAlpha}, it is
given by 
$ g_{ \alpha }( \lambda ): = \lambda^{-1} ( 1 - ( 1 - \learningrate \lambda  )^{ 1 / \alpha } ) $,
and the discretisation \( \alpha_{\iteration} = \iteration^{-1} \).
If $A^{\top}A$ is invertible, the estimators have the non-recursive representation
\begin{equation*}
    \estimator^{ (\iteration) }
  = 
    \learningrate 
    \sum_{ i = 0 }^{ \iteration - 1 } 
    ( I - \learningrate \designmatrix^{\top} \designmatrix )^{i} \designmatrix^{\top} \response 
  =
    (I-(I-\omega\designmatrix^{\top}\designmatrix)^{m})(\designmatrix^{\top}\designmatrix)^{-1}\designmatrix^{\top}\response, 
  \qquad m \geq 0. 
\end{equation*}
leading to the explicit error decomposition \eqref{eq:strong_risk} and \eqref{eq:weak_risk} of the strong and weak risk: 
\begin{equation}
\label{eq:bias_variance_landweber}
\begin{aligned}
\mathcal{R}(\estimator^{(\iteration)}, \truesignal) &=b_\iteration^{2}(\truesignal)+s_\iteration =
  \Vert (I-\omega
  \designmatrix^{\top}\designmatrix)^{m}\truesignal\Vert^{2}+\noiselevel^{2}\mathrm{tr}((\designmatrix^{\top}\designmatrix)^{-1}(I-(I-\omega\designmatrix^{\top}\designmatrix)^{\iteration})^{2}),\\
\mathcal{R}_{\designmatrix} (\estimator^{(\iteration)}, \truesignal) &= b_{\iteration,
  \designmatrix}^{2}(\truesignal) + s_{\iteration,\designmatrix} = \Vert \designmatrix(I-\omega
  \designmatrix^{\top}\designmatrix)^{\iteration}\truesignal \Vert^{2} +
  \delta^{2}\mathrm{tr}((I-(I-\omega\designmatrix^{\top}\designmatrix)^{\iteration})^{2}).
\end{aligned}
\end{equation}

\begin{remark}[Numerical implementation]
In the \package, the weak and strong biases are updated iteratively through
\begin{align*}
b_\iteration^{2}(\truesignal) = \Vert (I-\omega\designmatrix^{\top}\designmatrix)(\truesignal-\mathbb{E}\estimator^{(\iteration-1)}) \Vert^{2}, \quad 
b_{\iteration, \designmatrix}^{2}(\truesignal) =
\Vert \designmatrix(I-\omega\designmatrix^{\top}\designmatrix)(\truesignal-\mathbb{E}\estimator^{(\iteration-1)}) \Vert^{2},
\end{align*}
while the weak and strong variance can only be updated sequentially by computing the powers of the
  matrix $I-\omega \designmatrix^{\top}\designmatrix$.
The weak and strong variance can then be evaluated via
\begin{equation*}
\begin{aligned}
  s_m 
  =
  \noiselevel^{2} \omega^{2}\mathrm{tr}\Big( \Big(\sum_{i=1}^{m-1}(I-\omega
  \designmatrix^{\top}\designmatrix)^{i}\Big)^{2} A^{\top}A\Big),
  \quad
  s_{m,A} 
  =
  \noiselevel^{2}\omega^{2}\mathrm{tr}\Big( \Big(A\sum_{i=1}^{m-1}(I-\omega
  \designmatrix^{\top}\designmatrix)^{i}A^{\top}\Big)^{2} \Big),
\end{aligned}
\end{equation*}
which is even possible if $A$ is not injective.
\end{remark}


When, \( A^{\top} A \) is non-singular, simple calculations show that
\begin{align}
  \mathbb{E} \| Y - A \widehat{f}^{(m)} \|^{2} 
  & =
  b_{m, A}^{2}(f) + \delta^{2} \tr (I - \omega A^{\top} A )^{2 m},
\end{align}
which is equal to \( b_{m, A}^{2}(f) - s_{m, A} + \delta^{2} n \) up to an additional mixture term.
As in the case of the truncated SVD estimators, the discrepancy stopping time is closely related to
the weak balanced oracle \( m^{\mathfrak{b}}_{w} \).
\citet[Corollary 3.6]{blanchard2018optimal} derive an oracle inequality of the form 
\begin{equation}
  \label{eq_OracleInequalityGeneralRegularizationMethods}
  \strongrisk{\truesignal}{\DP} 
  \lesssim
  \max \left({\strongrisk{\truesignal}{m^\mathfrak{b}_s}, (m^\mathfrak{b}_w)^{2}
  \weakrisk{\truesignal}{m^\mathfrak{b}_w}}\right),
\end{equation}
which holds even for general spectral regularisation methods as introduced in
\eqref{eq:spectral_regularisation_methods} given signals with polynomially decaying singular values
$\lambda_i=i^{-1/\nu}$.
For the Landweber iteration, the authors also identify a class of signals for which 
Equation \eqref{eq_OracleInequalityGeneralRegularizationMethods} translates to an optimality result
of the form $ \strongrisk{\truesignal}{\DP} \lesssim
\strongrisk{\truesignal}{m^\mathfrak{b}_s}$.


\subsubsection{Conjugate gradient descent}
\label{sssec_ConjugateGradientDescent}

In this section, we consider conjugate gradient descent for the normal equation
\begin{align}
  \label{eq_NormalEquation}
  A^{\top} Y = A^{\top} A f
\end{align}
as stylised in \Cref{alg:conjugate_gradient_descent}. 
In the \package{}, the algorithm is computed in a more efficient manner as proposed in
\citet[Algorithm~7.4.1]{bjoerck1996}.
\begin{center}
\begin{minipage}{0.9\linewidth}
\begin{algorithm}[H]
\caption{Conjugate gradient descent for the normal equation}
\label{alg:conjugate_gradient_descent}
\begin{algorithmic}[1]
\State \( 
               \estimator^{(0)} \gets 0, \quad \mathbf{r}^{(0)}\gets\response-\designmatrix \estimator^{(0)}, \quad q^{(0)} \gets\designmatrix^T \mathbf{r}^{(0)}
\)
\For{ \( m = 0, 1, 2, \dots \) }
\State \(
    \alpha_\iteration \gets\|\designmatrix^{\top} \mathbf{r}^{(\iteration)}\|^2 /\|\designmatrix q^{(\iteration)}\|^2 \quad (\text{learning rate})
\)
\State \( 
\estimator^{(\iteration+1)}\gets\estimator^{(\iteration)}+\alpha_{\iteration} q^{(\iteration)}
\)
\State  \(
d^{(\iteration+1)} \gets \mathbf{r}^{(\iteration)}-\alpha_\iteration \designmatrix q^{(\iteration)} \quad \text{(residual vector)}
\)
\State \(
  q^{(\iteration+1)} \gets \designmatrix^{\top}\mathbf{r}^{(\iteration+1)}+
  \frac{\|\designmatrix^{\top} \mathbf{r}^{(\iteration+1)}\|^2}{\|\designmatrix^T \mathbf{r}^{(\iteration)}\|^2}   q^{(\iteration)} \quad \text{(search direction)}
\)
\EndFor
\end{algorithmic}
\end{algorithm}
\end{minipage}
\end{center}

The conjugate gradient algorithm does not have a direct representation in terms of a simple
approximating family \( (g_{\alpha})_{\alpha > 0} \).
Nevertheless, a similar representation based on the residual polynomials is presented in
\citet{hucker2025comparingregularisationpathsconjugate}.

We present the approach to early stopping for conjugate gradients following \citet{hucker2024early}. 
Under suitable assumptions on the design matrix $\designmatrix$ and the observation $\response$,
there is another formulation of the conjugate gradient algorithm obtained through minimising
residual polynomials, which we use for the theoretical analysis. 
Recall the convention that $g(\designmatrix \designmatrix^{\top})=\sum_{j=1}^n g(\lambda_j^{2}) v_j$ 
for a function \( g \) in terms of the SVD of $\designmatrix$ in \eqref{eq:SVD:designmatrix}.
The residual polynomial $r_\iteration$ with $\iteration \in \{0,\dots,n \}$ is then defined as
\begin{equation*}
r_{\iteration} \coloneqq \argmin_{p_{\iteration}} \Vert p_{\iteration}(\designmatrix \designmatrix^{\top}) \response\Vert^{2},
\end{equation*}
where the argmin is taken over all polynomials $p_{\iteration}$ of degree at most $\iteration$ satisfying $p_{\iteration}(0)=1$. The conjugate gradient estimator can now be expressed as
\begin{equation*}
\estimator^{(\iteration)} \coloneqq \designmatrix^{+}(I_{\samplesize}- r_{\iteration}(\designmatrix \designmatrix^{\top}))\response
\end{equation*}
with $A^+$ denoting the Moore--Penrose pseudoinverse of $A$. 
For the precise definition, we refer to \citet*[Definition 3.1]{hucker2024early}. For $t=\iteration
+\alpha$, with $\iteration \in \{0,\dots,n-1 \}$, $\alpha \in (0,1]$, the residual polynomial may be interpolated through $r_t \coloneqq (1-\alpha)r_{\iteration} + \alpha r_{\iteration +1}$, giving rise to the interpolated CG estimator $\estimator^{(t)} = (1-\alpha)\estimator^{(\iteration)} + \alpha \estimator^{(\iteration +1)}$.
\begin{remark}
\label{remark:interpolation}
To avoid the effect of \textit{overshooting} (see \citet{hucker2024early, miftachov2025early}), we interpolate linearly between the residual polynomials. In the \package{}, the default is set to the non-interpolated estimator and the interpolated estimator is also available as an option. From a computational perspective, the interpolation comes at a negligible additional cost since the required quantities are calculated in either case.
\end{remark}
Let $x_{1,t}$ be the smallest zero on $[0,\infty)$ of the $t$-th residual polynomial $r_t$ and denote by $r_{t,<}(x) \coloneqq r_t(x) \mathbf{1}(x<x_{1,t})$ the polynomial up to the first zero. 
Based on the residual polynomial, it is now possible to find a weak risk bound consisting of
an approximation and a stochastic error, which are data-dependent. 
The bound is given by
\begin{equation}
\label{eq:upperbound_error_decomposition_cg}
  \mathcal{R}_{A}(f^{*}, m): 
  = 
  \Vert \designmatrix (\estimator^{(t)} - \truesignal) \Vert^{2} 
  \leq
  2 (b_{t, \designmatrix} + s_{t, \designmatrix})
\end{equation}
with
\begin{align*}
  b_{t,A} 
  &\coloneqq
  \Vert r_{t,<}^{1/2}(\designmatrix \designmatrix^{\top})\designmatrix \truesignal \Vert^2 
  +
  \| Y - A \widehat{f}^{(t)} \|^{2}
  -
  \Vert r_{t,<}^{1/2}(\designmatrix \designmatrix^\top)\response \Vert^2, \\
\qquad s_{t,A} 
  &\coloneqq
  \Vert (1-r_{t,<})^{1/2}(\designmatrix \designmatrix^\top)\noise \Vert^2.
\end{align*}
Note that as $b_{t,A}$ does not correspond exactly to the squared bias, it also does not need to be positive. The (data-dependent) balanced oracle is given by
\begin{equation*}
t_w^{\mathfrak{b}} \coloneqq \inf \{t \in [0, d] \colon b_{t,A} \leq s_{t,A}\}.
\end{equation*}
In contrast to the other algorithms, even with knowledge of the true signal, the balanced oracle is
not available to the user since it depends on the residual polynomial, which is very hard to compute
in practice. 
In the \package{}, we instead compute an empirical version of the weak classical oracle
\begin{equation*}
  t^{\mathfrak{o}} 
  =
  t^{\mathfrak{o}}(f^{*}) 
  =
  \argmin_{t \in [0,d]} \Vert\designmatrix (\estimator^{(t)} - \truesignal)\Vert.
\end{equation*}
Analogously to before, we consider the discrepancy stopping time 
\begin{equation*}
  \DP \coloneqq \inf\{ t \in [0,d] \colon \| Y - A \widehat{f}^{(t)} \|^{2} \leq \kappa\}.
\end{equation*}
Even though we do not have a classical bias-variance decomposition, the prediction risk of the discrepancy stop satisfies the weakly balanced oracle inequality \cite[Theorem~6.8]{hucker2024early}
\begin{equation}
\mathcal{R}_{\designmatrix}(\truesignal, \DP) \lesssim \mathbb{E}(s_{t_w^{\mathfrak{b}}, A}) + \sqrt{n}\delta^2 + |\kappa - n\delta^{2}|. \label{eq:bal:oracle:CG}
\end{equation}
Note that the right-hand side involves the expected stochastic error at the balanced oracle and not the risk $\mathcal{R}_{\designmatrix}(\truesignal, t_w^{\mathfrak{b}})$ as in \eqref{eq:bal:oracle:truncated:SVD}, since \eqref{eq:upperbound_error_decomposition_cg} only provides an upper bound. Nevertheless, it is possible to retrieve the optimal rate for $\mathcal{R}_{\designmatrix}(\truesignal, \DP)$ by controlling $\mathbb{E}(s_{t_w^{\mathfrak{b}}, A})$, see \citet[Corollary~6.9]{hucker2024early}. The result shows that the critical value $\threshold$ should be chosen such that $|\kappa - n\delta^{2}| \lesssim \sqrt{n}\delta^{2}$ as for truncated SVD. 
For the reconstruction error $\mathcal{R}((\truesignal)^+, \DP)$, where $(\truesignal)^+ = A^+ A \truesignal$, there is no immediate analogue to the oracle inequality \eqref{eq:bal:oracle:CG}. 
Instead, the reconstruction error at $\DP$ has a relatively rough bound involving the error terms at $\DP$ and the intrinsic regularising quantity $\abs{r_{\DP}'(0)}$. 
Still, under a source condition on $\truesignal$ and polynomial spectral decay of order $p$, an optimal rate for $\mathcal{R}((\truesignal)^+, \DP)$ can be shown over a given regime of regularity parameters, see \citet[Theorem~7.10]{hucker2024early}.


\subsection{Boosting in high-dimensional linear models}
\label{ssec_BoostingInHighDimensionalLinearModels}

Following the approach in \citet{stankewitz2024early}, \citet{Ing2020ModelSelection} and
\citet{kueck2023estimation}, we consider sequential early stopping for an iterative boosting
algorithm applied to data from a high-dimensional linear model
\begin{equation}
  \label{eq_IV_1_HDLinearModel}
  Y_{i} =   f^{*}( X_{i} ) + \varepsilon_{i} 
        =   \sum_{ j = 1 }^{p} \beta_{j}^{*} X_{i}^{ (j) } 
          +
            \varepsilon_{i}, 
  \qquad i = 1, \dots, n,
\end{equation}
where \( 
  f^{*}(x) = \sum_{ j = 1 }^{p} \beta^{*}_{j} x^{ (j) }, 
             x \in \mathbb{R}^{p}
\), is a linear function of the columns of the design matrix, \( 
  \varepsilon: = ( \varepsilon_{i} )_{ i \le n }
\) is the vector of centered noise terms in our observations, and the parameter
size \( p \) is potentially much larger than the sample size \( n \). 
In order to consistently estimate \( f^{*} \) in this setting, the problem has to be regularised,
either explicitly as in the Lasso or implicitly by suitably iterating an iterative procedure.

As an estimation algorithm, we focus on \( L^{2} \)-boosting based on orthogonal
matching pursuit (OMP), which produces an estimate of the true signal \( f^{*} \) and performs
variable selection at the same time.
Empirical correlations between data vectors are measured via the \emph{empirical
inner product} \( \langle a, b \rangle_{n}: = n^{-1} \sum_{ i = 1 }^{n} a_{i} b_{i} \) with norm 
\( \| a \|_{n}: = \langle a, a \rangle_{n}^{ 1 / 2 } \), for \( a, b \in \mathbb{R}^{n} \).
By \( \widehat{ \Pi }_{J}: \mathbb{R}^{n} \to \mathbb{R}^{n}, \) we denote the orthogonal projection
with respect to \( \langle \cdot, \cdot \rangle_{n} \) onto the span of the columns \( \{ X^{ (j) }:
j \in J \} \) of the design matrix.
OMP is initialised at \( \estimator^{ (0) }: = 0 \) and then iteratively
selects the covariates \( X^{ (j) }, j \le p \), which maximise the empirical
correlation with the residuals \( Y - \estimator^{ (m) } \) at the current
iteration \( m \).  
The estimator is updated by projecting onto the subspace spanned by the selected
covariates.
Explicitly, the procedure is given by the following algorithm:

\begin{center}
\begin{minipage}{0.9\linewidth}
  \begin{algorithm}[H]
    \caption{Orthogonal matching pursuit (OMP)}
    \label{alg_IV_OMP}
    \begin{algorithmic}[1]
      \vspace{2px}
      \State \( 
               \estimator^{ (0) } \gets 0,
               \widehat{J}_{0} \gets \emptyset 
             \) 
      \vspace{2px}
      \For{ \( m = 0, 1, 2, \dots \) }
        \State \( 
                        \widehat{j}_{m + 1}
                 \gets 
                        \argmax_{ j \le p } 
                        \Big| 
                          \Big\langle 
                            Y - \estimator^{ (m) }, 
                            \frac{ X^{ (j) } }
                                 { \| X^{ (j) } \|_{n} }
                          \Big\rangle_{n}
                        \Big| 
              \) 
        \vspace{2px}
        \State \( 
                       \widehat{J}_{ m + 1 } 
                 \gets
                       \widehat{J}_{m} \cup \big\{ \widehat{j}_{ m + 1 } \big\}
               \)
        \vspace{2px}
        \State \( 
                        \estimator^{ ( m + 1 ) }
                  \gets
                        \widehat{ \Pi }_{ \widehat{J}_{ m + 1 } } Y
               \) 
        \vspace{2px}
        \vspace{2px}
      \EndFor
      \vspace{2px}
  \end{algorithmic}
\end{algorithm}
\end{minipage}
\end{center}

Other versions of \( L^{2} \)-boosting algorithms exist in this setting, where the
projection step 5 of Algorithm \ref{alg_IV_OMP} is replaced with a greedy gradient step in the direction of the selected covariate,
see, e.g., \citet{Buehlmann2006BoostingForHDLinearModels} or \citet{kueck2023estimation}.
Here, however, we focus on the OMP version for the algorithm as it is the only one for which
theoretical early stopping guarantees exist.
Due to the orthogonality of the projections, the empirical squared error of the estimation can be decomposed into an empirical bias and a stochastic error part
\begin{align}
  \label{eq_IV_EmipircalBiasVarianceDecomposition}
      \mathcal{R}(f^{*}, m): 
  & = 
      \| \estimator^{ (m) } - f^{*} \|_{n}^{2} 
    = 
      \| ( I - \widehat{ \Pi }_{m} ) f^{*} \|_{n}^{2} + \| \widehat{ \Pi }_{m} \varepsilon \|_{n}^{2} 
    = b_{m}^{2}                                       + s_{m}, 
\end{align}
matching the structure of the general error bound in \eqref{eq:general_error_decomposition}.
The squared bias \( m \mapsto b_{m}^{2} \) is monotonously decreasing and the stochastic error 
\( m \mapsto s_{m} \) is monotonously increasing in \( m \).
Both of the quantities, however, remain random because of the randomness of the variable selection
in Algorithm \ref{alg_IV_OMP}.
In the context of this model and algorithm, we consider two approaches for early stopping.

As in Section \ref{sec_EarlyStoppingForInverseProblems}, \citet{stankewitz2024early} considers a
discrepancy-type stopping rule, i.e. halting the algorithm at
\begin{equation}
\label{eq:discrepancy_principle_boosting}
  \tau^{\text{DP}}: = \inf \{ m \geq 0: \| Y - \estimator^{(m)} \|_{n}^{2} \le \kappa \}
\end{equation}
for some user-specified critical value \( \kappa > 0 \). 
By decomposing the residuals 
\begin{equation}
  \label{eq_DecompositionOfResiduals}
        \| Y - \estimator^{(m)} \|_{n}^{2} 
   =   
        b_{m}^{2} + 2 \langle (I - \widehat{\Pi}_{m}) f^{*}, \varepsilon \rangle_{n}
      + 
        \| \varepsilon \|_{n}^{2} - s_{m},
\end{equation}
we obtain that the stopping condition \( \| Y - \estimator^{(m)} \|_{n}^{2} \le \kappa \) is
equivalent to 
\begin{align}
      b_{m}^{2} + 2 \langle (I - \widehat{\Pi}_{m}) f^{*}, \varepsilon \rangle_{n} 
  \le 
      s_{m} + \kappa - \| \varepsilon \|_{n}^{2}.
\end{align}
Assuming that the cross term \( \langle (I - \widehat{\Pi}_{m}) f^{*}, \varepsilon \rangle_{n} \) is
negligible and we can choose the critical value \( \kappa \) as a suitable estimator 
\( \widehat{\sigma}^{2} \) of the squared empirical norm \( \| \varepsilon \|_{n}^{2} \) of the
error terms, the stopping time \( \tau^{\text{DP} } \) mimics the balanced oracle 
\begin{equation}
  m^{\mathfrak{b}} = m^{\mathfrak{b}}(f^{*})
   = 
  \inf \{ m \geq 0: b_{m}^{2} \le s_{m} \}.
\end{equation}
Under some additional assumptions, this intuition can be made mathematically rigorous.
Indeed, when the error terms are i.i.d. centered Gaussians with unknown variance 
\( \sigma^{2} \), fixing the critical value \( \kappa = \kappa_{m} = \widehat{\sigma}^{2} + 8
\widehat{\sigma}^{2} m \log n \) yields the 
general oracle inequality
\begin{equation}
  \label{eq_OracleInequality_L2-Boosting}
          \mathcal{R}(f^{*}, \DP)  
  \lesssim \mathcal{R}(f^{*}, m^{ \mathfrak{b}})
        + 
          \frac{\sigma^{2} m^{ \mathfrak{b} } \log p }{n}
        + 
          | \widehat{ \sigma }^{2} - \| \varepsilon \|_{n}^{2} |
\end{equation}
with probability converging to one for \( n \to \infty \).
The first term in Equation \eqref{eq_OracleInequality_L2-Boosting} is of optimal order.
In the classical sparse setting, when the number of non-zero coefficients \( \beta_{j}^{*} \) in
model \eqref{eq_IV_1_HDLinearModel} is given by an unknown \( s \in \mathbb{N} \), the same is true
for the second term.
Then, the oracle iteration \( m^{\mathfrak{b}} \) will be of order \( s \).
At the same time, \( f^{*} \) cannot be estimated with a faster rate than 
\( s \log(p) / n \), see \citet{RaskuttiEtal2009_MMRSparseLinearModel}.
Since the empirical noise level \( \| \varepsilon \|_{n}^{2} \) can also be approximated with this
rate by an estimator \( \widehat{\sigma}^{2} \), early stopping is fully adaptive to the sparsity 
\( s \) in this setting.

In \citet{kueck2023estimation}, the authors consider a sequential stopping rule which differs
from the discrepancy principle.
They propose stopping at the residual ratio stopping time
\begin{align}
  \label{eq_ResidualRationStop}
  \tau^{\text{RR}} : = 
  \inf \Big\{ m \geq 0: \frac{\| Y - \estimator^{(m + 1)} \|_{n}^{2}}
                            {\| Y - \estimator^{(m    )} \|_{n}^{2}}
                       \ge 
                            1 - \frac{4 K_{\text{RR}} \log(2 p / \alpha)}{n}
       \Big\},
\end{align}
where \( K_{\text{RR}} \) is a user-specified parameter and \( \alpha \) is a selected confidence
level.
From the perspective of the strongly \( s \)-sparse setting, we can derive a solid intuition for the
mechanics of this stopping rule.
Indeed, rearranging the condition in  \eqref{eq_ResidualRationStop}, we stop at
the first index \( m \) at which
\begin{equation}
  \label{eq_RearrangedResidualRationStop}
  \| Y - \estimator^{(m    )} \|_{n}^{2} 
  - 
  \| Y - \estimator^{(m + 1)} \|_{n}^{2} 
  \le 
  \| Y - \estimator^{(m)} \|_{n}^{2} 
  \frac{4 K_{\text{RR}} \log(2 p / \alpha)}{n}.
\end{equation}
As long as the stochastic error \( s_{m} \) is negligible compared to the bias, the left-hand side 
in Equation \eqref{eq_RearrangedResidualRationStop} reflects the reduction
\( b_{m}^{2} - b_{m + 1}^{2} \) of the squared bias.
Recalling that the minimax rate of convergence is \( \sigma^{2} s \log(p) / n \) and with each
additional iteration, we approximately incur an increase of the stochastic error of 
\( \sigma^{2} \log(p) / n \), any iteration with bias reduction larger than this quantity is
still beneficial.
Around the optimal iteration \( m \), the residuals \( \| Y - \estimator^{(m)} \|_{n}^{2} \) will
be close to \( \| \varepsilon \|_{n}^{2} \), which itself is an approximation of \( \sigma^{2} \).
Then, the right-hand side in Condition \eqref{eq_RearrangedResidualRationStop} is approximately of 
size \( \sigma^{2} \log(p) / n \), and the residual ratio stopping rule prescribes stopping at the
first index at which the reduction in the squared bias becomes smaller than the increase of the
stochastic error, i.e. it mimics a first-order condition of the risk.
Cleverly, by considering the residual ratio instead of the residual difference, the authors generate
the term \( \| Y - \estimator^{(m)} \|_{n}^{2} \) on the right-hand side of Equation
\eqref{eq_RearrangedResidualRationStop}, which avoids having to estimate 
\( \| \varepsilon \|_{n}^{2}\) as for the discrepancy principle.

As a theoretical guarantee, they obtain that for a constant \( C > 0 \) with probability at least 
\( 1 - \alpha \), 
\begin{equation}
  \mathcal{R}(f^{*},\tau^{\text{RR}} )
  \lesssim 
  \frac{\sigma^{2} s \log(p \lor n) \log n}{n},
\end{equation}
which guarantees optimal recovery of the signal \( f^{*} \) up to a \( \log \)-term.

Both the discrepancy and the residual ratio stopping rule can be sensitive to the noise estimation
and the choice of the constant \( K_{\text{RR}} \) respectively.
In practice, relying on the estimator \( \estimator^{(\tau)} \) with \( \tau = \tau^{(\text{DP})} \)
or \( \tau = \tau^{(\text{RR})} \) directly retains some unwanted variance.
Therefore, it can be beneficial to combine early stopping with a high-dimensional Akaike model
selection criterion applied to the path up to the stopping time and consider the two-step procedure
\begin{align}
  \label{eq_HDAIC}
  \tau^{(\text{2-step})}: = \argmin_{m \le \tau} \text{AIC}(m) 
  \qquad \text{with} \qquad 
  \text{AIC}(m):
  =
  \| Y - A \widehat{f}^{(m)} \|^{2} + \frac{ K_{ \text{AIC} }  \widehat{\sigma}^{2} m \log p }{n}, 
  \quad m \geq 0
\end{align}
with \( \tau \) again equal to either \( \tau^{\text{DP}} \) or \( \tau^{\text{RR}} \).
Applying the AIC criterion to the full path yields optimal recovery of \( f^{*} \), see, e.g., 
\citet{stankewitz2024early}.
Clearly, the same is true as long as \( \tau \) in Equation \eqref{eq_HDAIC} is large enough.
This provides some clear direction for the use in applications:
\begin{enumerate}
  \item Compute the estimators up to the early stopping index \( \tau \), where any parameters in
    the algorithm may be tuned to slightly favour larger stopping times.

  \item Instead of using the stopping time \( \tau \) directly, use the iteration 
    \( \tau^{(\text{2-step})} \) from the two-step procedure, which considers the whole path up to
    the stopping time.
\end{enumerate}
For the minor additional cost of introducing a slight upward bias in the stopping times
\( \text{AIC}(m) \) for \( m \le \tau \), we can therefore stabilise the procedures by reducing the
variability stemming from tuning constants while maintaining the computational advantage from early
stopping.


\subsection{Regression trees}
\label{ssec_RegressionTrees}

The classification and regression tree (CART) by \citet{breiman1984classification} is a learning algorithm
that constructs estimates of an unknown function by iteratively partitioning its domain.
We focus on a non-parametric regression setting, in which we observe independent, identically
distributed pairs \( (X_{1}, Y_{1}), \dots, (X_{n}, Y_{n}) \in \mathbb{R}^{p} \times \mathbb{R} \)
according to the data-generating process 
\begin{align}
  \label{EqRegr}
  Y_i = \truesignal(X_i) + \noise_i, \quad i&=1,\ldots,n,  
  \qquad \text{ with } \qquad 
  \mathbb{E}(\noise_i\,|\,X_i)=0 \text{ and }  \text{Var}(\noise_i)=\sigma^2.
\end{align}

In order to obtain an estimator of the unknown regression function \( f^{*} \), the CART algorithm
starts with a parent node $A=\mathbb{R}^{\dimension}$, which is recursively partitioned into
non-overlapping $\dimension$-dimensional (hyper-)rectangles. 
The partitioning begins by selecting a coordinate $j=1,\ldots,\dimension$, and a threshold
$c\in\mathbb{R}$, dividing $A=\mathbb{R}^{\dimension}$ into the left child node $A_L (j,c) = \{x \in
A : x_j < c\}$ and the right child node $A_R (j,c) = \{x \in A : x_j \geq c\}$. 
The child nodes then serve as parent nodes for subsequent iterations, continuing the recursive
partitioning process. 
At each instance, the {\it terminal nodes} (nodes without children) define a {\it partition} of
$\mathbb{R}^{\dimension}$, which is refined as the tree grows. 
Following \citet{miftachov2025early}, the regression tree is grown based on the breadth-first search
principle, resulting in the simultaneous splitting of all terminal nodes that contain more than one
observation \( X_{i} \), \( i \le n \), when iterating from one level to the next level.
At each split, the coordinates \( j^{*} \) and the threshold \( c^{*} \) for a generic parent node 
\( A \) are chosen by greedily minimising the residuals (or in-sample training errors) of the child
nodes according to
\begin{equation*}
  (j^*, c^*) \in \underset{(j,c)}{\operatorname{argmin}} \left(\sum_{i:X_i \in A_{L}(j,c)}\left(Y_i-\bar{Y}_{A_{L}(j,c)}\right)^2 + \sum_{i:X_i \in A_{R}(j,c)}\left(Y_i-\bar{Y}_{A_{R}(j,c)}\right)^2\right)
\end{equation*}
with $\bar{Y}_{A} = \frac{1}{n_A} \sum_{i:X_i \in A} Y_i$ being the node average with local sample size $n_A=\lvert \{i: X_i \in A \} \rvert$.

At iteration (or level) $ m \geq 0 $ of the CART algorithm, this procedure yields a
partition $P_m=\{A^{m}_1,\ldots,A^{m}_K\}$ of $\mathbb{R}^{\dimension}$ such that the $A^{m}_k$
contain at least one design point $X_i$.
Associated to the  partition is an orthogonal projection $\Pi_{P_m}:L_n^2 \rightarrow L_n^2$
\begin{equation}
          \Pi_{m} f(x):
  =
          \frac{1}{n^m_k} \sum_{i: X_i \in A^m_k} f\left(X_i\right) 
  \qquad
          \text {for } x \in A^m_k, \label{eq:orth_projection}
\end{equation}
which is the average of $f(X_i)$ for the $X_i$ in the set $A^m_k$ at iteration $m$ which contains
$x$.
By slightly overloading this notation, we can formulate the CART-estimator at iteration \( m \) as
\begin{equation}
  \widehat{f}^{(m)}(x): 
  = 
  \Pi_{m} Y:
  = 
  \frac{1}{n_k^m} \sum_{i:X_i \in A_k^m} Y_i \qquad \text{for}\ x\in A^m_k.
\end{equation}

Common approaches to determine a suitable level of the regression tree are post-pruning
\citep{breiman1984classification}, stopping when a pre-specified depth is reached
\citep{klusowski2023large}, or growing the tree until its maximal depth
\citep{scornet2015consistency}. 
In practice, these techniques require cross-validation to get the optimal tree depth, which is not
desired due to the computational costs. 
By utilising the stopping rule based on the discrepancy principle as introduced in the previous
chapters, we avoid cross-validation for the regression tree while being computationally efficient
and interpretable. 
Oracle inequalities for the early-stopped regression tree are established in
\citet{miftachov2025early}, with the main result based on the CART algorithm summarised below.

Similar to the linear interpolation described in \Cref{remark:interpolation}, we interpolate the
regression tree estimator between two consecutive iterations to avoid the issue of
\textit{overshooting}. 
The continuous parameter allows to balance between over- and underfitting more granularly for
regression trees and avoids additional discretisation errors in the analysis. 
Note that \citet{miftachov2025early} introduce a \textit{generalised projection flow}, which has a
much broader scope. 
For example, it includes gradient descent, ridge regression, smoothing splines, and other
estimators. 
In this work, however, we solely focus on the generalised projection 
$\left(\Pi_t\right)_{t \in[0, n]}$ flow applied to the regression tree estimators, which defines the
linear interpolation between two consecutive orthogonal projections at iteration $m$ and $m+1$ as
\begin{equation}
      \Pi_{t}:=(1-\alpha) \Pi_{P_m}+\alpha \Pi_{P_{m+1}}, \label{eq:proj_flow}
  \qquad \text{ for } t = m + \alpha \quad\text{and} \quad \alpha \in[0,1]. 
\end{equation}
The associated interpolated regression tree estimator is \( \estimator_t= \Pi_t Y \). 
The interpolation comes at no additional computational cost since the required quantities are
calculated in either case. In our Python implementation, we include both estimators, one is based on
the orthogonal projection and the other on the projection flow.

As in Equation \eqref{eq:general_error_decomposition}, the risk of this family of estimators can be
decomposed into an approximation error term and a stochastic error term
$$
\mathcal{R}(\truesignal, t) := \| \estimator^{ (t) } - f^{*} \|_{n}^{2}  \lesssim \left\|\left(I-\Pi_t\right) \truesignal\right\|_n^2+\left\|\Pi_t \varepsilon\right\|_n^2=: b_t + s_t,
$$
where $t \mapsto b_t$ decreases continuously from $\norm{f}_n^2$ to 0 and $t \mapsto s_t$ increases continuously from 0 to $\norm{\varepsilon}_n^2$.
The (random) balanced oracle $t^\mathfrak{b}_w$ is given by 
\[ t^\mathfrak{b}_w=\inf\{t\in[0,n]\,|\, b_t  \le s_t \}\]
and for a threshold $\kappa>0$, as before, a data-driven discrepancy-type stopping rule $\tau$ is
given by
\[ \DP=\inf\{t\in[0,n]\,|\, \norm{Y-\estimator_t}_n^2 \le\kappa\}.\]
Assuming that the noise vector $\noise$ is Gaussian with unknown variance $\sigma^2$,
Theorem 6.16 in \citet{miftachov2025early} provides the following oracle-type inequality for
the risk at the interpolated early stopping estimator under the original CART algorithm
\begin{equation}
\mathbb{E}\mathcal{R}(\truesignal, \DP)  
\lesssim \mathbb{E} \Big(\inf_{t\in[0,n]}\Big( b_t^2 + s_t \Big)\Big)+ \mathbb{E}\abs{\threshold-\norm{\noise}_n^2} + \frac{\sigma^2 \log((\dimension \vee 2)n)}{n}\mathbb{E}(t^\mathfrak{b}_w+1).
\end{equation}
The third term, which we refer to as the cross term, evolves due to the complex dependence between $\Pi_t, \noise $, and $t^\mathfrak{b}_w$; see Proposition 4.3 in \citet{miftachov2025early}. At iteration t, there are $\mathcal{O}\left((p n)^{t}\right)$ possible splits, which contribute to the cross term.
The early-stopping error $\mathbb{E}\abs{\kappa-\norm{\noise}_n^2}$ is usually of order $\sigma^2n^{-1/2}$ or larger and thus dominates the last term in this bound. Typically, the first term, which is the oracle error, has an additional $\log(pn)$-factor to the best achievable minimax rate $n^{-2/(d+2)}$ over classical function spaces. Then, the early stopped regression tree adapts to the oracle since $n^{-2/(p+2)}\ge n^{-1/2}$ already for $p\ge 2$ and $f$ Lipschitz.


\section{Implementation}
\label{sec:implementation}
This section illustrates an example code explaining how the \package{} can be used and replicates simulation results from \citet{blanchard2018early, blanchard2018optimal}, \citet{miftachov2025early}, \citet{stankewitz2024early} and \citet{hucker2024early}. To install the package, please follow the instructions in the \href{https://earlystop.github.io/EarlyStopping/}{documentation}.

Each of the iterative estimation procedures introduced in \Cref{sec_BackgroundOnEarlyStopping} has
its own class, stylised by \Cref{code:example_class}, named \pythoninline{TruncatedSVD},
\pythoninline{Landweber}, \pythoninline{ConjugateGradients}, \pythoninline{L2_boost} and
\pythoninline{RegressionTree}. Each instance of one of these classes requires the specification of a
design-matrix \pythoninline{design} and a response variable \pythoninline{response} so that the iterative estimation procedure may be applied. Some quantities, such as the bias, risk and balanced oracle, depend on the unknown \pythoninline{true_signal} and are called oracle quantities. The user can additionally specify the optional parameters \pythoninline{true_signal} and \pythoninline{true_noise_level} to allow all oracle quantities to be tracked while the iterative estimation procedure is executed. Alongside the parameters shared among classes, algorithm-specific parameters, such as the learning rate for the \pythoninline{Landweber}-class, can additionally be specified.

\begin{python}[caption={Exemplary class.},label=code:example_class]
class Algorithm:
   %\texttt{{\#} Class attributes}%
   self.sample_size %---- Sample size $\samplesize$.%
   self.parameter_size %---- Dimension $\dimension$.%
   self.design %---- Design matrix $A$.%
   self.response %---- Response $Y$.%
   
   self.iteration %---- Current maximal achieved iteration $\iteration$.%
   self.algorithm_estimate_list %---- All computed estimators up to self.iteration.%
   self.residuals %---- Squared residuals between the observed data and the estimator.%
   
   %\texttt{{\#} Methods}%
   def __init__(self,design,response,true_signal=None,true_noise_level=None)
     %Collect model parameters and initialise computational quantities.%
     
   def iterate(self, number_of_iterations)
     %Execute a specified number of iterations of the iterative estimation procedure.%
     
   def get_estimate(self, iteration)
     %Return the estimator at a specified iteration.%
     
   def get_discrepancy_stop(self, critical_value, max_iteration)
     %Iterate to/return the discrepancy stop \eqref{eq:discrepancy_stop}%.
     
   def get_balanced_oracle(self)
      %Iterate to/return the weak balanced oracle \eqref{eq:balanced_oracles_weak_strong}%.
\end{python}

The \pythoninline{iterate}-method exists within all of the classes and executes a specified number of iterations of the iterative estimation procedure.  All other methods are of the form \pythoninline{get_quantity} where quantity is the name of the quantity that should be returned. In particular, \pythoninline{get_quantity} will perform iterations of the iterative estimation procedure until either the desired quantity can be returned or a maximal iteration is reached. The computations are performed using \citetalias{numpy} and \citetalias{scipy}.

All classes are built around the \textit{iterate-once} philosophy and keep track of the maximal achieved iteration. If the execution of a particular method does not require iterating any further, the output will be based on the computations that were already made previously. On the other hand, if the number of iterations may be data-dependent, the user will be required to specify a maximal iteration.

The constructor \pythoninline{Algorithm.__init__} initialises an instance \pythoninline{alg} of the \pythoninline{Algorithm}-class. Based on the \pythoninline{design}, the sample size and dimension of the linear model are computed and stored within the parameters \pythoninline{alg.sample_size} and 
\pythoninline{alg.parameter_size}, respectively. The variable \pythoninline{alg.iteration} starts at zero and stores the maximal achieved iteration of the  \pythoninline{Algorithm}-object \pythoninline{alg}. 

The estimates corresponding to the iterations that have been computed are stored within the list \pythoninline{alg.algorithm_estimate_list}. The residuals required for the discrepancy principle are similarly stored for each iteration within a \pythoninline{np.array} that can be accessed via \pythoninline{alg.residuals}. If the user has also specified \pythoninline{alg.true_signal} and \pythoninline{alg.true_noise_level}, oracle quantities that can be computed for the specified \pythoninline{Algorithm} are initialised. 
The iterative method \pythoninline{alg.iterate} executes the private method \pythoninline{alg.__algorithm_one_iteration} for a specified number of times.

Each application of the function \pythoninline{alg.__algorithm_one_iteration} computes the next iteration of the estimator \pythoninline{alg.alorithm_estimate}, stores it in the list \pythoninline{alg.algorithm_estimate_list} and increases the maximally computed number of iterations \pythoninline{alg.iteration} by one. The method \pythoninline{alg.__algorithm_one_iteration} also updates and saves the residuals and updates theoretical quantities, for instance, \pythoninline{alg.__update_oracle_quantity} provided that \pythoninline{alg.true_signal} and also \pythoninline{alg.true_noise_level} were specified by the user. 

The function \pythoninline{alg.get_discrepancy_stop} is a general example of the stopping rule based on the discrepancy principle \eqref{eq:discrepancy_stop}. Suppose the residuals up to the maximally achieved iteration are smaller than the \pythoninline{critical_value} $\threshold$ specified by the user. In that case, the stopping index is determined based on the residuals computed so far. Otherwise, the method \pythoninline{alg.__algorithm_one_iteration} will be executed until the residuals drop below \pythoninline{critical_value} or the maximal number of iterations \pythoninline{max_iteration} is reached. Similarly, the oracles can be retrieved using \pythoninline{alg.get_balanced_oracle} provided that the true signal \pythoninline{true_signal} and the true noise level \pythoninline{true_noise_level} were specified. 

In addition to the classes for the different iterative estimation procedures, the three additional classes \pythoninline{SimulationWrapper}, \pythoninline{SimulationParameters} and \pythoninline{SimulationData} facilitate Monte-Carlo simulations. The \pythoninline{SimulationParameters}-class groups the parameters required to execute a Monte-Carlo simulation with the \pythoninline{SimulationWrapper}. It further completes several sanity checks on the input data and may detect errors in the setup of the statistical experiment. The \pythoninline{SimulationWrapper} itself contains functions like \pythoninline{run_simulation_truncated_svd} and \pythoninline{run_simulation_conjugate_gradients}, which perform Monte-Carlo simulations, which can be distributed onto a desired number of CPU kernels. The simulation then collects several crucial theoretical quantities of interest and returns them as a \citetalias{pandas} data frame object, which may be saved in the .csv format. Finally, the \pythoninline{SimulationData} may be used to create simulation data for several interesting examples of statistical inverse problems. A more detailed description of all classes and functions is available in the \href{https://earlystop.github.io/EarlyStopping/}{documentation}.  

\subsection{Truncated SVD}
\label{sec:svd}
\begin{figure}[htb]
    \centering
    \begin{subfigure}[b]{0.49\textwidth}
        \centering
          \includegraphics[width=\textwidth]{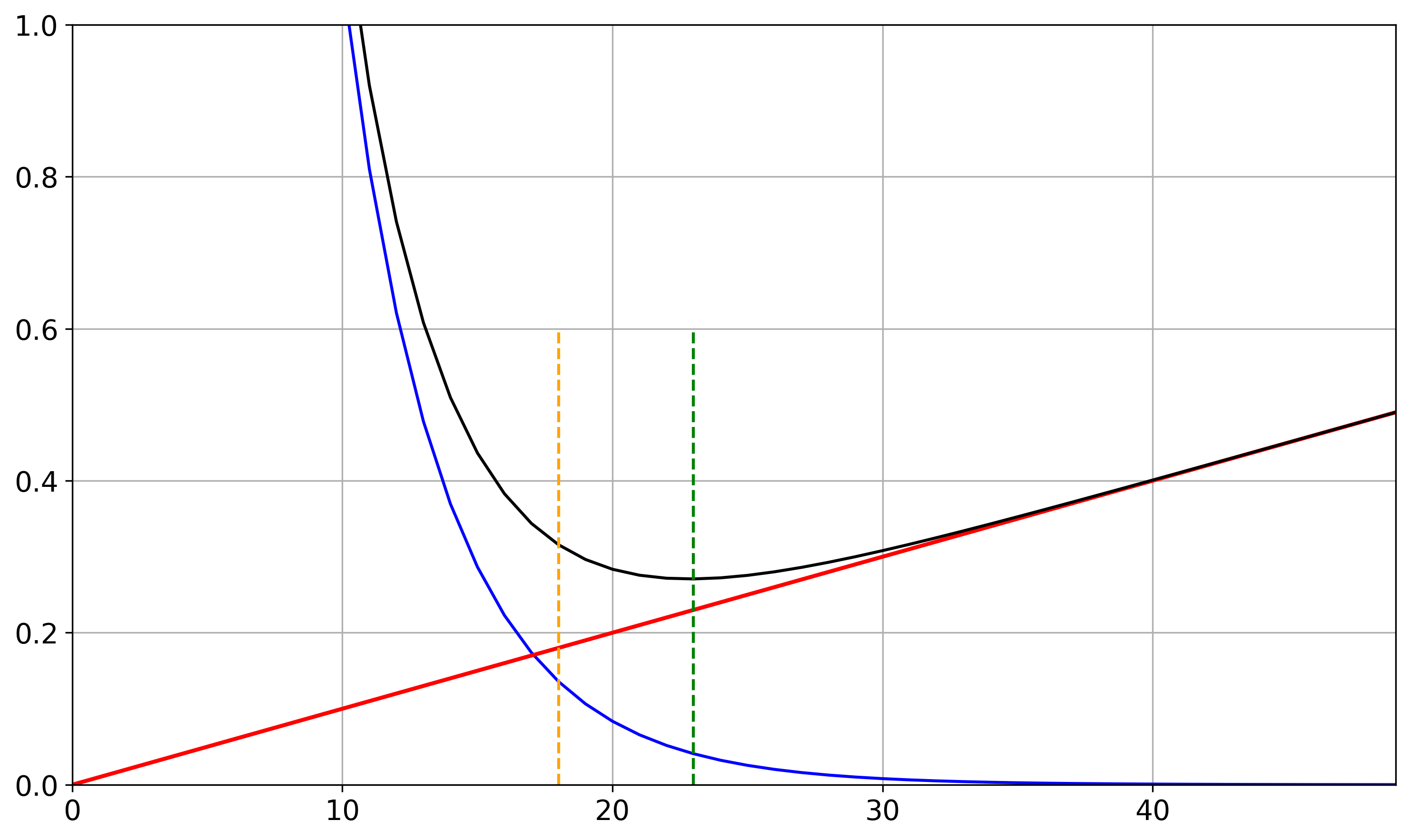}
          \caption{Weak error decomposition}
          \label{fig:error_demonstration_weak}
    \end{subfigure}
    \hfill
    \begin{subfigure}[b]{0.49\textwidth}
        \centering
          \includegraphics[width=\textwidth]{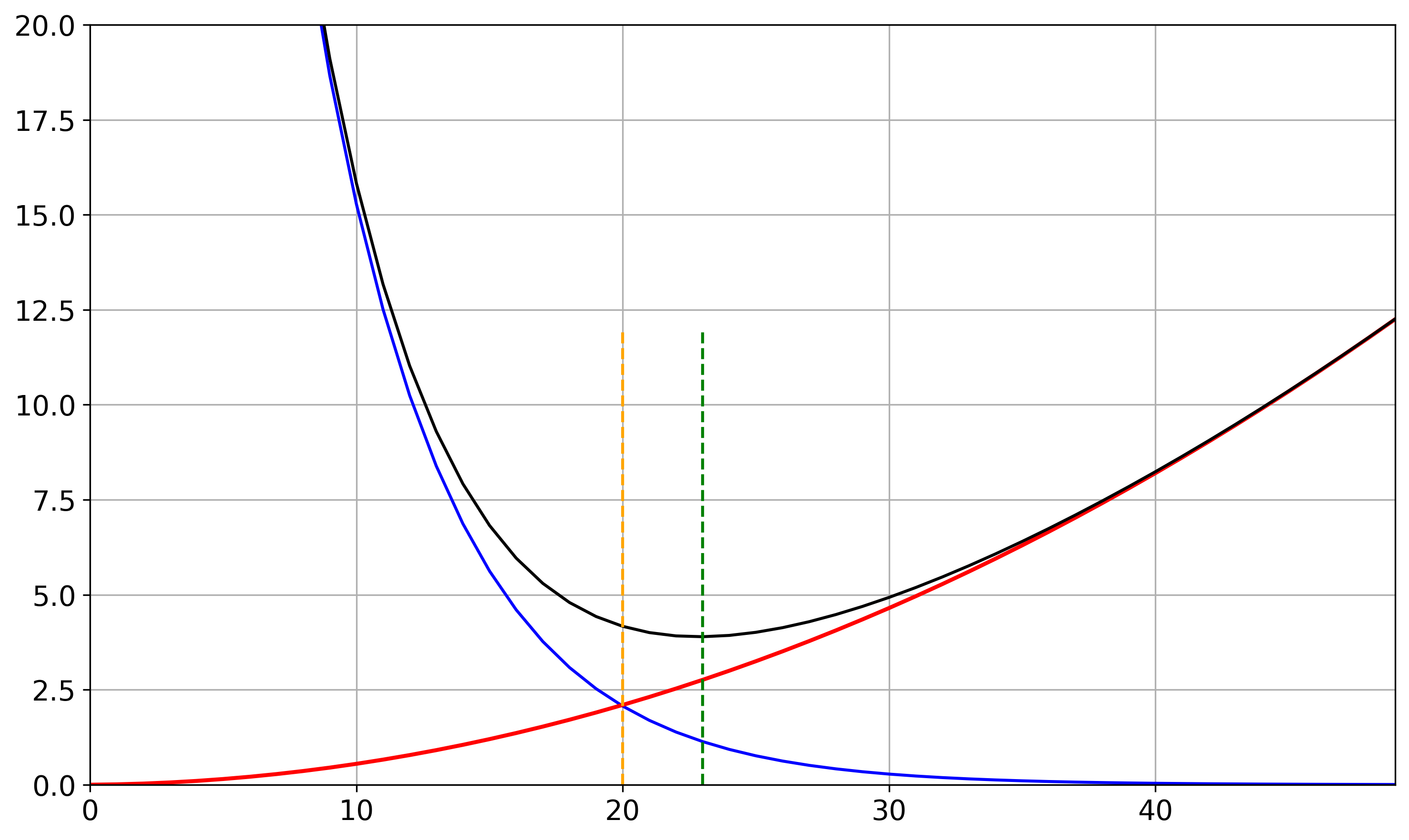}
          \caption{Strong error decomposition}
          \label{fig:error_demonstration_strong}
    \end{subfigure}
    \hfill
    \begin{subfigure}[b]{0.49\textwidth}
        \centering
  \includegraphics[width=\textwidth]{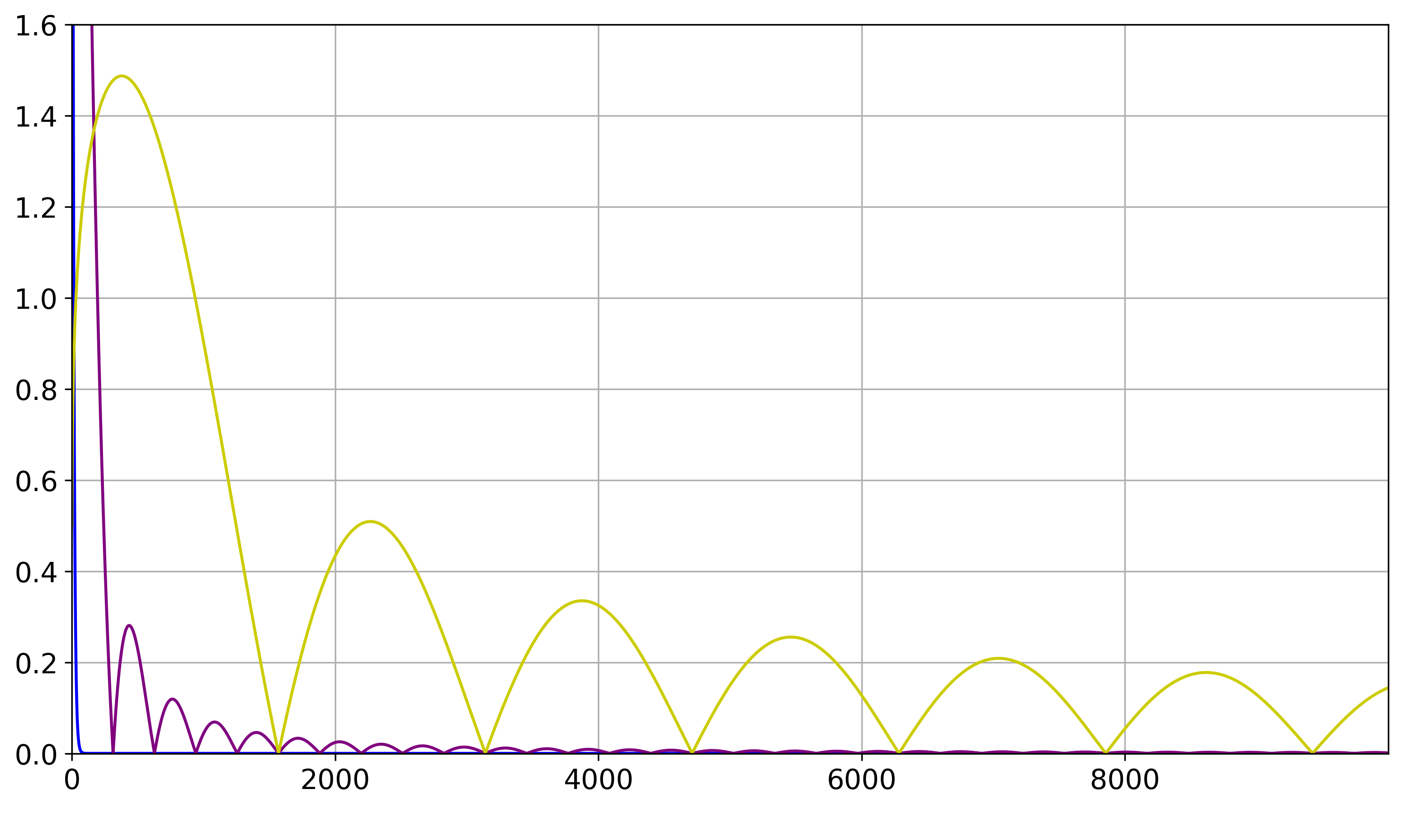}
  \caption{Signals}
  \label{fig:signals}
    \end{subfigure}
    \hfill
    \begin{subfigure}[b]{0.49\textwidth}
        \centering
        \includegraphics[width=\textwidth]{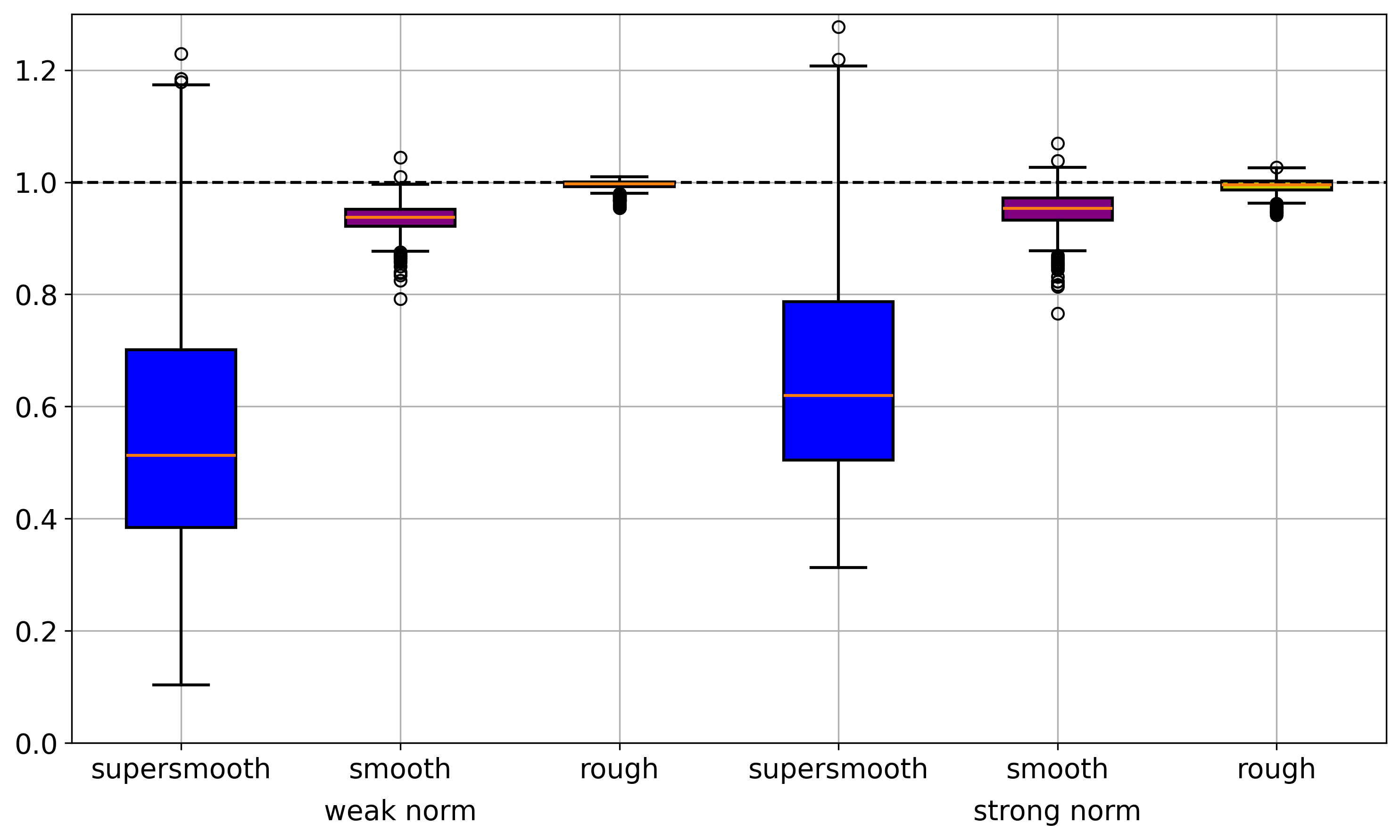}
  \caption{Relative efficiency}
  \label{fig:svd}
    \end{subfigure}
    \caption{Weak (a) and strong risk decomposition (b) for the smooth signal, with bias (blue), variance (red), error (black), balanced oracle (orange), and stopping time (green). (c) SVD representation of a super-smooth (blue), a smooth (purple), and a rough (olive) signal. (d) Relative efficiency: \eqref{eq:relative_efficiency_tSVD}. \href{https://github.com/EarlyStop/EarlyStopping/blob/main/simulations/TruncatedSVD_Replication.py}{\faGithub}
   }
    \label{fig:simulation_svd}
\end{figure}

In this section, we replicate the empirical results from \citet{blanchard2018early} based on \pythoninline{TruncatedSVD}-class and explain its functionality. 
We begin by including the required dependencies. 
\begin{python}
import numpy as np
import EarlyStopping as es
\end{python}
In \Cref{code:define_signal}, we define the smooth-signal from \citet{blanchard2018early}, which is visualised in \Cref{fig:simulation_svd}. 

\begin{python}[caption={Defining the signal.},label=code:define_signal]
sample_size   = 10000
indices       = np.arange(sample_size) + 1
signal_smooth = 5000 * np.abs(np.sin(0.01 * indices)) * indices ** (-1.6)
true_signal   = signal_smooth
\end{python}

By further defining a diagonal design matrix with $\lambda_j=j^{-1/2}$ for $j=1, \dots, \dimension$, multiplying it with a signal from \Cref{code:define_signal} and adding noise, we can now define the response variable for the statistical inverse problem \eqref{eq:inverse_problem}.
\begin{python}[caption={Defining the design matrix, response (observation) and noise.},label=code:setup_data]
true_noise_level = 0.01
noise            = true_noise_level * np.random.normal(0, 1, sample_size)
eigenvalues      = 1 / np.sqrt(indices)

from scipy.sparse import dia_matrix
design           = dia_matrix(np.diag(eigenvalues))
response         = eigenvalues * true_signal + noise
\end{python}
The data from \Cref{code:setup_data} and \Cref{code:define_signal} is fed into the class associated
with the desired iterative estimation procedure, in this case the \pythoninline{TruncatedSVD} class,
and executes a specified number of iterations through the \pythoninline{iterate}-method.
\begin{python}[caption={Initialisation for truncated SVD.},label=code:init_svd]
alg = es.TruncatedSVD(design, response, true_signal, true_noise_level,
                      diagonal=True)
alg.iterate(number_of_iterations=3000)
\end{python}
Several base attributes in the \pythoninline{TruncatedSVD}-class and several theoretical quantities, for example, the strong and weak bias and variance from \eqref{eq:truncatedSVD_weak_strong_tradeoff}, are now available since \pythoninline{true_signal} and \pythoninline{true_noise_level} were both specified. The strongly and weakly balanced oracles are computed by applying \pythoninline{alg.get_weak_balanced_oracle} and \pythoninline{alg.get_strong_balanced_oracle}, respectively.
\begin{python}[caption={Gathering theoretical quantities manually.},label=code:manual_quantities_1]
weak_variance          = alg.weak_variance
weak_bias2             = alg.weak_bias2
weak_balanced_oracle   = alg.get_weak_balanced_oracle(max_iteration=3000)

strong_variance        = alg.strong_variance
strong_bias2           = alg.strong_bias2
strong_balanced_oracle = alg.get_strong_balanced_oracle(max_iteration=3000)
\end{python}
Now that we have computed the oracle quantities, they can easily be compared with the data-driven discrepancy principle from \eqref{eq:discrepancy_stop} obtained by applying \pythoninline{alg.get_discrepancy_stop}. In fact, we already obtain \Cref{fig:error_demonstration_weak} and \Cref{fig:error_demonstration_strong} by plotting the weak and strong quantities from \Cref{code:manual_quantities_1}, respectively.
Using \pythoninline{alg.get_estimate}, we can obtain the algorithm's estimate at the determined stopping time.
\begin{python}[caption={Computing the discrepancy stop.},label=code:manual_quantities_2]
critical_value   = sample_size * true_noise_level**2

discrepancy_time = alg.get_discrepancy_stop(critical_value, 
                max_iteration=3000)
                
estimated_signal = alg.get_estimate(discrepancy_time)
\end{python}
In many cases, it is not necessary to manually collect the different parameters from \Cref{code:manual_quantities_1} and \Cref{code:manual_quantities_2}.
Instead, if we wish to perform a complete Monte-Carlo simulation study and would like to collect all theoretical quantities, we can use the \pythoninline{SimulationWrapper}.
It is most easily used in combination with the \pythoninline{SimulationParameters} class, which is set up in \Cref{code:setup_data_simulation_parameters}.
\begin{python}[caption={Using the simulation parameters class.},label=code:setup_data_simulation_parameters]
parameters_smooth = es.SimulationParameters(design=design, 
                true_signal=true_signal, true_noise_level=0.01, 
                monte_carlo_runs=500, cores=10)
\end{python}
We can now create a \pythoninline{SimulationWrapper} object from the \pythoninline{SimulationParameters} and run the simulation based on the \pythoninline{TruncatedSVD} class. The \pythoninline{data_set_name} parameter of the functions running the simulation, e.g. \pythoninline{run_simulation_truncated_svd}, is optional and is used to save the simulation results as a .csv file. If the parameter is not specified, the results will not be saved and are returned as a \pythoninline{pd.DataFrame}. 
\begin{python}
simulation_smooth   = es.SimulationWrapper(**parameters_smooth.__dict__)
simmulation_results = simulation_smooth.run_simulation_truncated_svd(
                max_iteration=500, diagonal=True)
\end{python}
Using this methodology, we can replicate the results from \citet{blanchard2018early} based on the \pythoninline{TruncatedSVD}-class. The Monte-Carlo experiment is performed using the super smooth, smooth and rough signals in \Cref{fig:signals} and $M=500$ Monte-Carlo iterations. The noise level is set to \(\delta = 0.01\), and the dimension is chosen as \(n = 10,000\). The threshold is set as $\threshold = n \noiselevel^2=1$. \Cref{fig:svd} illustrates the relative efficiency in RMSE as 
\begin{equation}
\label{eq:relative_efficiency_tSVD}
\begin{aligned}
&\text{min}_{m \geq 0}\mathbb{E}[\Vert \estimator^{(m)}-\truesignal \Vert^{2}_{A}]^{1/2} / \mathbb{E}[\Vert \estimator^{(\DP)}-\truesignal \Vert^{2}_{A}]^{1/2},\quad \quad \ \ \text{weak relative efficiency},\\
&\text{min}_{m \geq 0}\mathbb{E}[\Vert \estimator^{(m)}-\truesignal \Vert^{2}]^{1/2} / \mathbb{E}[\Vert \estimator^{(\DP)}-\truesignal \Vert^{2}]^{1/2},\quad \quad \ \ \ \text{strong relative efficiency}.
\end{aligned}
\end{equation}
These quantities are also automatically computed within the \pythoninline{SimulationWrapper}-class. The balanced oracle iterations (weak) are (34, 316, 1356), and
classical oracle iterations in strong norm are (43, 504, 1331). The results match the results from \citet{blanchard2018early}, and we refer to the former paper for a detailed interpretation of the simulation results.

\subsection{Landweber}
\label{sec:land}
Since the signals and design from 
\Cref{code:setup_data} and \Cref{code:define_signal} are contained within the \pythoninline{SimulationData} class, they do not need to be created manually, see \Cref{code:setup_data_simulation_data}. 
\begin{python}[caption={Using the SimulationData class.},label=code:setup_data_simulation_data]
design, response, true_signal = es.SimulationData.diagonal_data(
                sample_size=2000, type="smooth")
\end{python}
This time, we use the data from \Cref{code:setup_data_simulation_data} as an input for the  \pythoninline{Landweber} class. The variables \pythoninline{true_signal} and \pythoninline{true_noise_level} are optional and required only for the theoretical analysis. The \pythoninline{learning_rate} can also be specified in the case of the \pythoninline{Landweber} class. 
\begin{python}[caption={Initialisation for Landweber.},label=code:init_landweber]
alg = es.Landweber(design, response, learning_rate=1, 
                true_signal=true_signal, 
                true_noise_level=true_noise_level)
                
alg.iterate(number_of_iterations = 3000)
\end{python}
Using the \pythoninline{iterate}-method, we have further executed a desired number of iterations. Several base attributes in the \pythoninline{Landweber}-class and several theoretical quantities, for instance, from \eqref{eq:bias_variance_landweber} are now available since \pythoninline{true_signal} and \pythoninline{true_noise_level} were both specified. Based on the bias and variance from \Cref{code:bias_variance_landweber}, the balanced oracles are computed through \pythoninline{alg.get_weak_balanced_oracle} and \pythoninline{alg.get_strong_balanced_oracle}, see \Cref{code:manual_quantities_landweber}.

\begin{python}[caption={Gathering bias and variance for Landweber.},label=code:bias_variance_landweber]
weak_variance   = alg.weak_variance
weak_bias2      = alg.weak_bias2

strong_variance = alg.strong_variance
strong_bias2    = alg.strong_bias2
\end{python}

\begin{figure}[htb]
    \centering
    \begin{subfigure}[b]{0.49\textwidth}
          \centering
          \includegraphics[width=\textwidth]{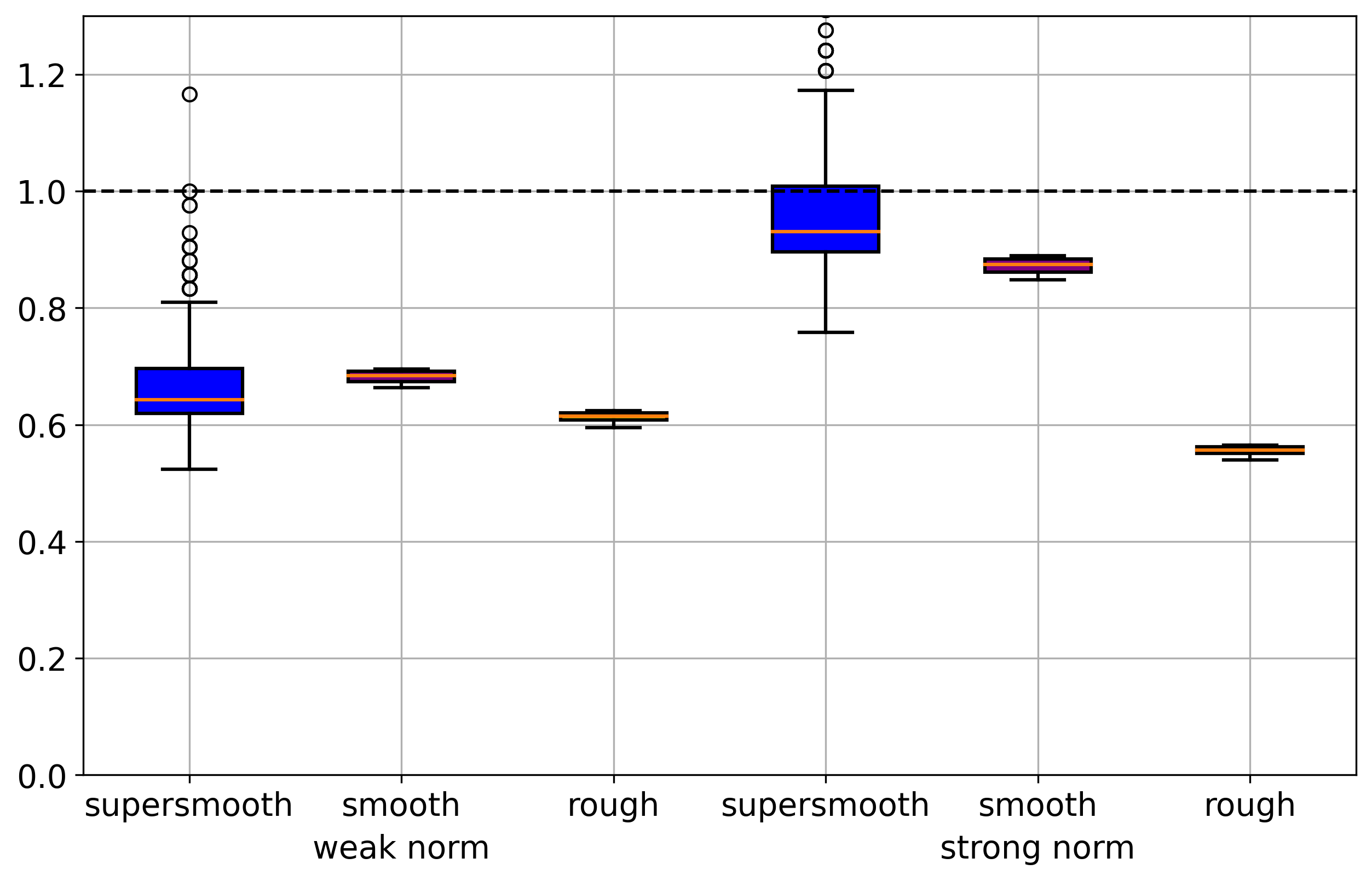}
          \caption{Relative number of iterations}
          \label{subfigure:boxplot_iteration_landweber}
    \end{subfigure}
    \hfill
    \begin{subfigure}[b]{0.49\textwidth}
          \centering
          \includegraphics[width=\textwidth]{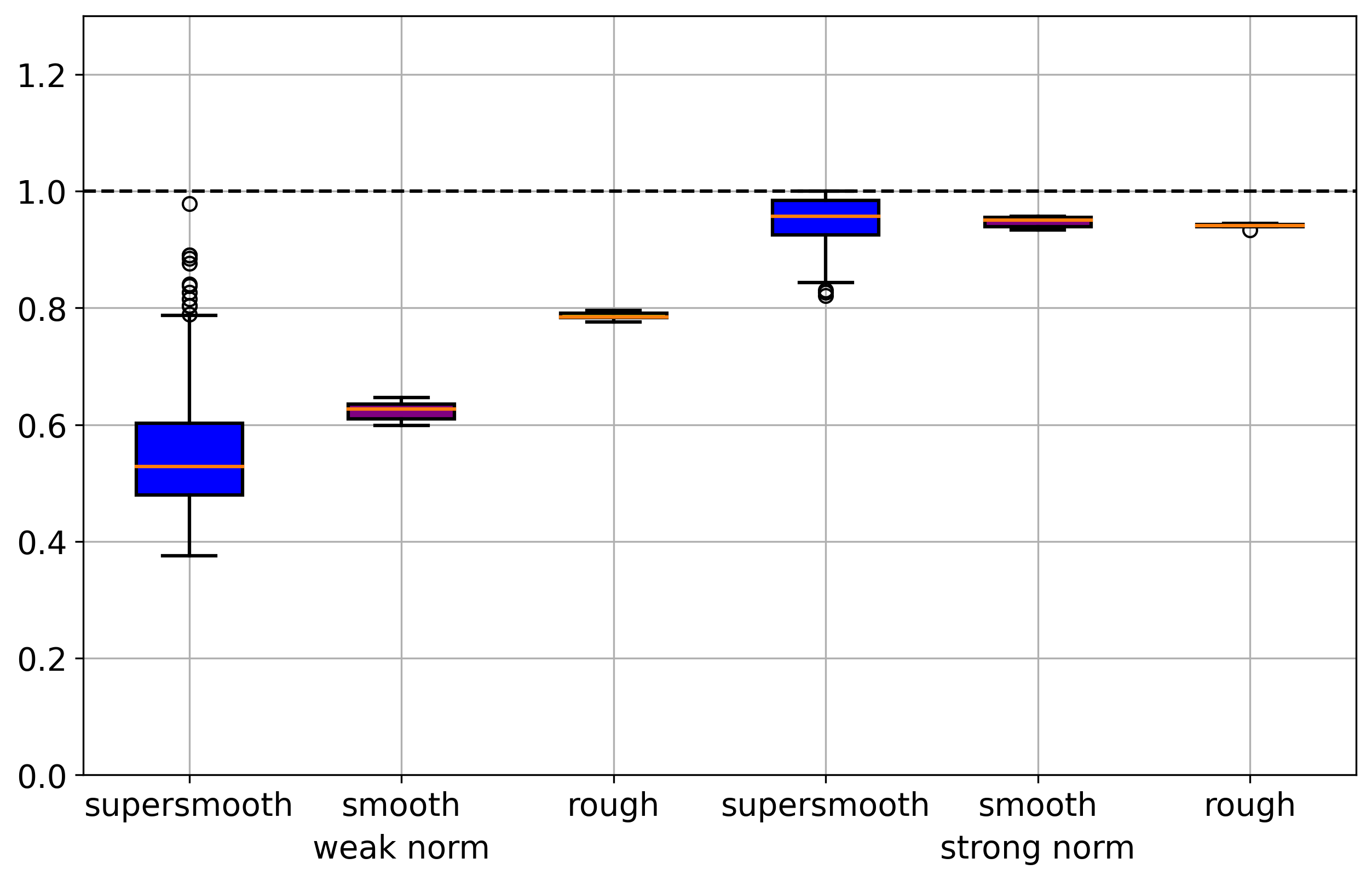}
          \caption{Relative efficiency}
          \label{subfigure:boxplot_efficiency_landweber}
    \end{subfigure}
    \caption{(a) Relative number of Landweber iterations for $\DP$ divided by the balanced oracle iteration. (b) Relative efficiency: \eqref{eq:relative_efficiency_tSVD}. \href{https://github.com/EarlyStop/EarlyStopping/blob/main/simulations/Landweber_Replication.py}{\faGithub}}
    \label{fig:simulation_bhr}
\end{figure}

\begin{python}[caption={Gathering stopping times for Landweber.},label=code:manual_quantities_landweber]
weak_balanced_oracle   = alg.get_weak_balanced_oracle(3000)
strong_balanced_oracle = alg.get_strong_balanced_oracle(3000)

critical_value         = sample_size * true_noise_level**2
discrepancy_time       = alg.get_discrepancy_stop(critical_value,
                max_iteration=3000)

estimated_signal       = alg.get_estimate(discrepancy_time)
\end{python}
As in the case of the \pythoninline{TruncatedSVD}-class, we can use the 
\pythoninline{SimulationWrapper}-class in combination with the \pythoninline{SimulationParameters} class, which is set up in \Cref{code:setup_data_simulation_parameters}.
\begin{python}[caption={Using the SimulationParameters class.},label=code:setup_wrapper_landweber]
simulation_smooth   = es.SimulationWrapper(**parameters_smooth.__dict__)
simmulation_results = simulation_smooth.run_simulation_landweber(
                max_iteration=2000)
\end{python}
The data frame returned by the \pythoninline{SimulationWrapper} already contains all required information to successfully replicate the simulations of \citet{blanchard2018optimal}, as illustrated in \Cref{fig:simulation_bhr} for $500$ Monte-Carlo iterations, where the noise level is $\noiselevel=0.01$, and the dimension is $\samplesize=10.000$. We are interested in the early stopping iteration $\DP$ for the threshold $\threshold = \samplesize \noiselevel^2=1$. The left plot shows the relative number of iterations in weak and strong norm. The weak balanced oracle iterations are (42, 312, 1074), and the strong balanced oracle iterations are (29, 244, 1185) for the supersmooth, smooth, and rough signals, respectively. The right plot shows the relative efficiencies in analogy to the previous section. For a precise analysis and theoretical interpretation of the results, we refer to Section 4.2 in \citet{blanchard2018optimal}.
\subsection{Conjugate Gradient Descent}
\begin{figure}[htb]
    \centering
    \begin{subfigure}[b]{0.49\textwidth}
        \centering
          \includegraphics[width=\textwidth]{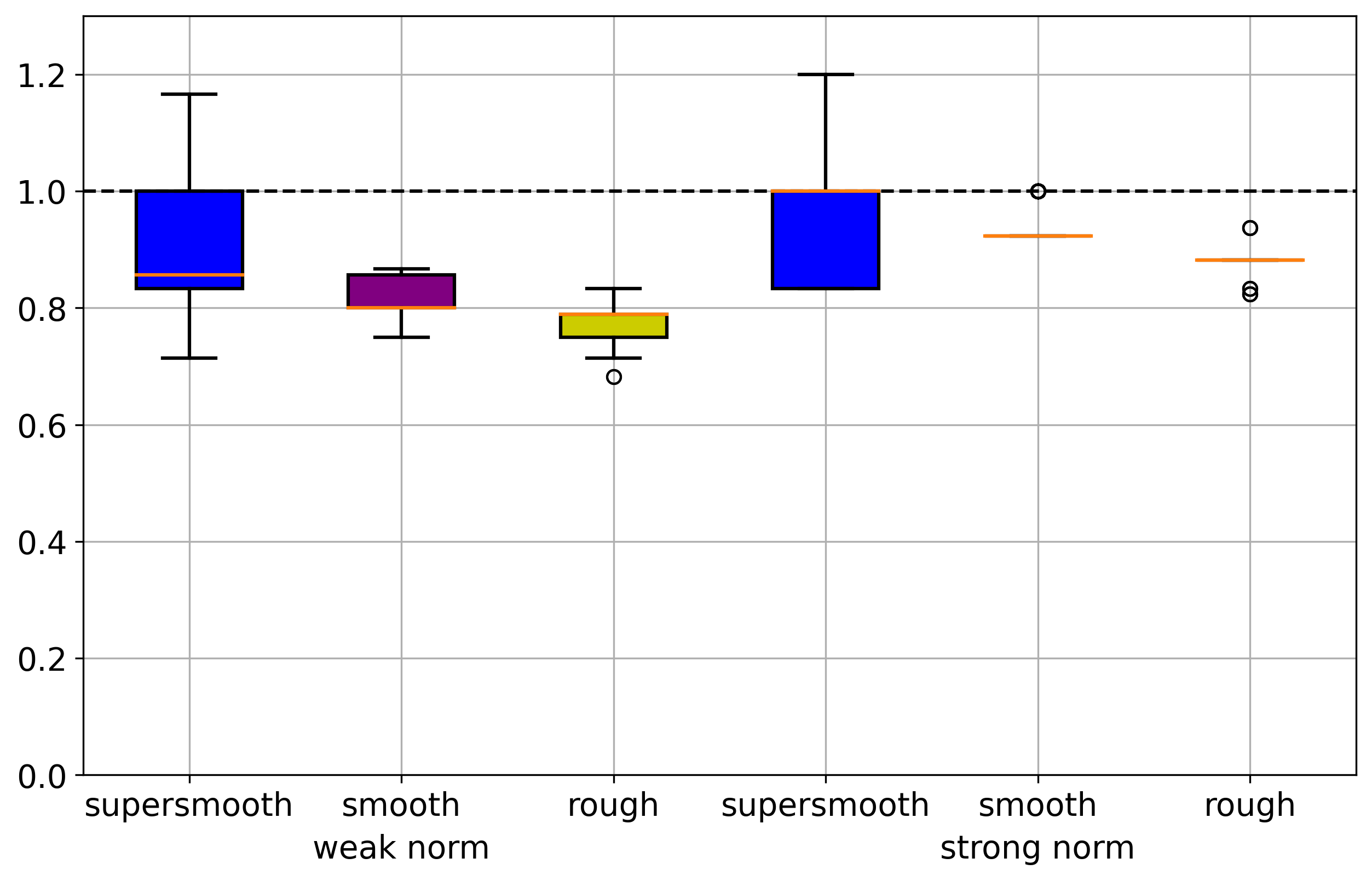}
          \caption{Relative number of iterations}
          \label{subfigure:relative_number_iterations_cg}
    \end{subfigure}
    \hfill
    \begin{subfigure}[b]{0.49\textwidth}
        \centering
        \includegraphics[width=\textwidth]{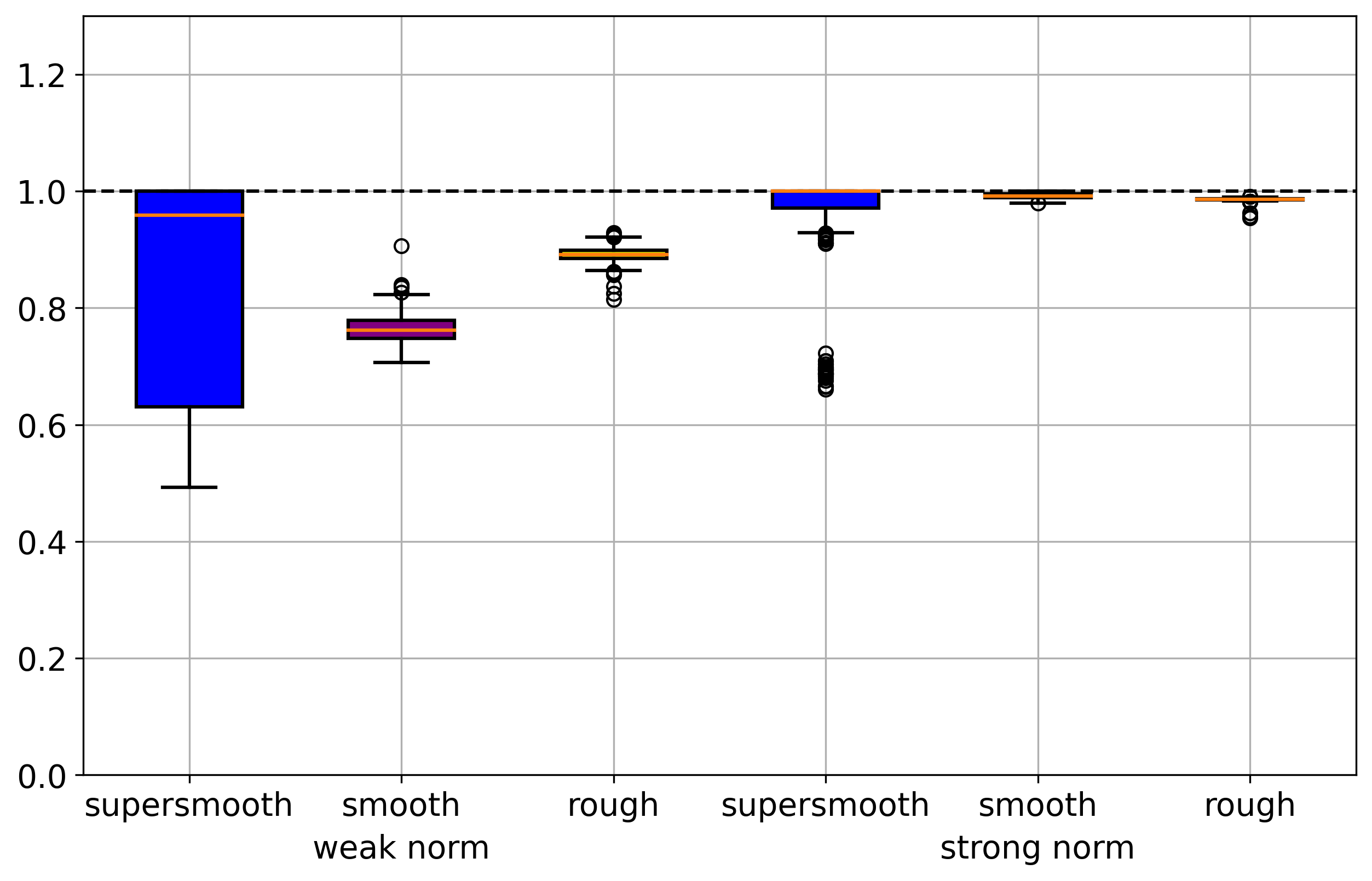}
        \caption{Relative efficiency}
         \label{subfigure:relative_efficiency_cg}
    \end{subfigure}
    \caption{(a) Relative number of conjugate gradient iterations for $\DP$ divided by the balanced oracle iteration. (b) Relative efficiency: \eqref{eq:relative_efficiency_cg}. \href{https://github.com/EarlyStop/EarlyStopping/blob/main/simulations/ConjugateGradients_Replication.py}{\faGithub}}
    \label{fig:simulation_cg}
\end{figure}

Based on the setup of \Cref{code:init_svd} and \Cref{code:init_landweber}, we demonstrate the functionality of the \pythoninline{ConjugateGradients}-class.
In \Cref{code:init_cg}, a \pythoninline{ConjugateGradients} instance is initialised analogously to \Cref{code:init_landweber} and \Cref{code:init_svd} for the \pythoninline{Landweber} and \pythoninline{TruncatedSVD}-class, respectively.
\begin{python}[caption={Initialisation for CG.},
label=code:init_cg]
alg = es.ConjugateGradients(design, response, true_signal=true_signal, 
                true_noise_level=true_noise_level,
                computation_threshold = 10 ** (-8))
    
alg.iterate(number_of_iterations=200)
\end{python}
Note that the additional parameter \pythoninline{computation_threshold} is required for the emergency stop, as described in \citet{hucker2024early}.
As usual, the \pythoninline{iterate}-method allows us to execute a specified number of iterations from the conjugate gradient algorithm. In contrast to the \pythoninline{Landweber} and the \pythoninline{TruncatedSVD}-class, several of the oracle quantities are empirical and remain dependent on the noise since the associated classical bias-variance decomposition is hard to access. The corresponding empirical quantities can be accessed as described in \Cref{code:empirical_oracle_cg}. Note that several quantities allow for an interpolated version, see \Cref{remark:interpolation}. 
\begin{python}[caption={Empirial oracle quantities.},label=code:empirical_oracle_cg]
strong_empirical_oracle = alg.get_strong_empirical_oracle(max_iteration=200, 
                interpolation=False)
weak_empirical_oracle   = alg.get_weak_empirical_oracle(max_iteration=200, 
                interpolation=False)

iteration = 100
get_strong_empirical_risk(max_iteration = iteration)
get_weak_empirical_risk(max_iteration = iteration)
\end{python}
The discrepancy stop is obtained in \Cref{code:discrepancy_stop_cg}. 
\begin{python}[caption={Discrepancy stop conjugate gradients.}, label=code:discrepancy_stop_cg]
critical_value   = sample_size * true_noise_level**2
discrepancy_time = alg.get_discrepancy_stop(critical_value, 
                max_iteration=100, interpolation=False)
\end{python}
We replicate the simulation study of \citet{hucker2024early} for $500$ Monte-Carlo iterations.
We can again use the \pythoninline{SimulationWrapper}-class for the Monte-Carlo simulation, see \Cref{code:wrapper_cg}.
\begin{python}[caption={Simulation wrapper conjugate gradients.}, label=code:wrapper_cg]
simulation_smooth   = es.SimulationWrapper(**parameters_smooth.__dict__)
simmulation_results = simulation_smooth.run_simulation_conjugate_gradients(
                max_iteration=2000)
\end{python}
The relative number of iterations and the relative efficiency are illustrated in \Cref{subfigure:relative_number_iterations_cg} and \Cref{subfigure:relative_efficiency_cg}, respectively.
The quantities in this figure are computed without additional interpolation, which is particularly visible for the supersmooth signal. Note that the weak and strong relative efficiencies are now empirical quantities computed according to:
\begin{equation}
\label{eq:relative_efficiency_cg}
\begin{aligned}
&\text{min}_{m \geq 0}\Vert \estimator^{(m)}-\truesignal \Vert_{A} / \Vert \estimator^{(\DP)}-\truesignal \Vert_{A},\quad \quad \text{weak (empirical) relative efficiency},\\
&\text{min}_{m \geq 0}\Vert \estimator^{(m)}-\truesignal \Vert / \Vert \estimator^{(\DP)}-\truesignal \Vert, \quad \quad \quad\ \text{strong (empirical) relative efficiency}.
\end{aligned}
\end{equation}

\subsection{$L^2$-Boosting}
\begin{figure}[htb]
  \centering
  \begin{subfigure}[b]{0.49\textwidth}
      \centering
        \includegraphics[width=\textwidth]{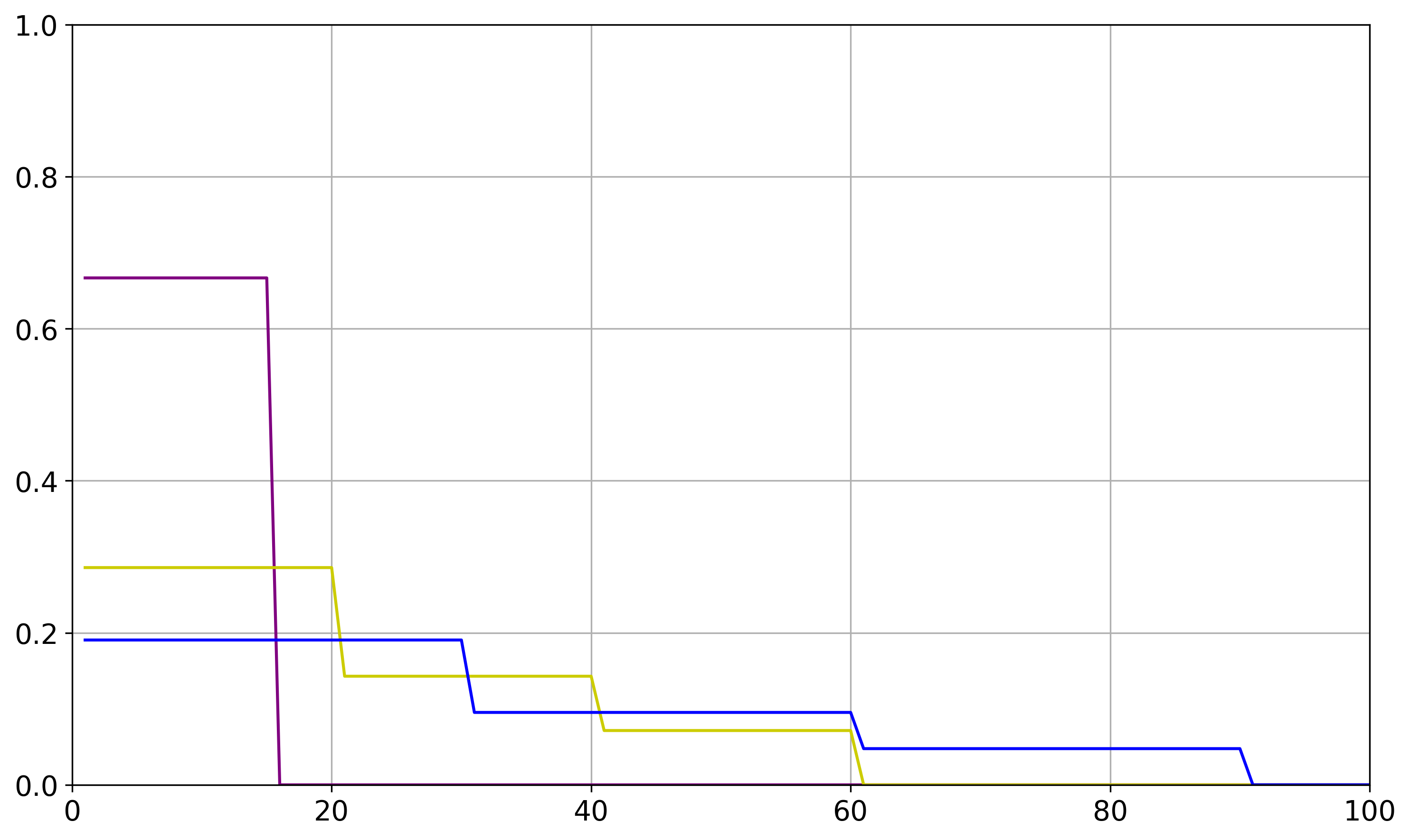}
        \caption{s-sparse signals: $\beta_{15}$ (purple), $\beta_{60}$ (olive), $\beta_{90}$ (blue) }
        \label{fig:boosting_signals_1}
  \end{subfigure}
  \hfill
  \begin{subfigure}[b]{0.49\textwidth}
      \centering
        \includegraphics[width=\textwidth]{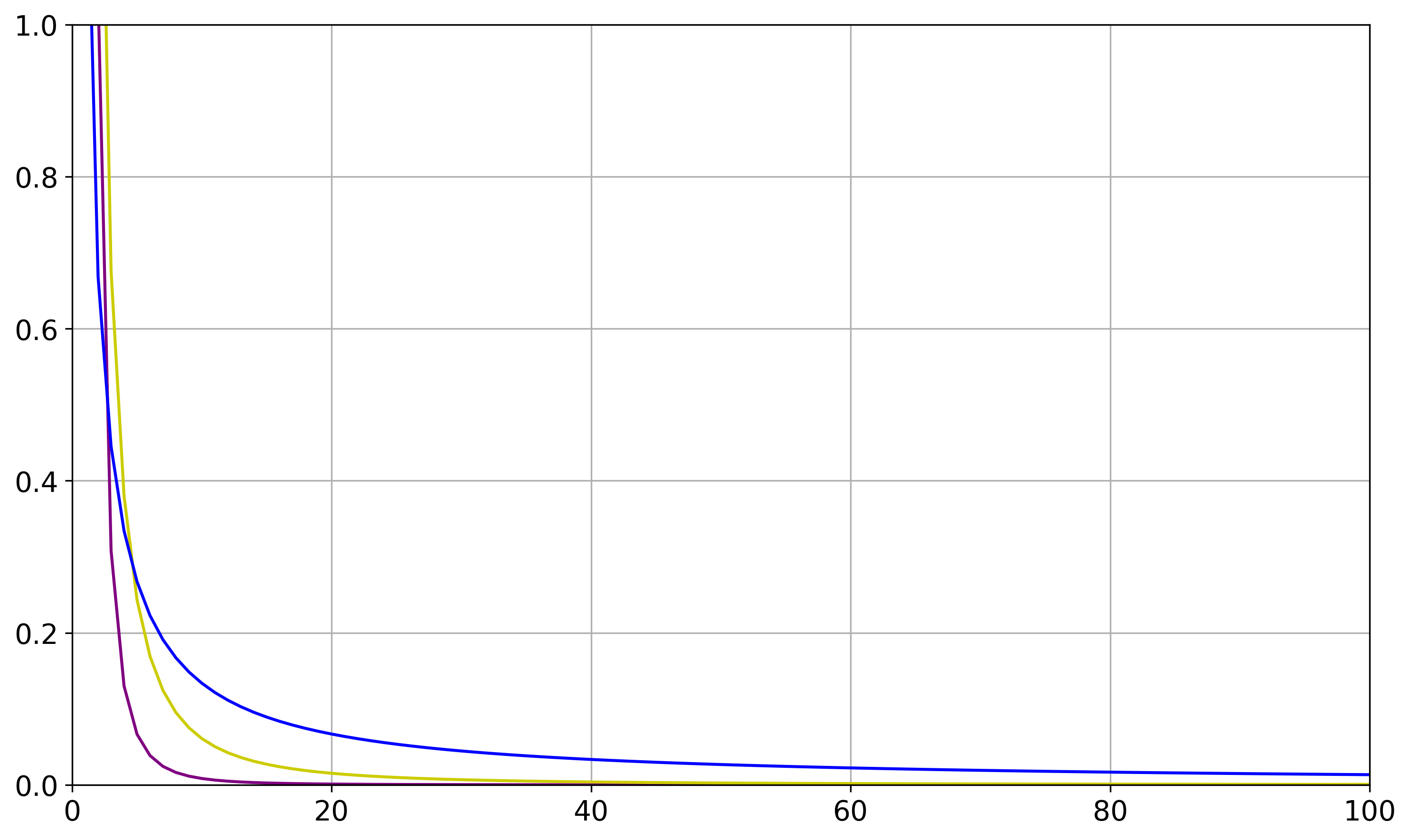}
        \caption{beta-smooth signals: $\beta_3$ (purple), $\beta_2$ (olive), $\beta_1$ (blue)}
        \label{fig:boosting_signals_2}
  \end{subfigure}
  \caption{Signals from the simulation in \citet{stankewitz2024early}.}
  \label{fig:boosting_signals}
\end{figure}

We consider the gamma-sparse and s-spare signals from the simulation in \citet{stankewitz2024early},
which are illustrated in \Cref{fig:boosting_signals}.

We define one of the signals and simulate data from the
high-dimensional linear model \eqref{eq_IV_1_HDLinearModel} in \Cref{code:L2-Boost_signals_gamma_sparse}.
\begin{python}[caption={Generate gamma-sparse signal.},
label=code:L2-Boost_signals_gamma_sparse]
sample_size    = 1000
parameter_size = 1000
design         = np.random.multivariate_normal(np.zeros(parameter_size),
                np.identity(parameter_size), sample_size)

beta_3         = 1 / (1 + np.arange(parameter_size)) ** 3
beta_3         = 10 * beta / np.sum(np.abs(beta))

noise          = np.random.normal(0, 1, sample_size)
response       = design @ beta_3 + noise
\end{python}
In \Cref{code:bosting_usage}, we instantiate the \pythoninline{L2_boost}-class, compute the boosting
path up to iteration 500 and access the theoretical quantities from \Cref{ssec_BoostingInHighDimensionalLinearModels}.
\begin{python}[caption={Using the boosting class.}, label=code:bosting_usage]
alg = es.L2_boost(design, response, beta_3)

alg.iterate(number_of_iterations = 500)

alg.bias2
alg.stochastic_error
alg.risk

alg.get_balanced_oracle()
\end{python}
For the discrepancy principle in \eqref{eq:discrepancy_principle_boosting}, the
\pythoninline{L2_boost}-class provides a noise estimation method based on the scaled Lasso algorithm
from \citet{sun2012scaled}. 
The \pythoninline{critical_value} argument of the method corresponds to \( \kappa \) in \eqref{eq:discrepancy_principle_boosting}. 
For practical use, this can be set equal to the noise estimate.
For the residual ratio stopping method from \citet{kueck2023estimation}, the hyperparameter defaults
to \( K = 1.2 \) and the confidence level is set to \( \alpha = 0.95 \).
The class logs the latest current iteration, here 500, up to which we can query the minimiser of the 
AIC criterion.
The penalty constant defaults to the common choice \( K_{\text{AIC}} = 2 \).
\begin{python}[caption={Methods for choosing a boosting iteration.}, label=code:choosing_boosting_iteration]
noise_estimate = alg.get_noise_estimate()
discrepancy_stop = alg.get_discrepancy_stop(critical_value=noise_estimate,  
                max_iteration=500)

residual_ratio_stop = alg.get_residual_ratio_stop(max_iteration=500, K=1.2, 
                alpha=0.95)

alg.iteration
alg.get_aic_iteration(K=2, max_iteration = 500)
alg.get_aic_iteration(K=2, max_iteration = discrepancy_stop)
alg.get_aic_iteration(K=2, max_iteration = residual_ratio_stop)
\end{python}
For a newly initialised instance of \pythoninline{L2_boost}-class, querying one of the sequential
stopping times iterates the learning procedure up to that index or the maximal iteration specified.
By passing an iteration to the \pythoninline{get_aic_iteration}-function, we compute the AIC minimiser up to that iteration. 
For the stopping times, this results in the two-step procedure described in \Cref{eq_HDAIC}.
The \pythoninline{L2_boost} class is also featured in the simulation wrapper and Monte-Carlo
simulation can be executed as shown in \Cref{code:wrapper_l2_boost}. 

\begin{python}[caption={Simulation wrapper for $L^2$-boosting.}, label=code:wrapper_l2_boost]
parameters_gamma_sparse = es.SimulationParameters(design = design,
                true_signal = beta_3, true_noise_level = 1,
                monte_carlo_runs = 100, cores=10)

simulation = es.SimulationWrapper(**parameters_gamma_sparse .__dict__)
simmulation_results = simulation.run_simulation_L2_boost(max_iteration=500)    
\end{python}
Using this, we can replicate a simulation of the relative efficiencies $\min_{m \geq 0} \| \widehat{f}^{(m)} - f^{*} \|_{n} / \| \widehat{f}^{(\tau)} - f^{*} \|_{n}$ from \citet{stankewitz2024early}.
Initially, we consider \( \tau = \tau^{\text{DP}} \) and \( f^{*} = X \beta^{*} \) for a standard
Gaussian design matrix \( X \) and \( \beta^{*} \) one of the signals in \Cref{fig:boosting_signals}. 
Comparing with the results when \( \tau = \tau^{(\text{RR})} \) shows that both sequential
stopping methods perform reasonably well, with neither outperforming the other over all signals.
\begin{figure}[H]
    \centering
    \begin{subfigure}[b]{0.49\textwidth}
        \centering
          \includegraphics[width=\textwidth]{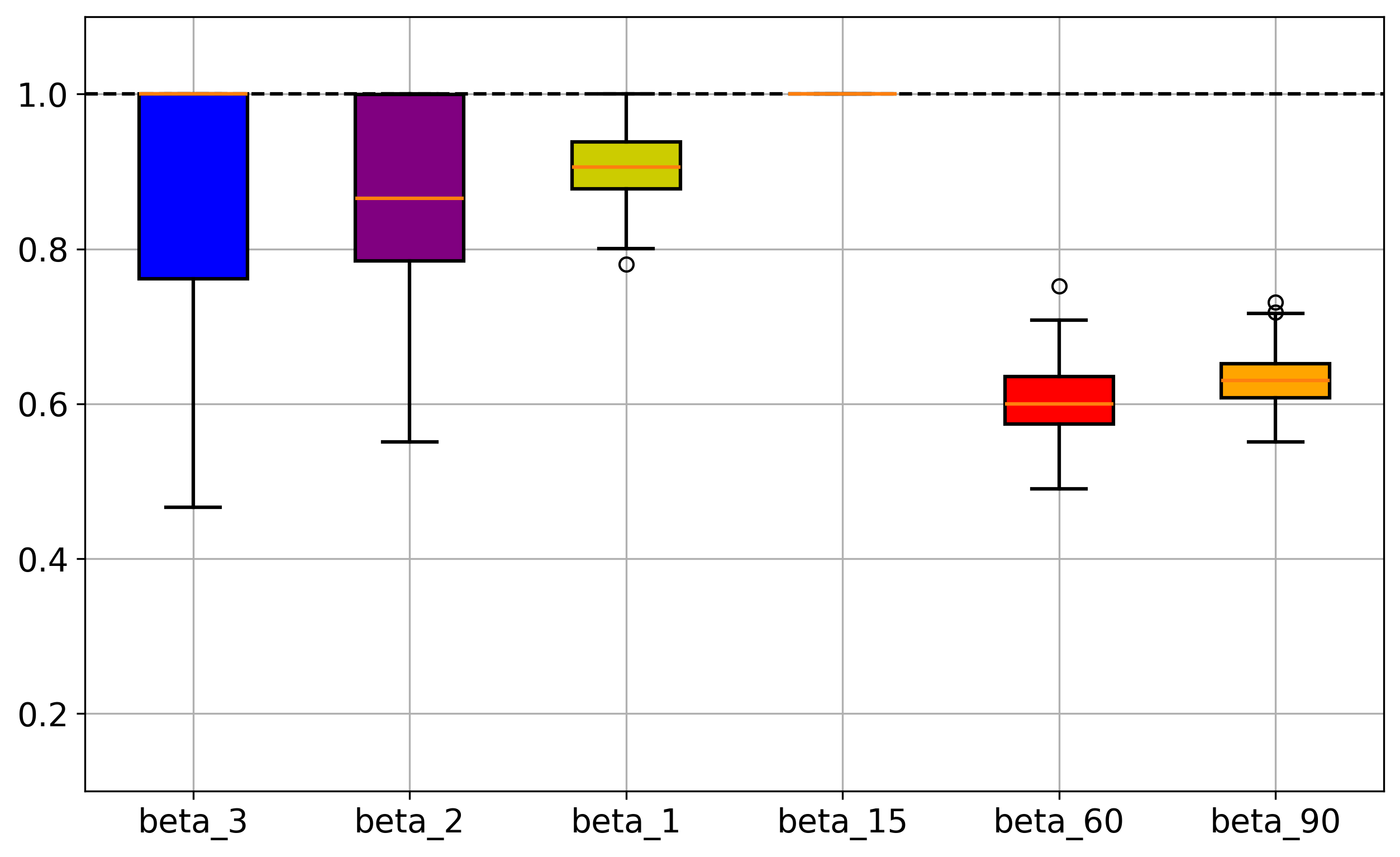}
          \caption{Discrepancy stop.}
    \end{subfigure}
    \hfill
    \begin{subfigure}[b]{0.49\textwidth}
        \centering
          \includegraphics[width=\textwidth]{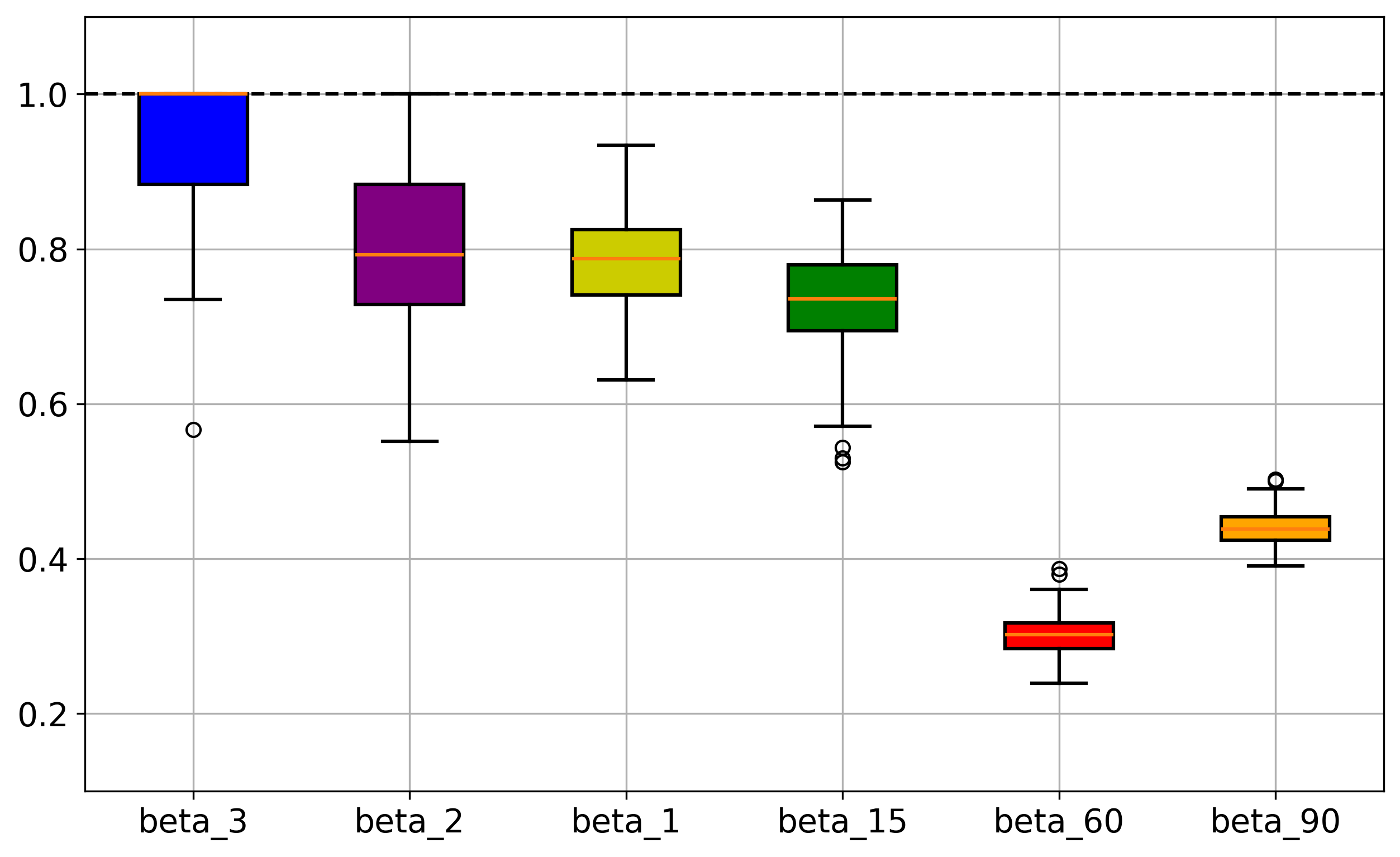}
          \caption{Two-step procedure for the residual ratio stop.}
    \end{subfigure}
    \hfill
    \begin{subfigure}[b]{0.49\textwidth}
        \centering
          \includegraphics[width=\textwidth]{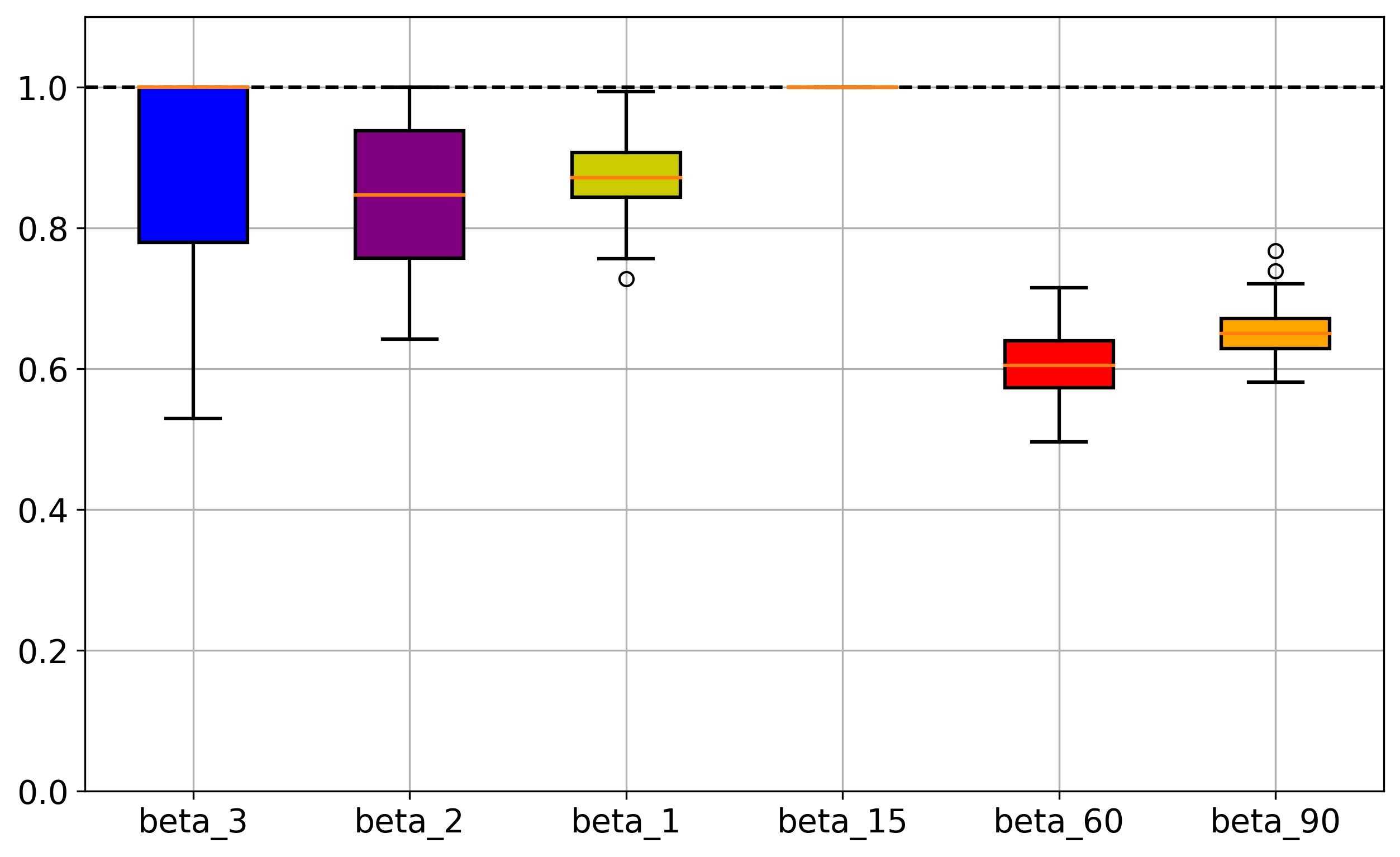}
          \caption{Two-step procedure for the discrepancy stop.}
    \end{subfigure}
    \hfill
    \begin{subfigure}[b]{0.49\textwidth}
        \centering
          \includegraphics[width=\textwidth]{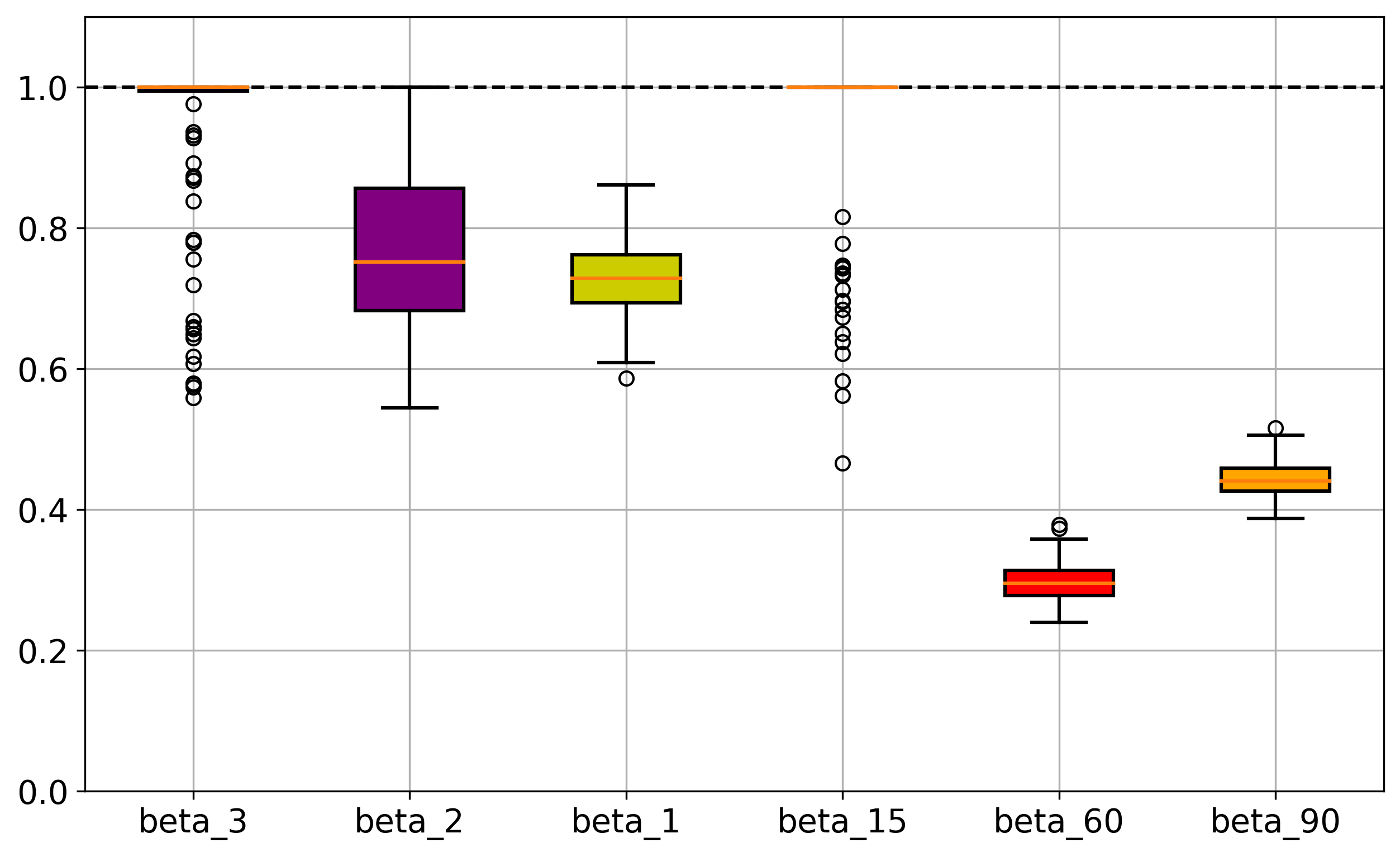}
          \caption{2-step residual ratio stop}
    \end{subfigure}
    \caption{Comparison of the relative efficiencies of the discrepancy stop and the residual ratio stop as well as their associated two-step procedures on the signals from \Cref{fig:boosting_signals}. \href{https://github.com/EarlyStop/EarlyStopping/blob/main/simulations/L2Boost_Replication.py}{\faGithub}}
\end{figure}
For the first four signals, stopping tends to be late.
Here, both methods profit from combining them with the Akaike criterion in the two-step procedure.
For the last two signals, stopping tends to be before the optimal index.
Consequently, the two-step procedure cannot provide additional improvement.
Again, this underlines the reasoning in Section \ref{ssec_BoostingInHighDimensionalLinearModels} for
slightly tuning the algorithm towards larger stopping times.
Simulation results in that setting are available in \citet{stankewitz2024early}.
\subsection{Regression tree}
\begin{figure}[htb]
    \centering
    \begin{subfigure}[b]{0.49\textwidth}
        \centering
        \includegraphics[width=\textwidth]{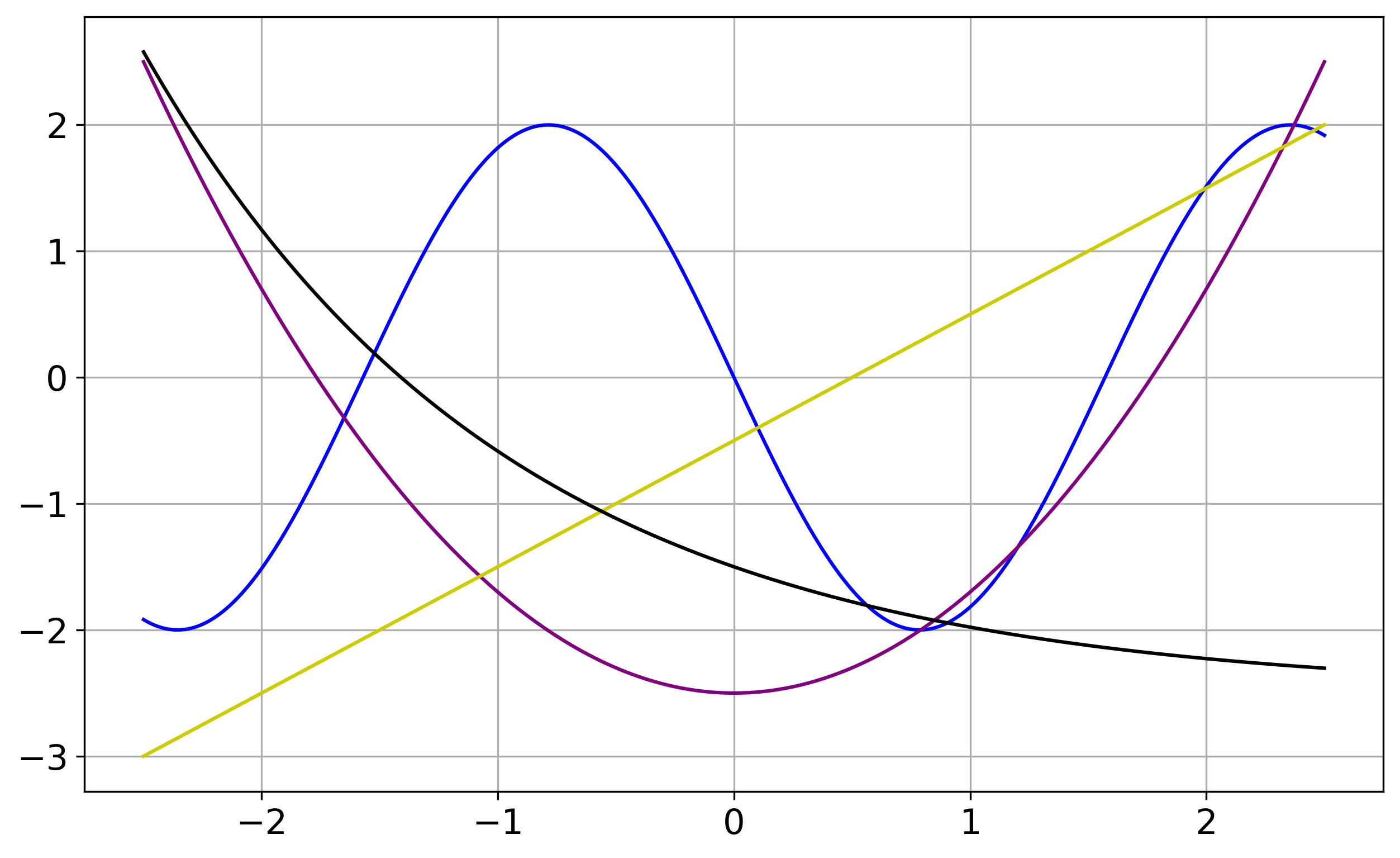}
  \caption{Smooth functions}
    \end{subfigure}
    \hfill
    \begin{subfigure}[b]{0.49\textwidth}
        \centering
        \includegraphics[width=\textwidth]{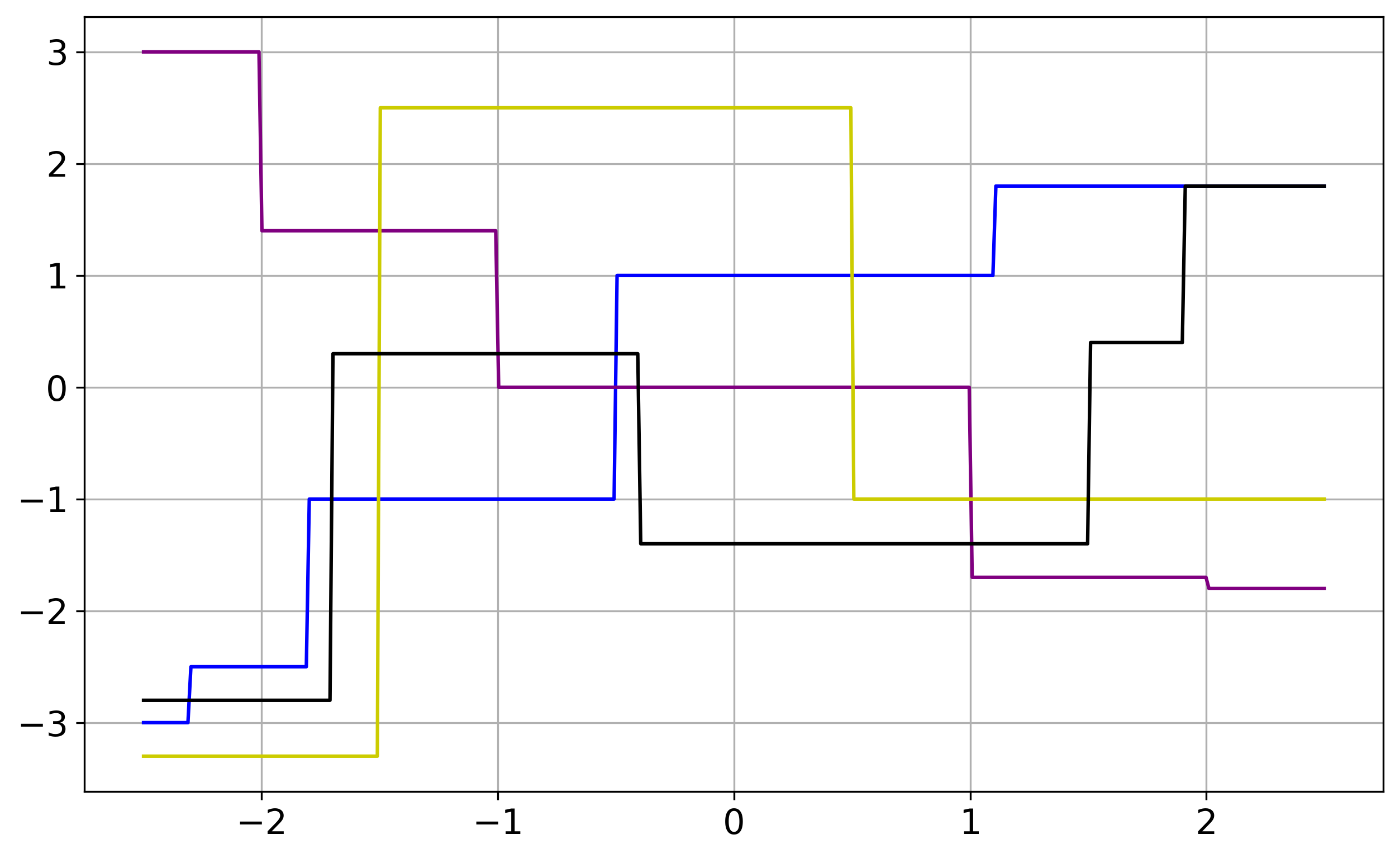}
  \caption{Step functions}
    \end{subfigure}\\
        \begin{subfigure}[b]{0.49\textwidth}
        \centering
                \includegraphics[width=\textwidth]{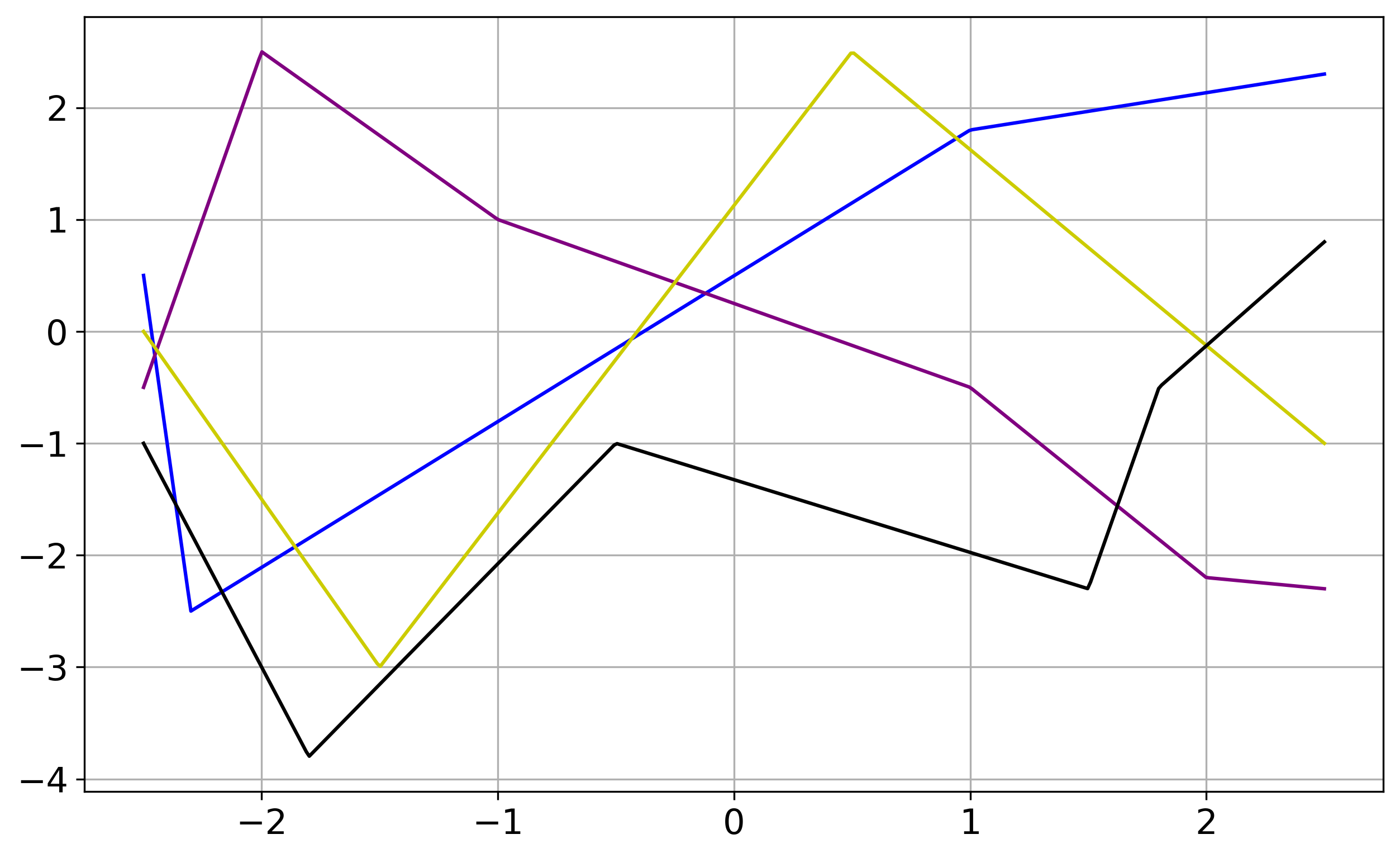}
  \caption{Piecewise linear functions}
    \end{subfigure}
    \hfill
    \begin{subfigure}[b]{0.49\textwidth}
        \centering
        \includegraphics[width=\textwidth]{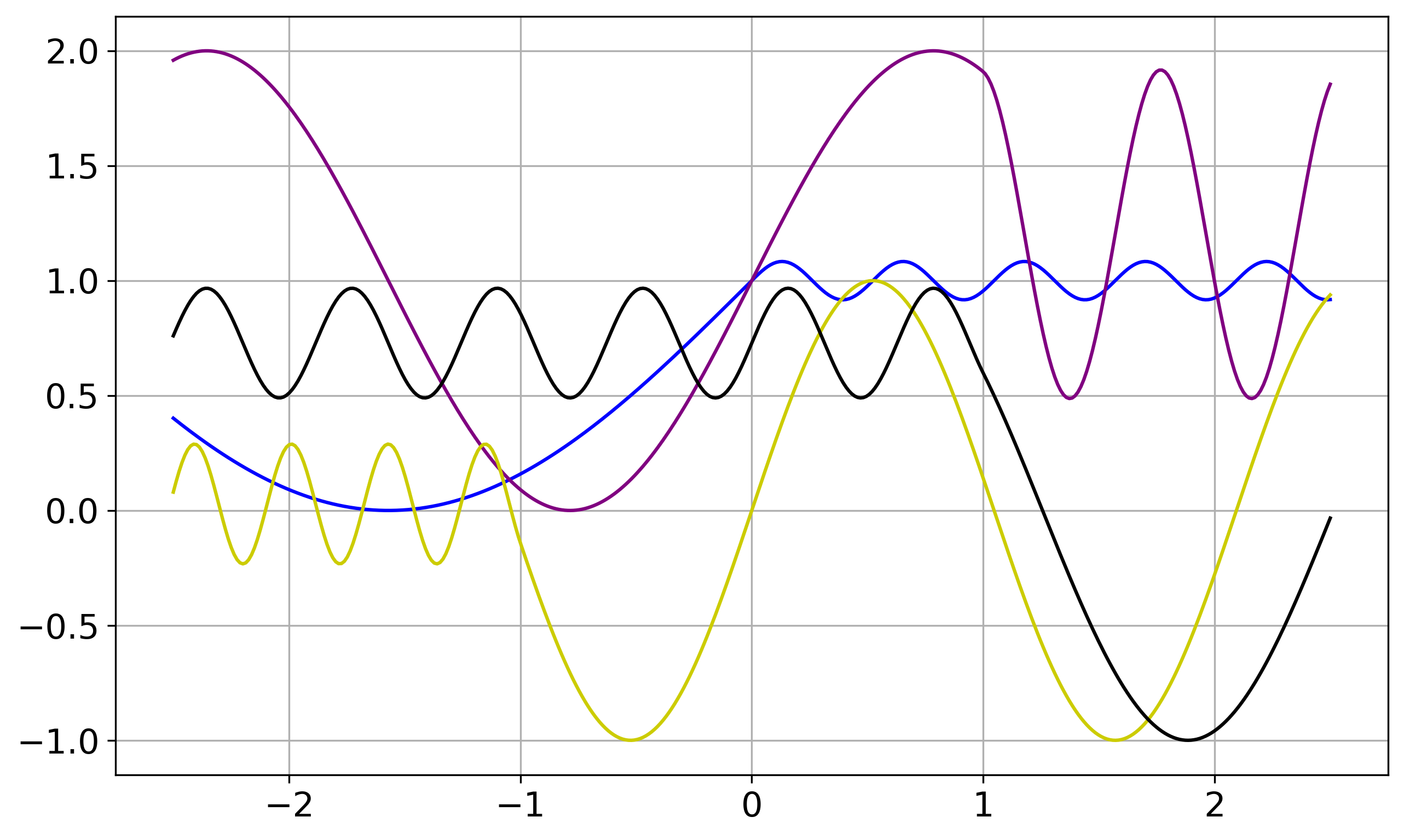}
  \caption{Hills-type functions}
    \end{subfigure}
    \caption{Signals $g_j$, $j=1,\ldots,4$ in the sparse high-dimensional additive model of the simulation in \citet{miftachov2025early}. \href{https://github.com/EarlyStop/EarlyStopping/blob/main/simulations/RegressionTree_additive_plots.py}{\faGithub}}
    \label{fig:tree_signals}
\end{figure}

In this section, we replicate the simulation study of \citet{miftachov2025early} by using the \pythoninline{RegressionTree} class. The Monte-Carlo simulation has dimensions $\dimension = 30$ with $n = 1000$ observations for both the training and the test set. The design is uniformly distributed with $X_1, \ldots, X_{30} \overset{\mathrm{iid}}{\sim} U(-2.5, 2.5)$, $i=1,\ldots,n$ and the response is generated by the additive non-parametric signal
\begin{align*}
    \truesignal(x) = g_1(x_1) + g_2(x_2) + g_3(x_3) + g_4(x_4),\quad x=(x_1,\ldots,x_{30})^{\top}\in\mathbb{R}^{30}.
\end{align*}
The functions $g_j$, for $j=1,\ldots,4$ are displayed in Figure \ref{fig:tree_signals} for different classes of functions.

By using the \pythoninline{SimulationData} class, these additive models are directly obtained. For example, the smooth additive model is generated by
\begin{python}
design, response, f = es.SimulationData.additive_smooth(
                sample_size=2000, noise_level=1)
\end{python}

We initialize an instance of \pythoninline{RegressionTree} in Codeblock \ref{code:regression_init}. Then, the \pythoninline{alg.iterate} function grows the regression tree, as described in \Cref{ssec_RegressionTrees}, until reaching the pre-specified \pythoninline{max_depth} parameter. The non-interpolated stopping iteration, referred to as \textit{global} early stopping in \citet{miftachov2025early}, is determined by \pythoninline{alg.get_discrepancy_stop}.

\begin{python}[caption={Discrepancy stop for the regression tree.},label=code:regression_init]
alg = es.RegressionTree(design=design, response=response, 
                min_samples_split=1, true_signal=f, true_noise_vector=noise)

alg.iterate(max_depth=30)
stopping_iteration = alg.get_discrepancy_stop(critical_value=1, 
                max_depth=30)
\end{python}

The squared bias, the variance and the balanced oracle iteration can be retrieved as shown in \Cref{code:oracle_regression_tree}. 
\begin{python}[caption={Oracle quantities for the regression tree.},label=code:oracle_regression_tree]
alg.bias2
alg.variance 
alg.get_balanced_oracle(max_depth=30)
\end{python}

\begin{figure}[htb]
    \centering
    \begin{subfigure}[b]{0.49\textwidth}
        \centering
          \includegraphics[width=\textwidth]{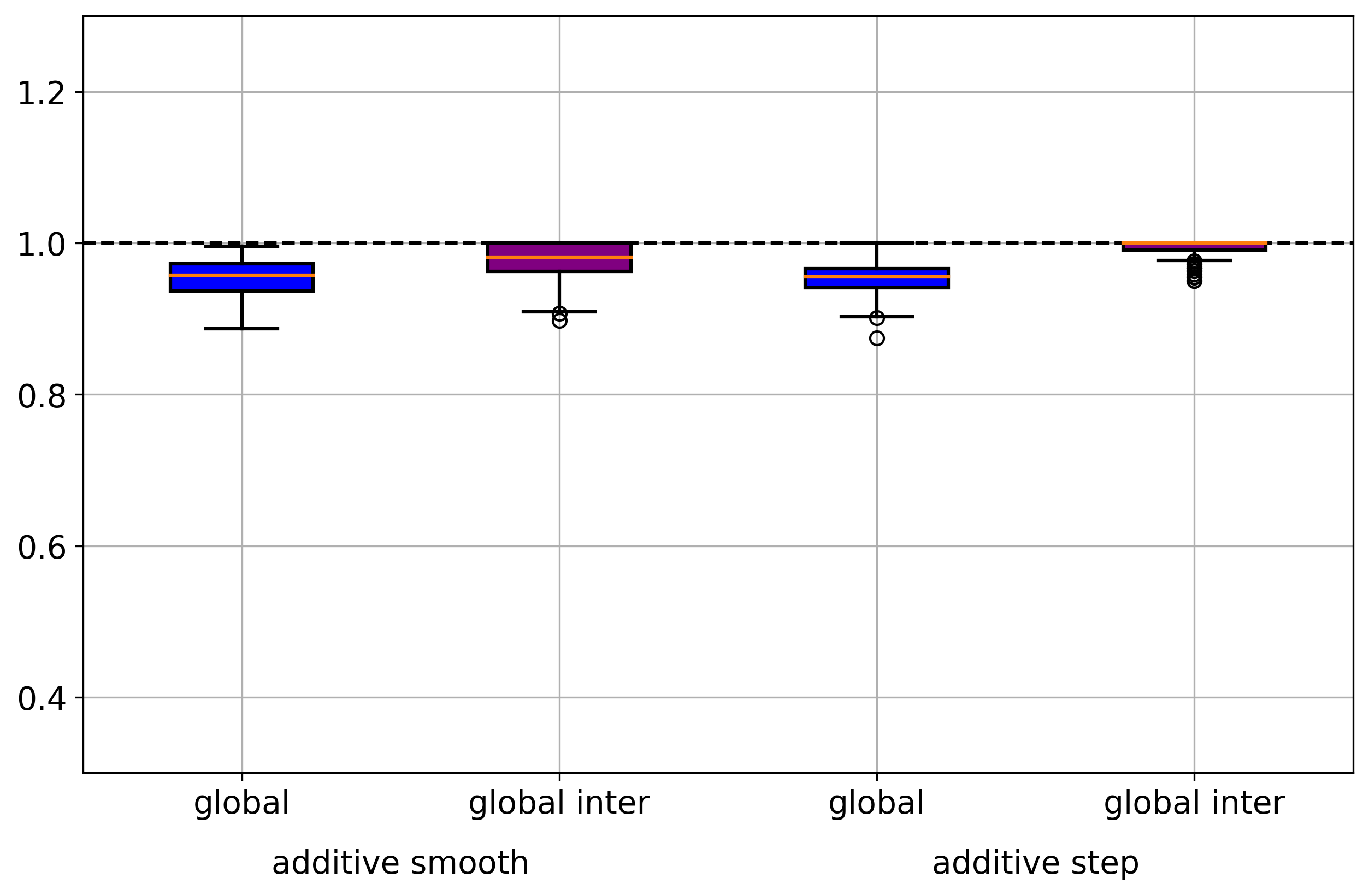}
          \caption{Smooth and step functions}
          \label{subfigure:regtree_plot_1}
    \end{subfigure}
    \hfill
    \begin{subfigure}[b]{0.49\textwidth}
        \centering
          \includegraphics[width=\textwidth]{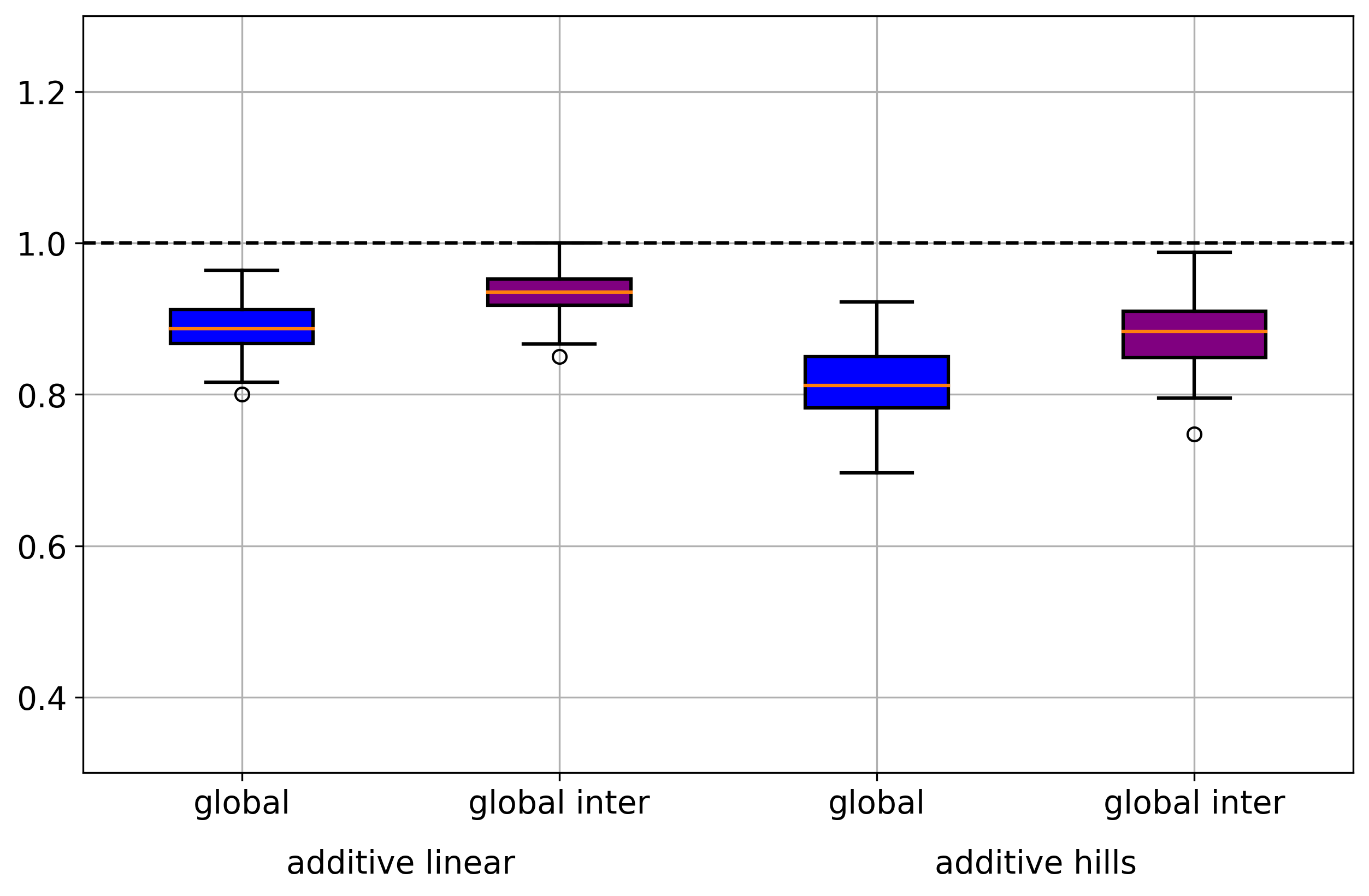}
          \caption{Linear and hills functions}
          \label{subfigure:regtree_plot_2}
    \end{subfigure}
    \caption{Relative efficiency for the global and interpolated regression tree estimator for different sparse high dimensional additive models, see \citet{miftachov2025early}. (a) additive smooth and additive step. (b) additive linear and additive hills. Higher values are better. \href{https://github.com/EarlyStop/EarlyStopping/blob/main/simulations/RegressionTree_Replication.py}{\faGithub}}
    \label{fig:simulation_tree}
\end{figure}

The regression tree is fitted on a training set and evaluated on an untouched test set, denoted by subscript $n^{\prime}$. The relative efficiency $\min _{t \in [0, n]}\|\estimator_{t} - f \|_{n^{\prime}} /\|\estimator_{\tau} - f \|_{n^{\prime}}$ is calculated for each Monte-Carlo iteration and displayed for each function in \Cref{fig:simulation_tree}. The global early stopping is based on the orthogonal projection \eqref{eq:orth_projection}, and the interpolated global early stopping uses the projection flow \eqref{eq:proj_flow}, see \Cref{ssec_RegressionTrees}.

We observe in \Cref{fig:simulation_tree} that the interpolation of the regression tree improves
prediction performance as the relative efficiency increases. The median relative efficiency is
approximately 0.9, which is high for a non-parametric adaptation problem. The reason is that the true underlying function is of sparse additive structure, and as shown in Proposition 1 of \citet{scornet2015consistency} the CART is able to successfully split only in the relevant coordinate directions. In particular, the sparse components contain relatively strong signals compared to the noise. Thus, the discrepancy principle is able to stop the algorithm very close to the oracle iteration. 
However, it gets slightly worse for the additive Hills-type functions since the stopping happens too late. This effect is evident from the tables in the simulation of \citet{miftachov2025early}, as the number of terminal
nodes in the early stopped tree is larger than that of the oracle.

For our replication study, we have used the stopping threshold $\kappa = \sigma^2$ to obtain the early stopped regression tree. In practice, the noise level $\sigma^2$ is usually unknown and must be estimated. A possible approach is the nearest-neighbour estimator proposed by \citet{devroye2018nearest} and used in \citet{miftachov2025early}, which is both simple and well-suited for random designs in higher-dimensional settings.

\subsection{Comparison between methods}

\begin{figure}[htb]
    \centering
    \begin{subfigure}[b]{0.49\textwidth}
        \centering
      \includegraphics[width=\textwidth]{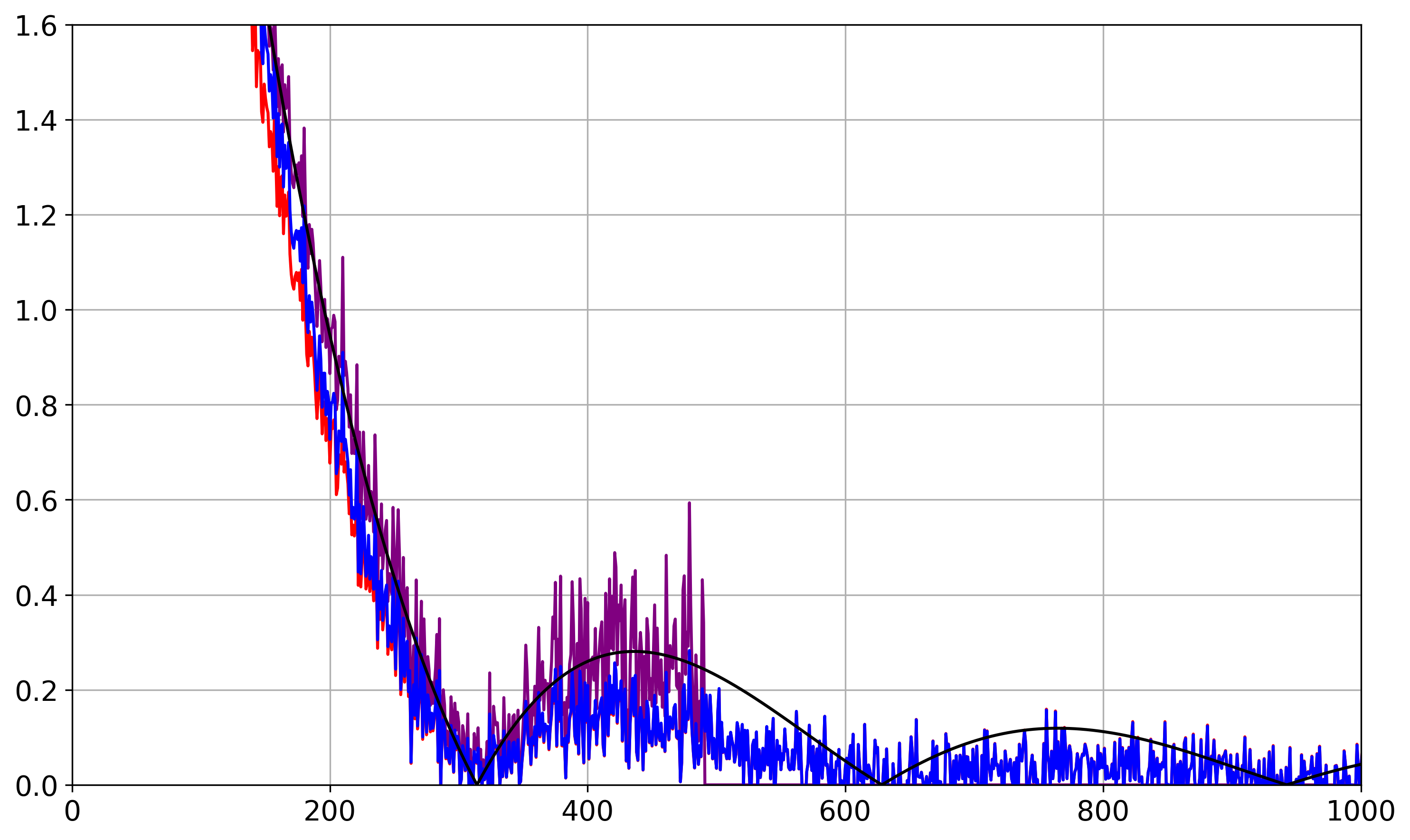}
      \caption{Signal estimate at $\tau$}
    \end{subfigure}
    \hfill
    \begin{subfigure}[b]{0.49\textwidth}
        \centering
  \includegraphics[width=\textwidth]{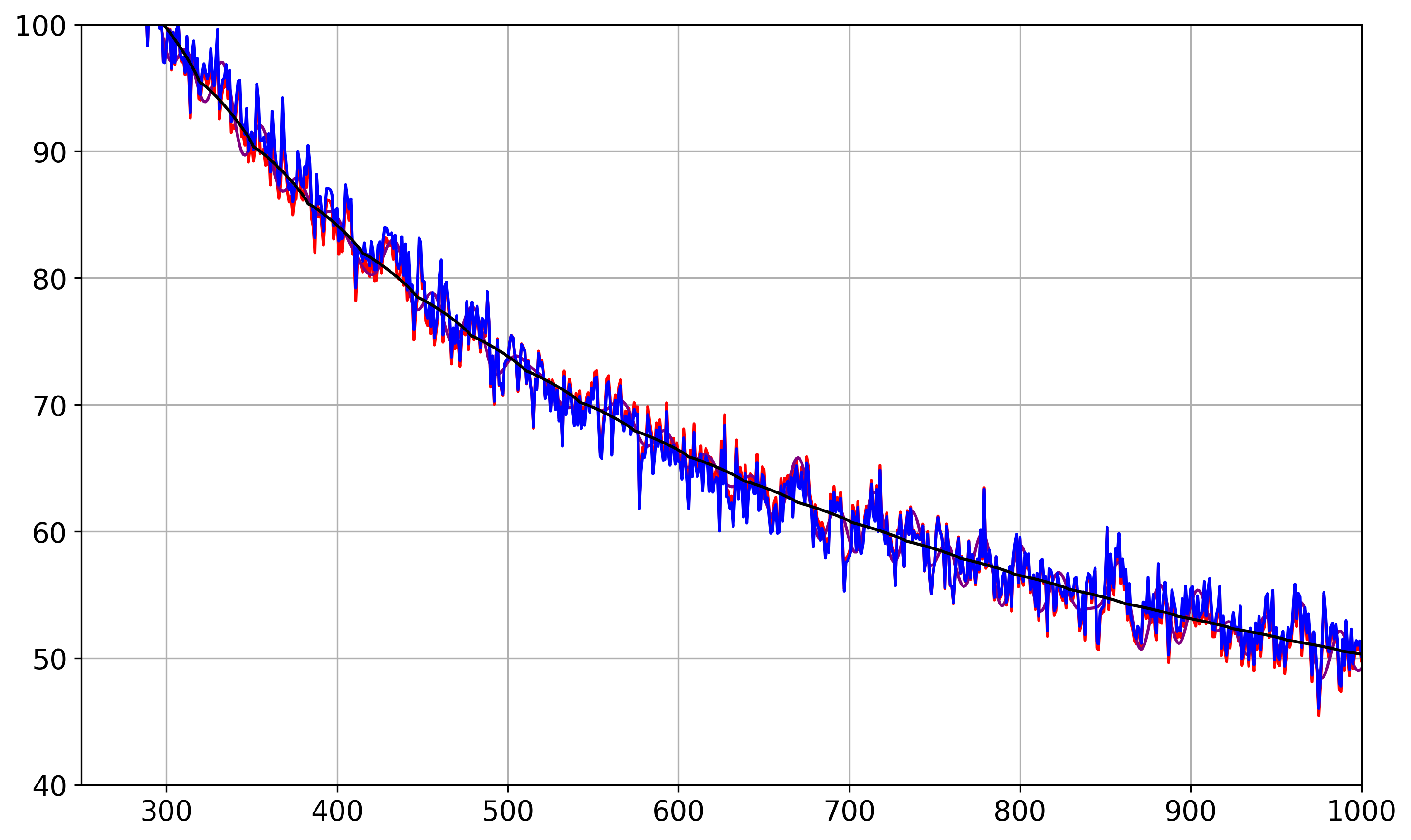}
  \caption{Fourier transform of signal estimate at $\tau$}
    \end{subfigure}
    \caption{Different estimated signals for the smooth example in \Cref{fig:simulation_svd}. Landweber (blue), tSVD (purple), conjugate gradient (red) and true signal (black).}
    \label{fig:signal_comparison}
\end{figure}

One of the main contributions of the \package{} is that its single coherent framework allows for
easy comparisons of different algorithms from the literature. 
As a simple example, in \Cref{fig:signal_comparison}, we compare the different estimates of the
smooth signal from \Cref{fig:signals} for the \pythoninline{Landweber}, \pythoninline{TruncatedSVD}
and \pythoninline{ConjugateGradients}-classes, respectively. 
While there is no clear structural difference between the \pythoninline{Landweber} and
\pythoninline{ConjugateGradients} estimate, the spectral cutoff of \pythoninline{TruncatedSVD} is markedly visible.

\begin{figure}[htb]
    \centering
    \begin{subfigure}[b]{0.49\textwidth}
        \centering
  \includegraphics[width=\textwidth]{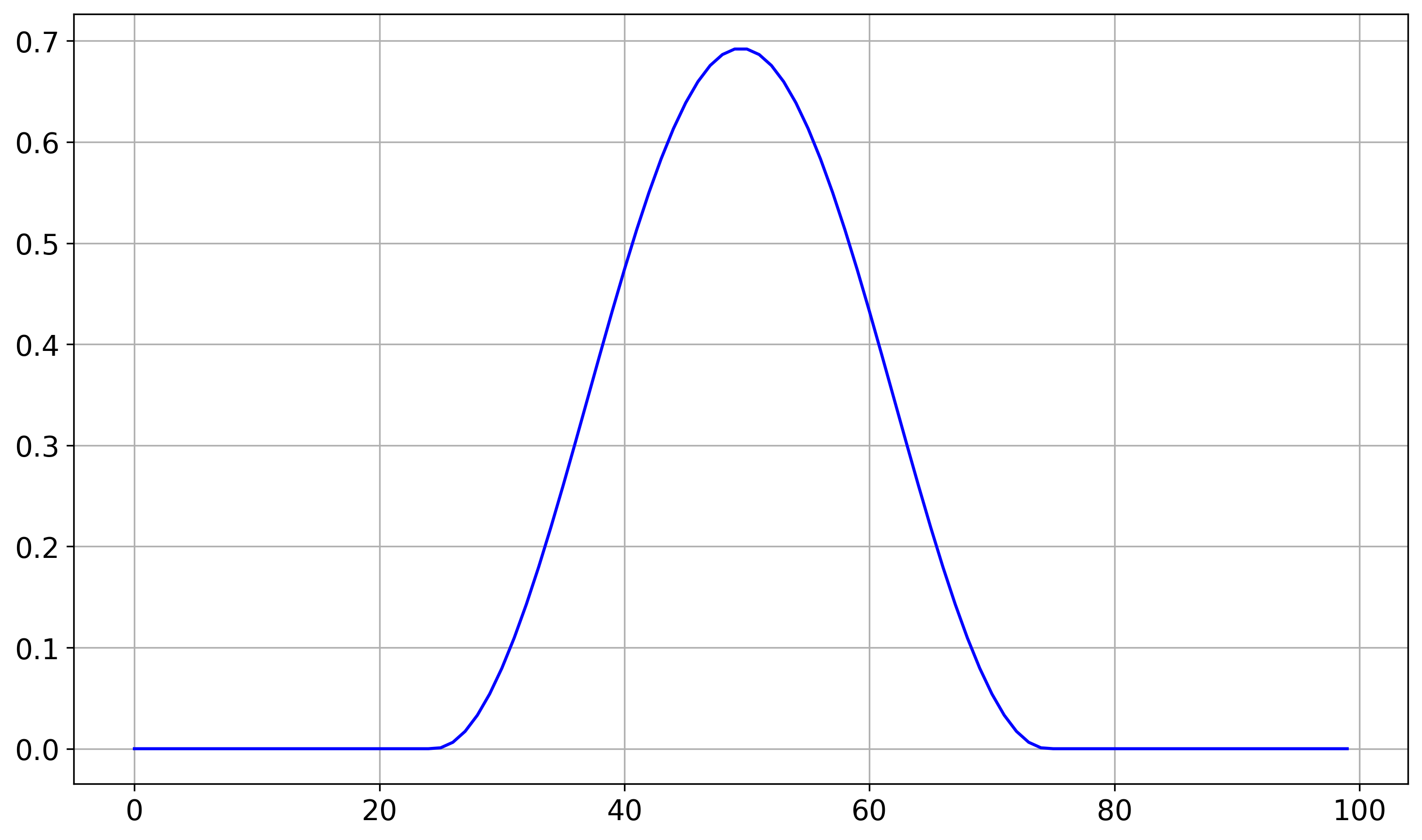}
  \caption{True signal}
    \end{subfigure}
    \hfill
    \begin{subfigure}[b]{0.49\textwidth}
        \centering
        \includegraphics[width=\textwidth]{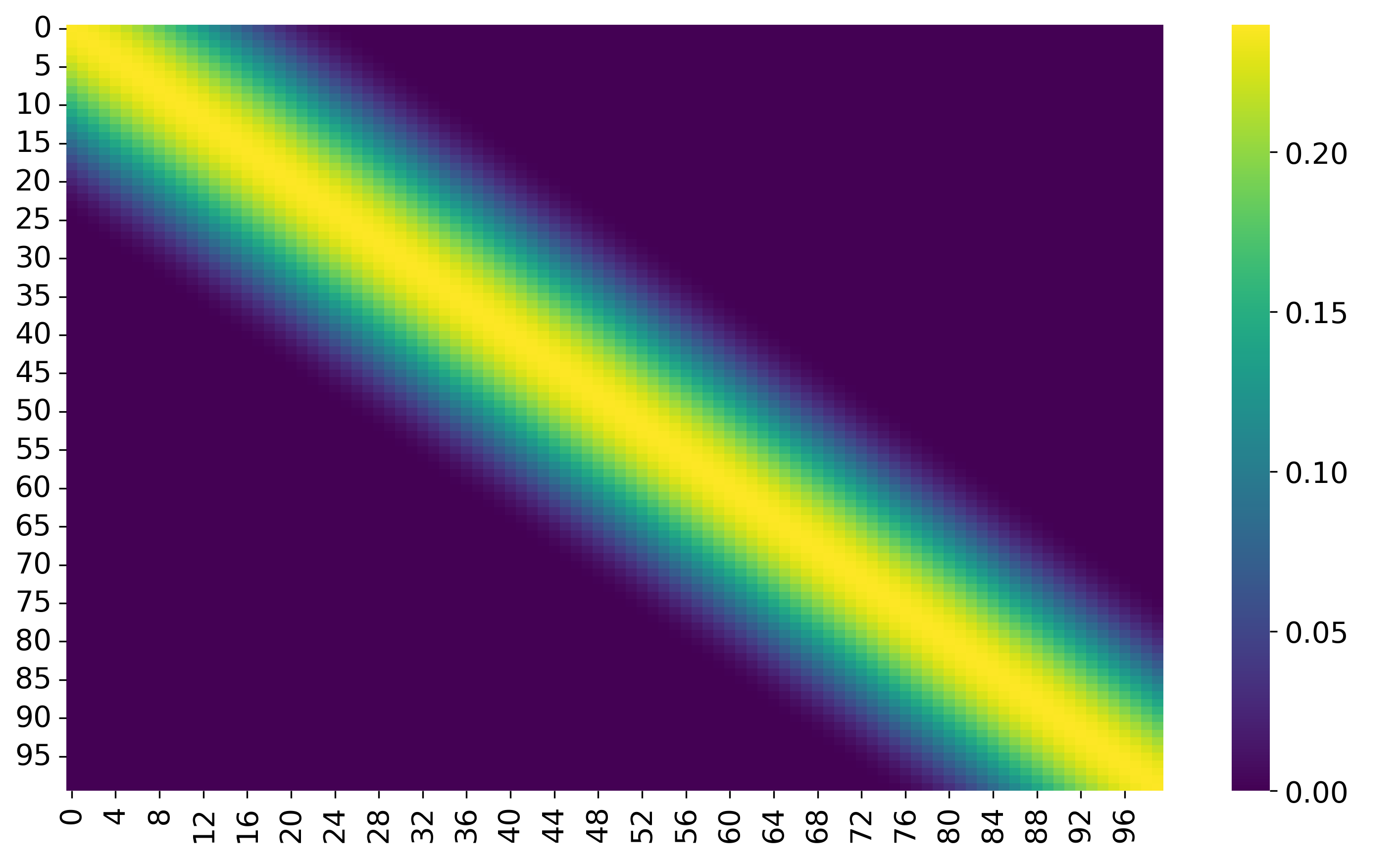}
  
  \caption{Design}
    \end{subfigure}
    \caption{Visualisation of the true signal $\truesignal$ and the design matrix $\designmatrix$ for the Phillips example.}
    \label{figure:phillips}
\end{figure}

More interestingly, we can also compare the performance of the same stopping method over the
different algorithms.
For a somewhat pathological example, let us consider the well-known Phillips' test problem, which
emerges as the discretisation of a Fredholm integral of the first-kind, see
\citet{regtools}. 
The true signal $\truesignal$ and design matrix $\designmatrix$ from the Phillips' example are visualised
in \Cref{figure:phillips}.

\begin{figure}[htb]
    \centering
    \begin{subfigure}[b]{0.49\textwidth}
          \centering
          \includegraphics[width=\textwidth]{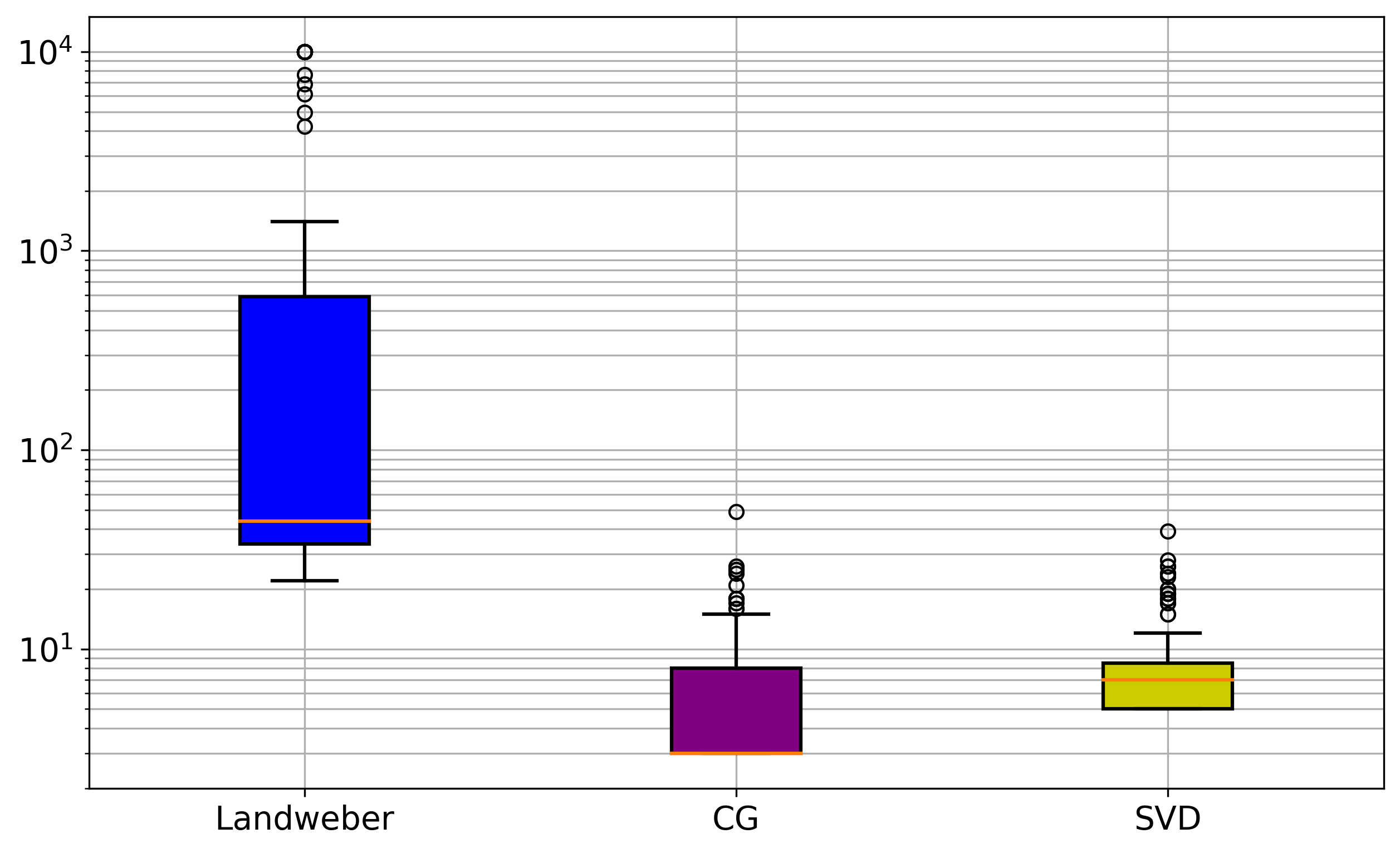}
          \caption{Stopping times}
          \label{fig:stopping_times_boxplots}
    \end{subfigure}
    \hfill
  \hfill
    \begin{subfigure}[b]{0.49\textwidth}
        \centering
        \includegraphics[width=\textwidth]{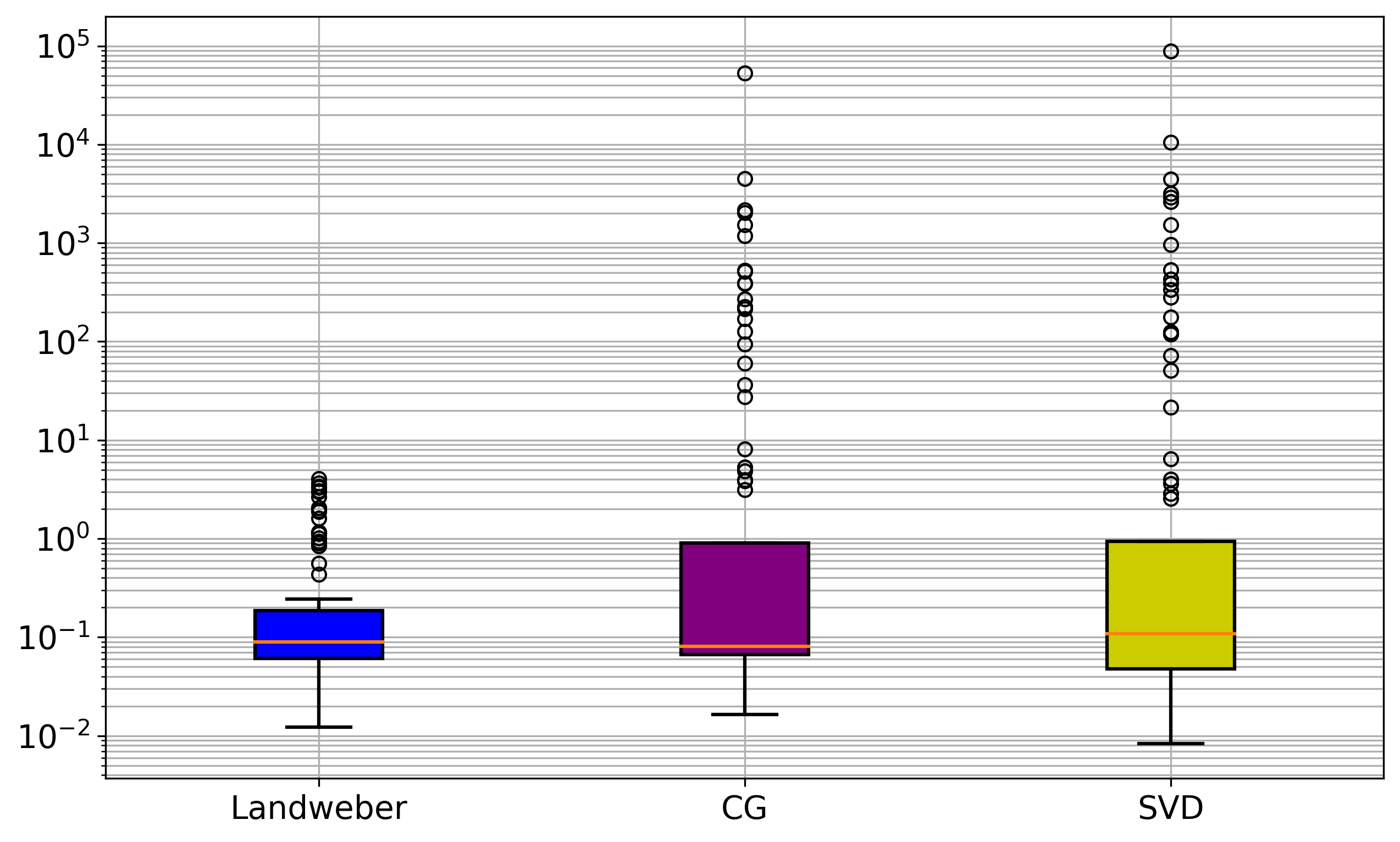}
        \caption{Empirical errors at $\tau^{\text{DP}}$}
  \label{fig:errors_phillips}
  \end{subfigure}
    \caption{Comparison of the relative efficiencies and variation of stopping times for the ill-posed Phillips problem from \citet{regtools}. Logarithmic scale capped at $10^{4}$. \href{https://github.com/EarlyStop/EarlyStopping/blob/main/simulations/ComparisonStudy.py}{\faGithub}}
\end{figure}

\Cref{fig:errors_phillips} shows a comparison between the strong empirical errors at the discrepancy stop
for the Landweber, truncated SVD and conjugate gradient algorithm on Phillips
example for $100$ Monte Carlo iterations, $\samplesize = 100$ and true noise level 
$\noiselevel = 0.1$. 
Due to its severe ill-posedness, the performance of the estimation procedures is visibly worse than
in previous examples, and the deviation of the discrepancy stop from the weak and strong balanced
oracle is larger and more susceptible to outliers. 
Comparatively, the latter effect is particularly severe for the Landweber iteration.
Having access to the oracle quantities of the simulation allows us to understand this difference on
deeper structural level.

We consider the truncated SVD procedure first and focus on the weak risk decomposition as the
discrepancy stopping time mimics the weak oracle.
From \eqref{eq:bal:oracle:truncated:SVD} and the lower bound in \citet{blanchard2018early}, we expect a difference of order
\( \sqrt{n} \delta^{2} = 0.1 \) between the optimal risk and the risk at the stopping time, which
provides a clear explanation for why the stopping times mostly stay below 50.
Both in the weak and the strong risk, this is enough to loose adaptivity.
However, this can be salvaged by applying the two-step procedure from \eqref{eq:two_step_tSVD}, since
we only lose about a factor of five in the computational load from the deviations in the stopping time.

\begin{table}[htb]
\centering
\begin{tabular}{l|l|l|l}
\hline
Data & \pythoninline{TruncatedSVD} & \pythoninline{Landweber} & \pythoninline{ConjugateGradients} \\ \hline
Gravity       & $1$   & 0.2 & $9\text{e-}4$ \\
Phillips'     & $1.8$ & 0.2 & $7.3\text{e-}3$  \\
Smooth        & $9\text{e-}4$ & $7\text{e-}3$ & $2\text{e-}4$ \\
Supersmooth   & $8\text{e-}4$  & $6\text{e-}3$ & $2\text{e-}4$ \\
Rough         & $9\text{e-}4$  & $6\text{e-}3$ & $2\text{e-}4$ \\
\hline
\end{tabular}
\caption{Average execution time for 100 iterations in seconds.}
\label{figure:times}
\end{table}

For the Landweber iteration, this changes structurally.
While, as in the case of both versions of the risk for truncated SVD, the strong risk still
exhibits a U-shape, this geometry is lost for the weak risk.
Consequently, deviations of the size \( \sqrt{n} \delta^{2} = 0.1 \) from the weak optimal risk can
result in enormous deviations in the stopping time.
The simulation shows deviations up to size \( 10^{4} \) because this was set as the maximal iteration but
even larger times of size \( n^{3} = 10^{6} \) can occur.
This situation, in which the deviations are of the same size \( n^{3} \) as the whole learning
trajectory constitutes a truly pathological example for early stopping since the two-step
procedure becomes qualitatively equivalent to model selection over the full path.



\begin{figure}[H]
    \centering
    \begin{subfigure}[b]{0.49\textwidth}
        \centering
        \includegraphics[width=\textwidth]{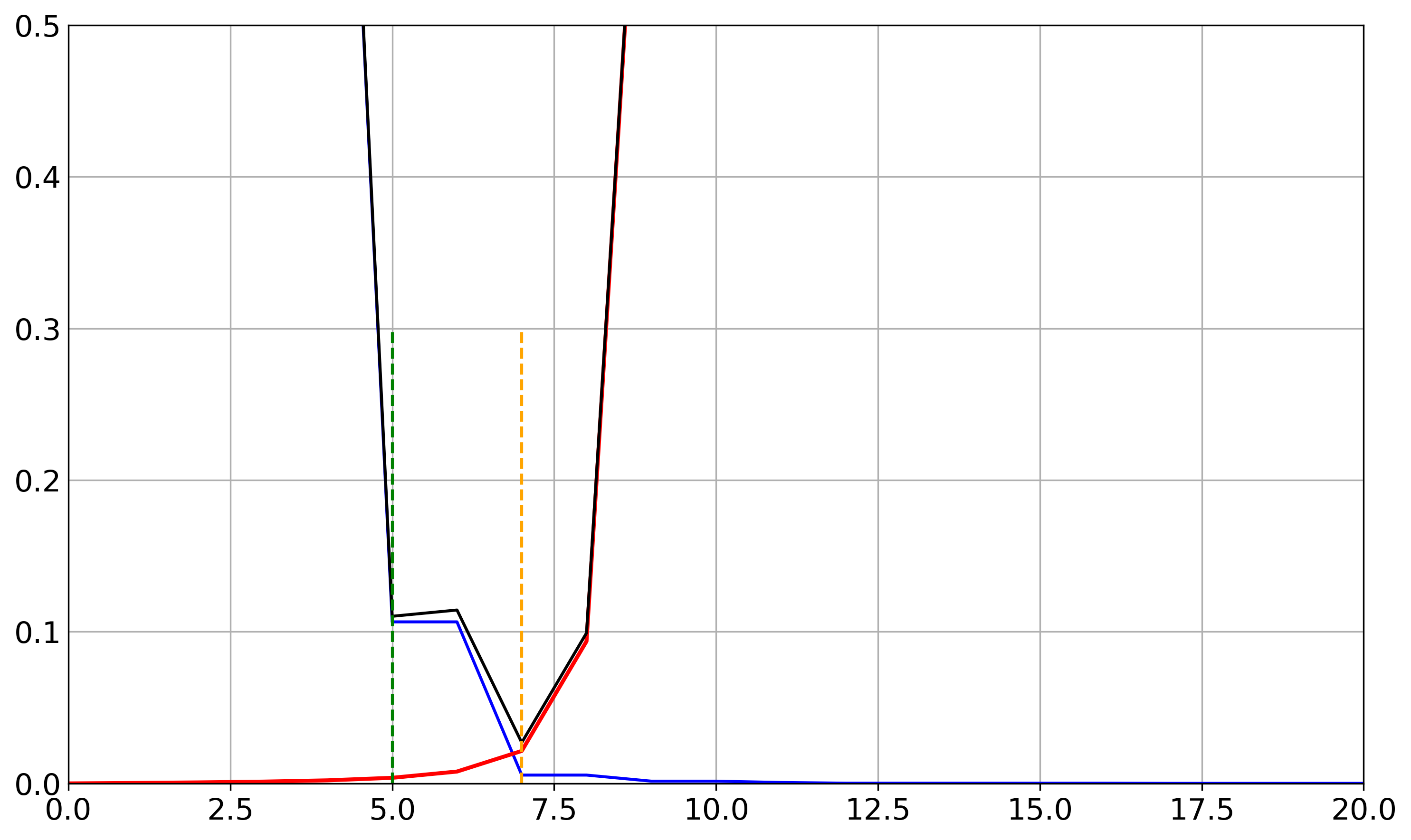}
        \caption{tSVD in strong norm}
        \label{fig:phillips_tSVD_strong}
    \end{subfigure}
    \hfill
    \begin{subfigure}[b]{0.49\textwidth}
        \centering
        \includegraphics[width=\textwidth]{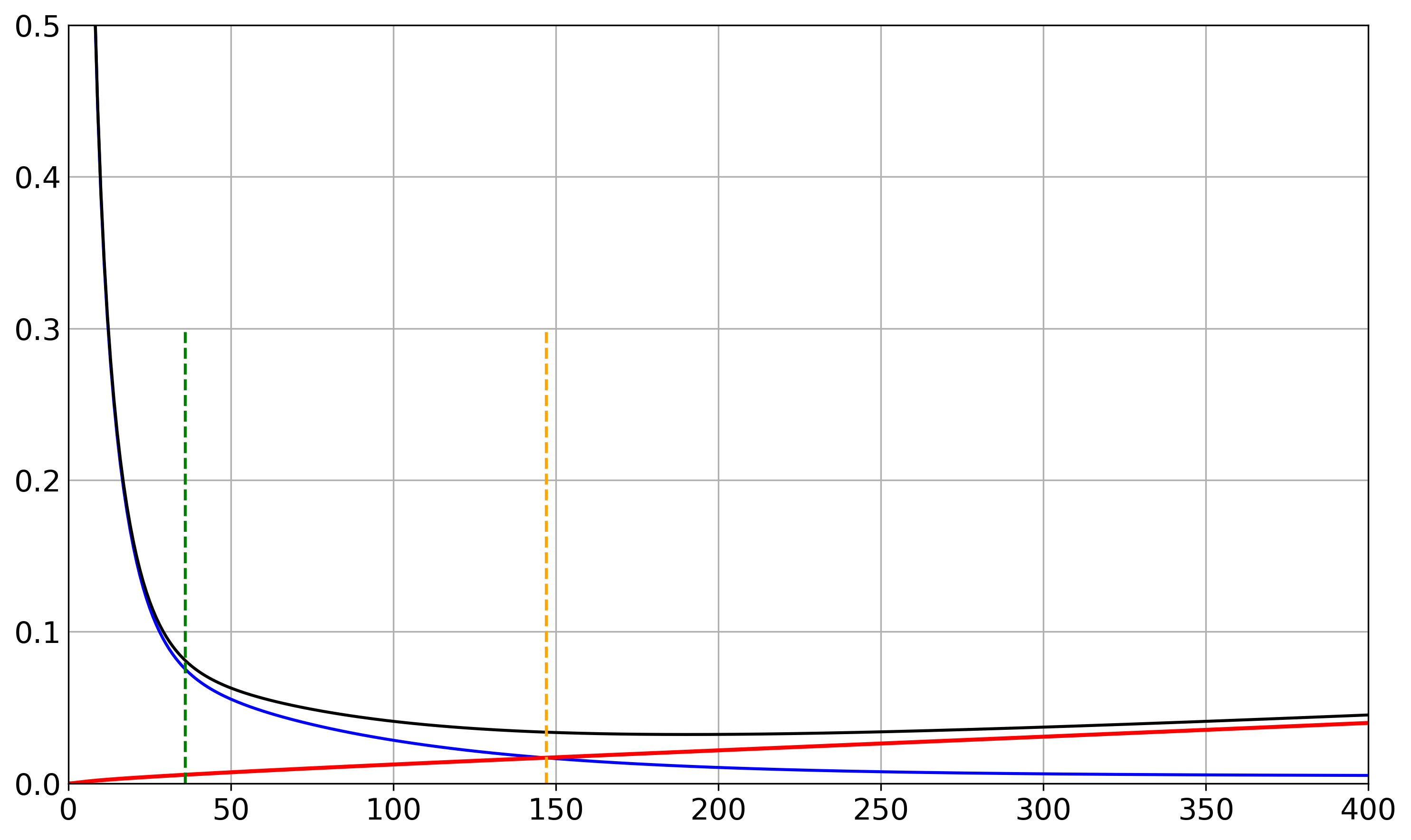}
        \caption{Landweber in strong norm}
        \label{fig:phillips_landweber_weak}
    \end{subfigure}\\
        \begin{subfigure}[b]{0.49\textwidth}
        \centering
                \includegraphics[width=\textwidth]{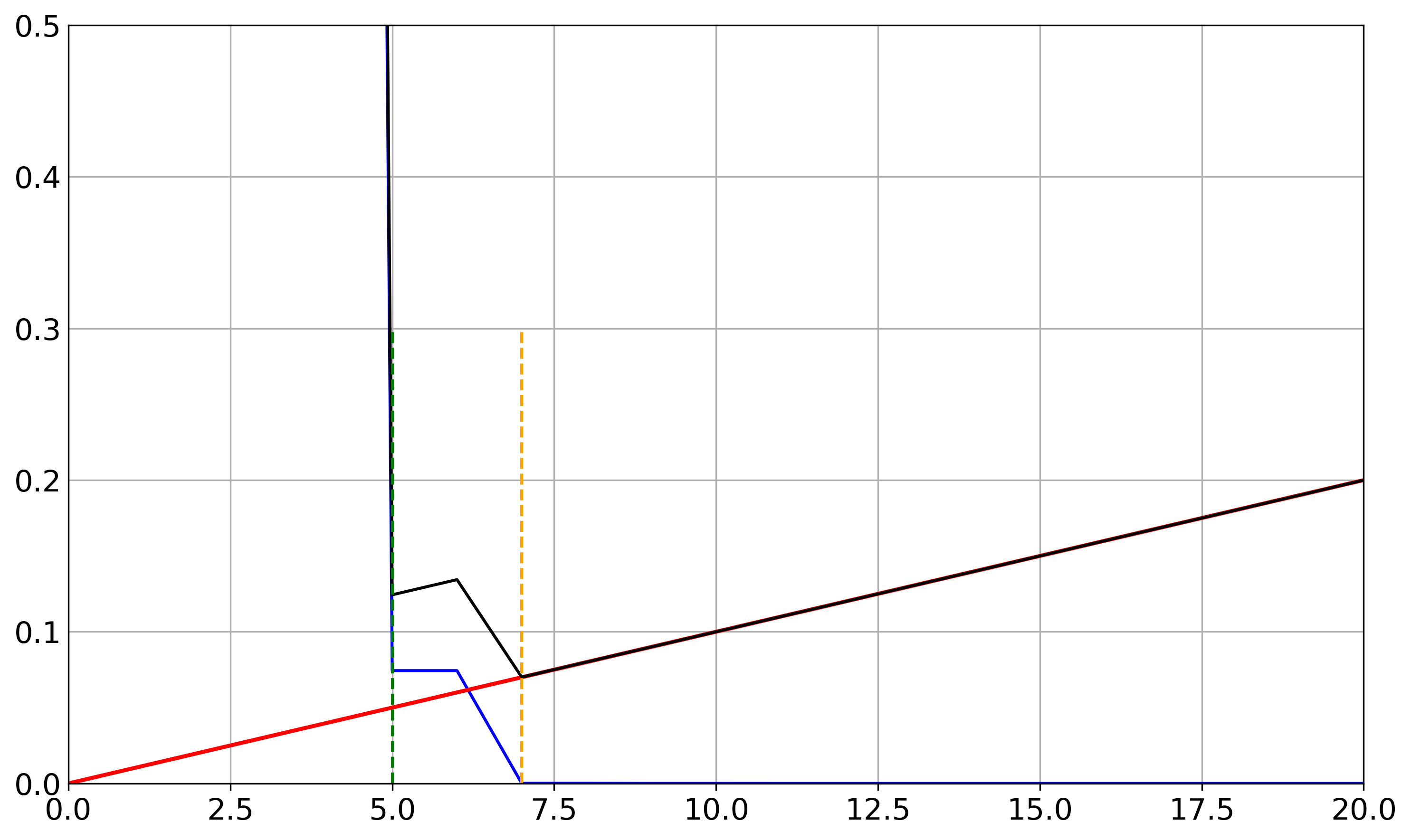}
        
        \caption{tSVD in weak norm}
        \label{fig:phillips_tSVD_weak}
    \end{subfigure}
    \hfill
    \begin{subfigure}[b]{0.49\textwidth}
        \centering
        \includegraphics[width=\textwidth]{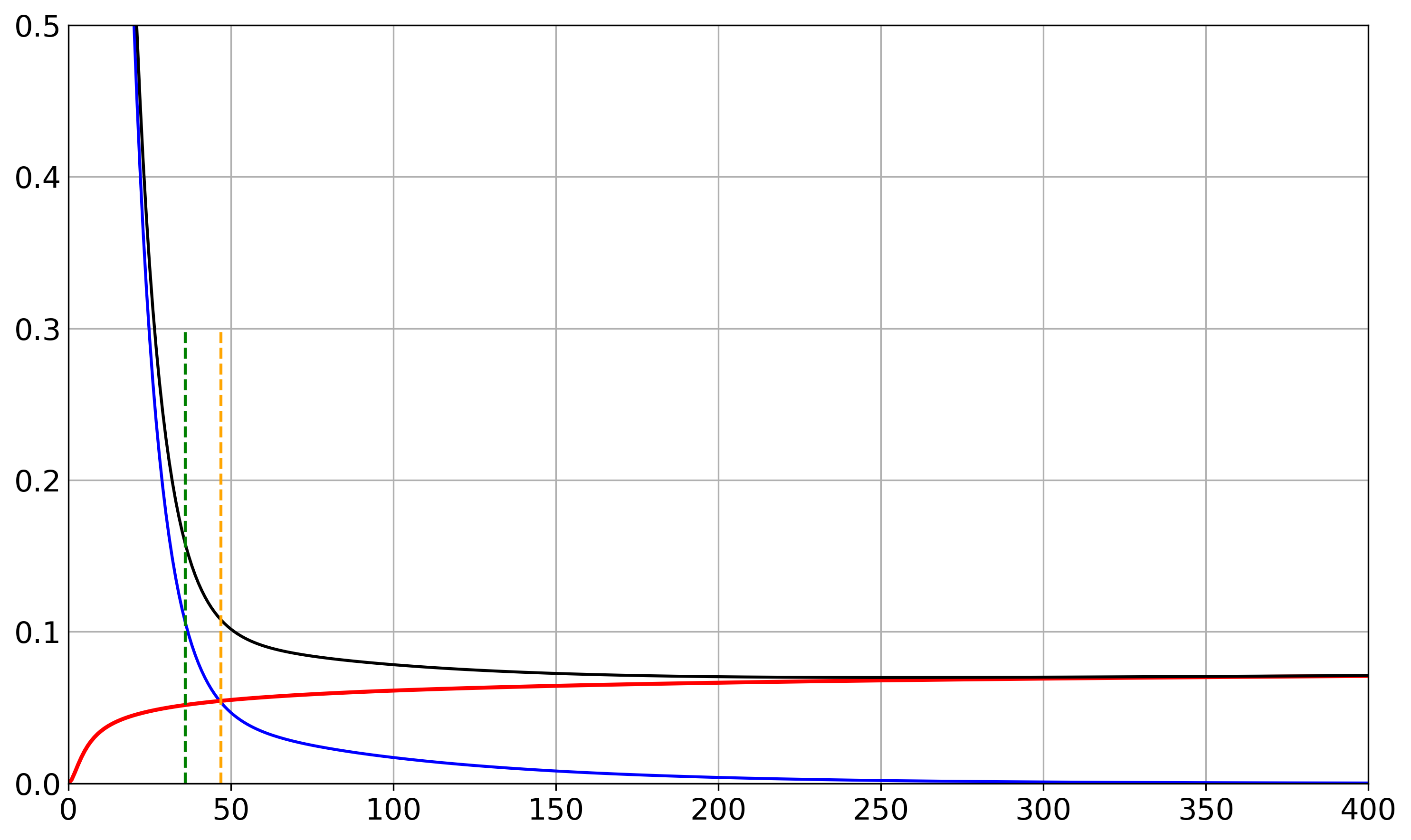}
        \caption{Landweber in weak norm}
   \label{fig:phillips_landweber_strong}
    \end{subfigure}
    \caption{Comparison of the risk (black), squared bias (blue), variance (red), discrepancy stop (green), balanced oracle (orange) in weak and strong norm for truncated SVD and the Landweber iteration for the Phillips example. \href{https://github.com/EarlyStop/EarlyStopping/blob/main/simulations/Simulation_counterexample_landweber.py}{\faGithub}}
    \label{fig:phillips}
\end{figure}
We conclude this section with a brief comparison of the average execution times for 100 iterations of \pythoninline{Landweber}, \pythoninline{ConjugateGradients} and \pythoninline{TruncatedSVD} compared across several different datasets given $\samplesize = 1000$, see \Cref{figure:times}. Note that these values do not include the additional runtime required for the computation of the theoretical quantities. 


\clearpage
\section*{Acknowledgments}
This research has been partially funded by the Deutsche Forschungsgemeinschaft
(DFG) – Project-ID 318763901 - SFB1294, Project-ID 460867398 - 
Research Unit 5381 and the German Academic Scholarship Foundation.
Co-funded by the European Union (ERC, BigBayesUQ, project number: 101041064).
Views and opinions expressed are, however, those of the author(s) only and
do not necessarily reflect those of the European Union or the European Research Council.
Neither the European Union nor the granting authority can be held responsible for them.

\bibliographystyle{abbrvnat}
\bibliography{references}

\end{document}